\documentclass[10pt]{article}
\usepackage{hyperref}
\usepackage{./TeX/note_pdf}
\usepackage{cancel}
\usepackage{algorithm}
\usepackage{algorithmic}

\usepackage{caption}
\begin{document}
%================%
%     Header     %
%================%
\begin{center}
\LARGE\textsf{\textbf{Tutorial on Diffusion Models for Imaging and Vision}}

\vspace{1ex}
\large{Stanley Chan}\footnote{School of Electrical and Computer Engineering, Purdue University, West Lafayette, IN 47907.

Email: \url{stanchan@purdue.edu}.}

\vspace{2ex}
\today
\end{center}

\normalsize
\textbf{Abstract}. The astonishing growth of generative tools in recent years has empowered many exciting applications in text-to-image generation and text-to-video generation. The underlying principle behind these generative tools is the concept of \emph{diffusion}, a particular sampling mechanism that has overcome some longstanding shortcomings in previous approaches. The goal of this tutorial is to discuss the essential ideas underlying these diffusion models. The target audience of this tutorial includes undergraduate and graduate students who are interested in doing research on diffusion models or applying these tools to solve other problems.

\tableofcontents

\newpage

\newpage
\section{Variational Auto-Encoder (VAE)}
\setcounter{figure}{0}

A long time ago, in a galaxy far far away, we wanted to build a generator — a generator that generates texts, speeches, or images from some inputs with which we give to the computer. While this may sound magical at first, the problem has actually been studied for a long time. To kick off the discussion of this tutorial, we shall first consider the \textbf{variational autoencoder} (VAE). VAE was proposed by Kingma and Welling in 2014 \cite{Kingma_2014_ICLR}. According to their 2019 tutorial \cite{Kingma_2019_VAE}, the VAE was inspired by the Helmholtz Machine \cite{Dayan_1995_Helmholtz} as the marriage of graphical models and deep learning. In what follows, we will discuss VAE's problem setting, its building blocks, and the optimization tools associated with the training.

\subsection{Building Blocks of VAE}
We start by discussing the schematic diagram of a VAE. As shown in the figure below, the VAE consists of a pair of models (often realized by deep neural networks). The one located near the input is called an \textbf{encoder} whereas the one located near the output is called a \textbf{decoder}. We denote the input (typically an image) as a vector $\vx$, and the output (typically another image) as a vector $\widehat{\vx}$. The vector located in the middle between the encoder and the decoder is called a \textbf{latent variable}, denoted as $\vz$. The job of the encoder is to extract a meaningful representation for $\vx$, whereas the job of the decoder is to generate a new image from the latent variable $\vz$.

\begin{figure}[h]
\centering
\includegraphics[width=0.4\linewidth]{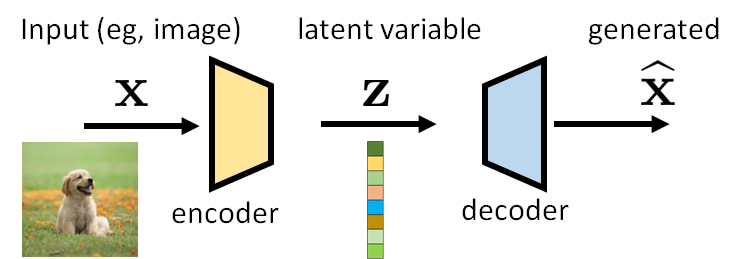}
\caption{A variational autoencoder consists of an encoder that converts an input $\vx$ to a latent variable $\vz$, and a decoder that synthesizes an output $\widehat{\vx}$ from the latent variable.}
\end{figure}

The latent variable $\vz$ has two special roles in this setup. With respect to the input, the latent variable encapsulates the information that can be used to describe $\vx$. The encoding procedure could be a lossy process, but our goal is to preserve the important content of $\vx$ as much as we can. With respect to the output, the latent variable serves as the ``seed'' from which an image $\widehat{\vx}$ can be generated. Two different $\vz$'s should in theory give us two different generated images.

A slightly more formal definition of a latent variable is given below.
\boxeddef{\textbf{Latent Variables\cite{Kingma_2019_VAE}}. In a probabilistic model, latent variables $\vz$ are variables that we do not observe and hence are not part of the training dataset, although they are part of the model.
}

\boxedeg{
Getting a latent representation of an image is not an alien thing. Back in the time of JPEG compression (which is arguably a dinosaur), we used discrete cosine transform (DCT) basis functions $\vvarphi_n$ to encode the underlying image/patches of an image. The coefficient vector $\vz = [z_1,\ldots,z_N]^T$ is obtained by projecting the image $\vx$ onto the space spanned by the basis, via $z_n = \langle \vvarphi_n, \vx \rangle$. So, given an image $\vx$, we can produce a coefficient vector $\vz$. From $\vz$, we can use the inverse transform to recover (i.e. decode) the image.
\begin{center}
\includegraphics[width=0.5\linewidth]{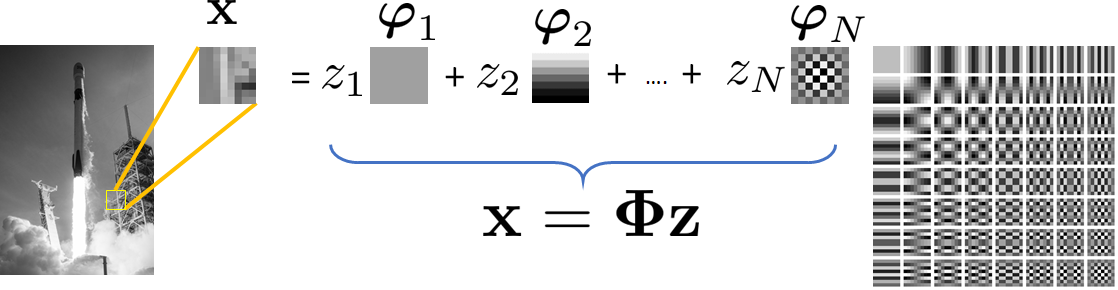}
\captionof{figure}{In discrete cosine transform (DCT), we can think of the encoder as taking an image $\vx$ and generating a latent variable $\vz$ by projecting $\vx$ onto the basis functions.}
\end{center}

In this example, the coefficient vector $\vz$ is the latent variable. The encoder is the DCT transform, and the decoder is the inverse DCT transform.
}

The term ``variational'' in VAE is related to the subject of calculus of variations which studies optimization over functions. In VAE, we are interested in searching for the optimal probability distributions to describe $\vx$ and $\vz$. In light of this, we need to consider a few distributions:
\begin{itemize}
\setlength\itemsep{-1ex}
\item $p(\vx)$: The true distribution of $\vx$. It is never known. The whole universe of diffusion models is to find ways to draw samples from $p(\vx)$. If we knew $p(\vx)$ (say, we have a formula that describes $p(\vx)$), we can just draw a sample $\vx$ that maximizes $\log p(\vx)$.
\item $p(\vz)$: The distribution of the latent variable. Typically, we make it a zero-mean unit-variance Gaussian $\calN(0,\mI)$. One reason is that linear transformation of a Gaussian remains a Gaussian, and so this makes the data processing easier. Doersch \cite{Doersch_2016_VAE} also has an excellent explanation. It was mentioned that any distribution can be generated by mapping a Gaussian through a sufficiently complicated function. For example, in a one-variable setting, the inverse cumulative distribution function (CDF) technique \cite[Chapter 4]{Chan_2021_book} can be used for any continuous distribution with an invertible CDF. In general, as long as we have a sufficiently powerful function (e.g., a neural network), we can learn it and map the i.i.d. Gaussian to whatever latent variable needed for our problem.
\item $p(\vz|\vx)$: The conditional distribution associated with the \textbf{encoder}, which tells us the likelihood of $\vz$ when given $\vx$. We have no access to it. $p(\vz|\vx)$ itself is \emph{not} the encoder, but the encoder has to do something so that it will behave consistently with $p(\vz|\vx)$.
\item $p(\vx|\vz)$: The conditional distribution associated with the \textbf{decoder}, which tells us the posterior probability of getting $\vx$ given $\vz$. Again, we have no access to it.
\end{itemize}

When we switch from the classical parameteric models to deep neural networks, the notion of latent variables is changed to \emph{deep} latent variables. Kingma and Welling \cite{Kingma_2019_VAE} gave a good definition below.
\boxeddef{
\textbf{Deep Latent Variables}\cite{Kingma_2019_VAE}. Deep Latent Variables are latent variables whose distributions $p(\vz)$, $p(\vx|\vz)$, or $p(\vz|\vx)$ are parameterized by a neural network.
}
The advantage of deep latent variables is that they can model very complex data distributions $p(\vx)$ even though the structures of the prior distributions and the conditional distributions are relatively simple (e.g. Gaussian). One way to think about this is that the neural networks can be used to estimate the mean of a Gaussian. Although the Gaussian itself is simple, the mean is a function of the input data, which passes through a neural network to generate a data-dependent mean. So the expressiveness of the Gaussian is significantly improved.

Let's go back to the four distributions above. Here is a somewhat trivial but educational example that can illustrate the idea:
\boxedeg{
Consider a random variable $\mX$ distributed according to a Gaussian mixture model with a latent variable $z \in \{1,\ldots,K\}$ denoting the cluster identity such that $p_Z(k) = \Pb[Z = k] = \pi_k$ for $k = 1,\ldots,K$. We assume $\sum_{k=1}^K \pi_k = 1$. Then, if we are told that we need to look at the $k$-th cluster only, the conditional distribution of $\mX$ given $Z$ is
\begin{align*}
p_{\mX|Z}(\vx|k) = \calN(\vx \,|\, \vmu_k, \sigma^2_k\mI).
\end{align*}
The marginal distribution of $\vx$ can be found using the law of total probability, giving us
\begin{equation}
p_{\mX}(\vx) = \sum_{k=1}^K p_{\mX|Z}(\vx|k) p_Z(k) = \sum_{k=1}^K \pi_k \calN(\vx \,|\, \vmu_k, \sigma_k^2\mI).
\end{equation}
Therefore, if we start with $p_{\mX}(\vx)$, the design question for the encoder is to build a magical encoder such that for every sample $\vx \sim p_{\mX}(\vx)$, the latent code will be $z \in \{1,\ldots,K\}$ with a distribution $z \sim p_Z(k)$.

To illustrate how the encoder and decoder work, let's assume that the mean and variance are known and are fixed. Otherwise we will need to estimate the mean and variance through an expectation-maximization (EM) algorithm. It is doable, but the tedious equations will defeat the educational purpose of this illustration.

\textbf{Encoder}: How do we obtain $z$ from $\vx$? This is easy because at the encoder, we know $p_{\mX}(\vx)$ and $p_{Z}(k)$. Imagine that you only have two classes $z \in \{1,2\}$. Effectively you are just making a binary decision of where the sample $\vx$ should belong to. There are many ways you can do the binary decision. If you like the maximum-a-posteriori decision rule, you can check
\begin{equation*}
p_{Z|\mX}(1|\vx) \gtrless^{\text{class 1}}_{\text{class 2}} p_{Z|\mX}(2|\vx),
\end{equation*}
and this will return you a simple decision: You give us $\vx$, we tell you $z \in \{1,2\}$.

\textbf{Decoder}: On the decoder side, if we are given a latent code $z \in \{1,\ldots,K\}$, the magical decoder just needs to return us a sample $\vx$ which is drawn from $p_{\mX|Z}(\vx|k) = \calN(\vx \,|\, \vmu_k, \sigma^2_k\mI)$. A different $z$ will give us one of the $K$ mixture components. If we have enough samples, the overall distribution will follow the Gaussian mixture.
}
This example is certainly oversimplified because real-world problems can be much harder than a Gaussian mixture model with known means and known variances. But one thing we realize is that if we want to find the magical encoder and decoder, we must have a way to find the two conditional distributions $p(\vz|\vx)$ and $p(\vx|\vz)$. However, they are both high-dimensional.

In order for us to say something more meaningful, we need to impose additional structures so that we can generalize the concept to harder problems. To this end, we consider the following two proxy distributions:
\begin{itemize}
\setlength\itemsep{-1ex}
\item $q_{\vphi}(\vz|\vx)$: The proxy for $p(\vz|\vx)$, which is also the distribution associated with the \emph{encoder}. $q_{\vphi}(\vz|\vx)$ can be any directed graphical model and it can be parameterized using deep neural networks \cite[Section 2.1]{Kingma_2019_VAE}. For example, we can define
    \begin{align}
    (\vmu, \vsigma^2) &= \text{EncoderNetwork}_{\vphi}(\vx), \notag \\
    q_{\vphi}(\vz|\vx)   &= \calN(\vz \;|\; \vmu, \text{diag}(\vsigma^2)).
    \end{align}
    This model is a widely used because of its tractability and computational efficiency.
\item $p_{\vtheta}(\vx|\vz)$: The proxy for $p(\vx|\vz)$, which is also the distribution associated with the \emph{decoder}. Like the encoder, the decoder can be parameterized by a deep neural network. For example, we can define
    \begin{align}
    f_{\vtheta}(\vz)               &= \text{DecoderNetwork}_{\vtheta}(\vz), \notag \\
    p_{\vtheta}(\vx|\vz) &= \calN(\vx \;|\; f_{\vtheta}(\vz), \sigma_{\text{dec}}^2\mI),
    \end{align}
    where $\sigma_{\text{dec}}$ is a hyperparameter that can be pre-determined or it can be learned.
\end{itemize}

The relationship between the input $\vx$ and the latent $\vz$, as well as the conditional distributions, are summarized in \fref{fig: econder-decoder, p q}. There are two nodes $\vx$ and $\vz$. The ``forward'' relationship is specified by $p(\vz|\vx)$ (and approximated by $q_{\vphi}(\vz|\vx)$), whereas the ``reverse'' relationship is specified by $p(\vx|\vz)$ (and approximated by $p_{\vtheta}(\vx|\vz)$).

\begin{figure}[h]
\vspace{-2ex}
\centering
\includegraphics[width=0.3\linewidth]{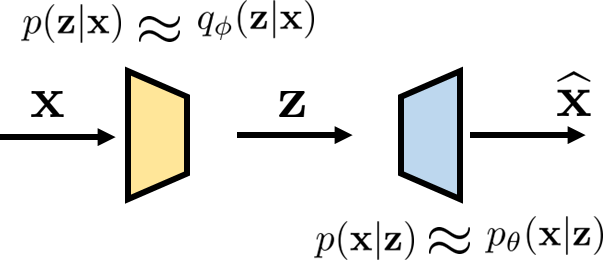}
\caption{In a variational autoencoder, the variables $\vx$ and $\vz$ are connected by the conditional distributions $p(\vx|\vz)$ and $p(\vz|\vx)$. To make things work, we introduce proxy distributions $p_{\vtheta}(\vx|\vz)$ and $q_{\vphi}(\vz|\vx)$.}
\label{fig: econder-decoder, p q}
\end{figure}

\boxedeg{
Suppose that we have a random variable $\vx \in \R^d$ and a latent variable $\vz \in \R^d$ such that
\begin{align*}
\vx & \sim  p(\vx) = \calN(\vx \,|\,\vmu,\sigma^2\mI),\\
\vz & \sim  p(\vz) = \calN(\vz \,|\,0,\mI).
\end{align*}
We want to construct a VAE. By this, we mean that we want to build two mappings Encoder$(\cdot)$ and Decoder$(\cdot)$. The encoder will take a sample $\vx$ and map it to the latent variable $\vz$, whereas the decoder will take the latent variable $\vz$ and map it to the generated variable $\widehat{\vx}$. If we \emph{knew} what $p(\vx)$ is, then there is a trivial solution where $\vz = (\vx-\vmu)/\sigma$ and $\widehat{\vx} = \vmu+\sigma\vz$. In this case, the true distributions can be determined and they can be expressed in terms of delta functions:
\begin{align*}
p(\vx|\vz) &= \delta\left(\vx-(\sigma\vz+\vmu)\right),\\
p(\vz|\vx) &= \delta\left(\vz- (\vx-\vmu)/\sigma\right).
\end{align*}

Suppose now that we do not know $p(\vx)$ so we need to build an encoder and a decoder to estimate $\vz$ and $\widehat{\vx}$. Let's first define the encoder. Our encoder in this example takes the input $\vx$ and generates a pair of parameters $\widehat{\vmu}(\vx)$ and $\widehat{\sigma}(\vx)^2$, denoting the parameters of a Gaussian. Then, we define $q_{\vphi}(\vz|\vx)$ as a Gaussian:
\begin{align*}
(\widehat{\vmu}(\vx), \; \widehat{\sigma}(\vx)^2)    &= \text{Encoder}_{\vphi}(\vx),\\
q_{\vphi}(\vz|\vx)                      &= \calN(\vz\;|\; \widehat{\vmu}(\vx), \; \widehat{\sigma}(\vx)^2\mI).
\end{align*}
For the purpose of discussion, we assume that $\widehat{\vmu}$ is an affine function of $\vx$ such that $\widehat{\vmu}(\vx) = a\vx+\vb$ for some parameters $a$ and $\vb$. Similarly, we assume that $\widehat{\sigma}(\vx)^2 = t^2$ for some scalar $t$. This will give us
\begin{align*}
q_{\vphi}(\vz|\vx) = \calN(\vz\;|\; a\vx+\vb, t^2\mI).
\end{align*}

For the decoder, we deploy a similar structure by considering
\begin{align*}
(\widetilde{\vmu}(\vz), \; \widetilde{\sigma}(\vz)^2)  &= \text{Decoder}_{\vtheta}(\vz),\\
p_{\vtheta}(\vx|\vz)                         &= \calN(\vx\;|\; \widetilde{\vmu}(\vz), \; \widetilde{\sigma}(\vz)^2\mI).
\end{align*}
Again, for the purpose of discussion, we assume that $\widetilde{\vmu}$ is affine so that $\widetilde{\vmu}(\vz) = c\vz+\vv$ for some parameters $c$ and $\vv$ and $\widetilde{\sigma}(\vz)^2 = s^2$ for some scalar $s$. Therefore, $p_{\vtheta}(\vx|\vz)$ takes the form of:
\begin{align*}
p_{\vtheta}(\vx|\vz)  = \calN(\vz\;|\; c\vx+\vv, s^2\mI).
\end{align*}
We will discuss how to determine the parameters later.
}

\subsection{Evidence Lower Bound}
How do we use these two proxy distributions to achieve our goal of determining the encoder and the decoder? If we treat $\vphi$ and $\vtheta$ as optimization variables, then we need an objective function (or the loss function) so that we can optimize $\vphi$ and $\vtheta$ through training samples. The loss function we use here is called the Evidence Lower BOund (ELBO) \cite{Kingma_2019_VAE}:
\boxeddef{
\label{def: ELBO}
(\textbf{Evidence Lower Bound}) The Evidence Lower Bound is defined as
\begin{equation}
\text{ELBO}(\vx) \bydef \E_{q_{\phi}(\vz|\vx)}\left[ \log \frac{p(\vx,\vz)}{q_{\vphi}(\vz|\vx)} \right].
\end{equation}
}
You are certainly puzzled how on the Earth people can come up with this loss function!? Let's see what ELBO means and how it is derived.

\begin{figure}[h]
\centering
\includegraphics[width=0.4\linewidth]{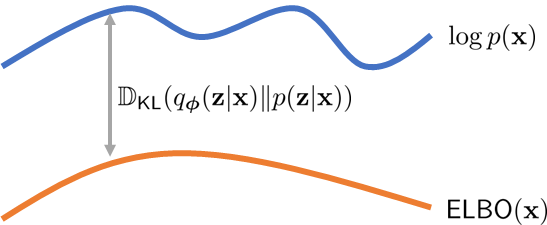}
\caption{Visualization of $\log p(\vx)$ and ELBO. The gap between the two is determined by the KL divergence $\mathbb{D}_{\text{KL}}( q_{\vphi}(\vz|\vx) \| p(\vz|\vx))$.}
\label{fig: ELBO gap}
\end{figure}

In a nutshell, ELBO is a \textbf{lower bound} for the prior distribution $\log p(\vx)$ because we can show that
\begin{align}
\log p(\vx)
&= \text{some magical steps to be derived} \notag \\
&= \E_{q_{\phi}(\vz|\vx)}\left[ \log \frac{p(\vx,\vz)}{q_{\vphi}(\vz|\vx)} \right] + \mathbb{D}_{\text{KL}}( q_{\vphi}(\vz|\vx) \| p(\vz|\vx)) \label{eq: ELBO with KL}\\
&\ge \E_{q_{\phi}(\vz|\vx)}\left[ \log \frac{p(\vx,\vz)}{q_{\vphi}(\vz|\vx)} \right] \notag \\
&\bydef \text{ELBO}(\vx) \notag,
\end{align}
where the inequality follows from the fact that the KL divergence is always non-negative. Therefore, ELBO is a valid lower bound for $\log p(\vx)$. Since we never have access to $\log p(\vx)$, if we somehow have access to ELBO and if ELBO is a good lower bound, then we can effectively maximize ELBO to achieve the goal of maximizing $\log p(\vx)$ which is the gold standard. Now, the question is how good the lower bound is. As you can see from the equation and also \fref{fig: ELBO gap}, the inequality will become an equality when our proxy $q_{\vphi}(\vz|\vx)$ can match the true distribution $p(\vz|\vx)$ exactly. So, part of the game is to ensure $q_{\vphi}(\vz|\vx)$ is close to $p(\vz|\vx)$.

The derivation of \eref{eq: ELBO with KL} is as follows.

\boxedthm{
\label{thm: VAE log likelihood decomposition}
\textbf{Decomposition of Log-Likelihood}. The log likelihood $\log p(\vx)$ can be decomposed as
\begin{equation}
\log p(\vx)
=
\underset{\bydef \text{ELBO}(\vx)}{\underbrace{\E_{q_{\phi}(\vz|\vx)}\left[ \log \frac{p(\vx,\vz)}{q_{\vphi}(\vz|\vx)} \right]}} + \mathbb{D}_{\text{KL}}( q_{\vphi}(\vz|\vx) \| p(\vz|\vx)).
\end{equation}
}

\boxedproof{
\textbf{Proof}. The trick is to use our magical proxy $q_{\vphi}(\vz|\vx)$ to poke around $p(\vx)$ and derive the bound.
\begin{alignat}{2}
\log p(\vx)
&= \log p(\vx) \times \underset{=1}{\underbrace{\int q_{\vphi}(\vz|\vx) d\vz}}
&&\qquad \text{\textcolor{purple}{(multiply 1)}} \notag\\
&= \int \underset{\text{some constant wrt $\vz$}}{\underbrace{\log p(\vx)}} \times \underset{\text{distribution in $\vz$}}{\underbrace{q_{\vphi}(\vz|\vx)}} d\vz
&&\qquad \text{\textcolor{purple}{(move $\log p(\vx)$ into integral)}}\notag \\
&= \E_{q_{\vphi}(\vz|\vx)}[ \log p(\vx) ],
\end{alignat}
where the last equality is the fact that $\int a \times p_Z(z) dz = \E[a] = a$ for any random variable $Z$ and a scalar $a$.

See, we have already got $\E_{q_{\vphi}(\vz|\vx)}[\cdot]$. Just a few more steps. Let's use Bayes theorem which states that $p(\vx,\vz) = p(\vz|\vx)p(\vx)$:
\begin{alignat}{2}
\E_{q_{\vphi}(\vz|\vx)}[ \log p(\vx) ]
&= \E_{q_{\vphi}(\vz|\vx)}\left[ \log \frac{p(\vx,\vz)}{p(\vz|\vx)} \right]                                                                         &&\qquad \text{\textcolor{purple}{(Bayes Theorem)}} \notag \\
&= \E_{q_{\vphi}(\vz|\vx)}\left[ \log \frac{p(\vx,\vz)}{p(\vz|\vx)} \times \textcolor{blue}{\frac{q_{\vphi}(\vz|\vx)}{q_{\vphi}(\vz|\vx)}}\right]   &&\qquad \text{\textcolor{purple}{(Multiply $\tfrac{q_{\vphi}(\vz|\vx)}{q_{\vphi}(\vz|\vx)}$)}}\notag \\
&=
\underset{\text{ELBO}}{\underbrace{\E_{q_{\vphi}(\vz|\vx)}\left[ \log \frac{p(\vx,\vz)}{\textcolor{blue}{q_{\vphi}(\vz|\vx)}} \right]}} +
\underset{\mathbb{D}_{\text{KL}}(q_{\vphi}(\vz|\vx)\|p(\vz|\vx))}{\underbrace{\E_{q_{\vphi}(\vz|\vx)}\left[ \log \frac{\textcolor{blue}{q_{\vphi}(\vz|\vx)}}{p(\vz|\vx)} \right]}}, \label{eq: ELBO with KL 1}
\end{alignat}
where we recognize that the first term is exactly ELBO, whereas the second term is exactly the KL divergence. Comparing \eref{eq: ELBO with KL 1} with \eref{eq: ELBO with KL}, we complete the proof.
}

\boxedeg{
Using the previous example, we can minimize the gap between $\log p(\vx)$ and $\text{ELBO}(\vx)$ if we \emph{knew} $p(\vz|\vx)$. To see that, we note that $\log p(\vx)$ is
\begin{align*}
\log p(\vx) = \text{ELBO}(\vx) + \mathbb{D}_{\text{KL}}( q_{\vphi}(\vz|\vx) \| p(\vz|\vx)) \ge \text{ELBO}(\vx).
\end{align*}
The equality holds if and only if the KL-divergence term is zero. For the KL divergence to be zero, it is necessary that $q_{\vphi}(\vz|\vx) = p(\vz|\vx)$. However, since $p(\vz|\vx)$ is a delta function, the only possibility is to have
\begin{align}
q_{\vphi}(\vz|\vx)
&= \calN(\vz \;|\; \tfrac{\vx-\vmu}{\sigma}, \;\; 0) \notag \\
&= \delta(\vz - \tfrac{\vx-\vmu}{\sigma}),
\end{align}
i.e., we set the standard deviation to be $t = 0$. To determine $p_{\vtheta}(\vx|\vz)$, we need some additional steps to simplify ELBO.
}

We now have ELBO. But this ELBO is still not too useful because it involves $p(\vx,\vz)$, something we have no access to. So, we need to do a little more work.
\boxedthm{
\textbf{Interpretation of ELBO}. ELBO can be decomposed as
\begin{equation}
\text{ELBO}(\vx) =
\underset{\text{how good your decoder is}}{\underbrace{\E_{q_{\phi}(\vz|\vx)}[ \log
\overset{\text{a Gaussian}}{\overbrace{p_{\vtheta}(\vx|\vz)}}]}} \qquad  - \qquad
\underset{\text{how good your encoder is}}{\underbrace{\mathbb{D}_{\text{KL}}\Big(
\overset{\text{a Gaussian}}{\overbrace{q_{\phi}(\vz|\vx)}}
\,\|\,
\overset{\text{a Gaussian}}{\overbrace{p(\vz)}}
 \Big)}}.
\label{eq: ELBO for VAE}
\end{equation}
}
\boxedproof{
\textbf{Proof}. Let's take a closer look at ELBO
\begin{alignat*}{2}
\text{ELBO}(\vx)
&\bydef \E_{q_{\vphi}(\vz|\vx)}\left[ \log \frac{p(\vx,\vz)}{q_{\vphi}(\vz|\vx)} \right]
&&\qquad \text{\textcolor{purple}{(definition)}}\\
&= \E_{q_{\vphi}(\vz|\vx)}\left[ \log \frac{ \textcolor{blue}{p(\vx|\vz)p(\vz)}}{q_{\vphi}(\vz|\vx)} \right]
&&\qquad \textcolor{purple}{(p(\vx,\vz) = p(\vx|\vz)p(\vz))}\\
&= \E_{q_{\vphi}(\vz|\vx)}\left[ \log p(\vx|\vz) \right] + \E_{q_{\vphi}(\vz|\vx)} \left[ \log \frac{p(\vz)}{q_{\vphi}(\vz|\vx)}\right]
&&\qquad \text{\textcolor{purple}{(split expectation)}}\\
&= \E_{q_{\vphi}(\vz|\vx)}\left[ \log \textcolor{blue}{p_{\vtheta}(\vx|\vz)} \right] - \mathbb{D}_{\text{KL}}(q_{\vphi}(\vz|\vx) \| p(\vz)),          &&\qquad \text{\textcolor{purple}{(definition of KL)}}
\end{alignat*}
where we replaced the inaccessible $p(\vx|\vz)$ by its proxy $p_{\vtheta}(\vx|\vz)$.
}

This is a \emph{beautiful} result. We just showed something very easy to understand. Let's look at the two terms in \eref{eq: ELBO for VAE}:
\begin{itemize}
\setlength\itemsep{-1ex}
\item \textbf{Reconstruction}. The first term is about the \emph{decoder}. We want the decoder to produce a good image $\vx$ if we feed a latent $\vz$ into the decoder (of course!!). So, we want to \emph{maximize} $\log p_{\vtheta}(\vx|\vz)$. It is similar to maximum likelihood where we want to find the model parameter to maximize the likelihood of observing the image. The expectation here is taken with respect to the samples $\vz$ (conditioned on $\vx$). This shouldn't be a surprise because the samples $\vz$ are used to assess the quality of the decoder. It cannot be an arbitrary noise vector but a meaningful latent vector. So, $\vz$ needs to be sampled from $q_{\phi}(\vz|\vx)$.
\item \textbf{Prior Matching}. The second term is the KL divergence for the \emph{encoder}. We want the encoder to turn $\vx$ into a latent vector $\vz$ such that the latent vector will follow our choice of distribution, e.g., $\vz \sim \calN(0,\mI)$. To be slightly more general, we write $p(\vz)$ as the target distribution. Because the KL divergence is a distance (which increases when the two distributions become more dissimilar), we need to put a negative sign in front so that it increases when the two distributions become more similar.
\end{itemize}

\boxedeg{
Following up on the previous example, we continue to assume that we \emph{knew} $p(\vz|\vx)$. Then the reconstruction term in ELBO will give us
\begin{align*}
\E_{q_{\vphi}(\vz|\vx)}[\log p_{\vtheta}(\vx|\vz)]
&= \E_{q_{\vphi}(\vz|\vx)}[\log \calN(\vx \;|\; c\vz+\vv, \; s^2\mI)]\\
&= \E_{q_{\vphi}(\vz|\vx)}\left[ -\frac{1}{2}\log 2\pi - \log s - \frac{\|\vx - (c\vz+\vv)\|^2}{2s^2} \right]\\
&= -\frac{1}{2}\log 2\pi - \log s - \frac{c^2}{2s^2} \E_{q_{\vphi}(\vz|\vx)}\left[ \|\vz - \tfrac{\vx-\vv}{c}\|^2\right]\\
&= -\frac{1}{2}\log 2\pi - \log s - \frac{c^2}{2s^2} \E_{\delta\left(\vz - \tfrac{\vx-\vmu}{\sigma}\right)}\left[ \|\vz - \tfrac{\vx-\vv}{c}\|^2\right]\\
&= -\frac{1}{2}\log 2\pi - \log s - \frac{c^2}{2s^2} \left[ \|\tfrac{\vx-\vmu}{\sigma} - \tfrac{\vx-\vv}{c}\|^2\right]\\
&\le -\frac{1}{2}\log 2\pi - \log s,
\end{align*}
where the upper bound is tight if and only if the norm-square term is zero, which holds when $\vv = \vmu$ and $c = \sigma$. For the remaining terms, it is clear that $-\log s$ is a monotonically decreasing function in $s$ with $-\log s \rightarrow \infty$ as $s \rightarrow 0$. Therefore, when $\vv = \vmu$ and $c = \sigma$, it follows that $\E_{q_{\vphi}(\vz|\vx)}[\log p_{\vtheta}(\vx|\vz)]$ is maximized when $s = 0$. This implies that
\begin{align}
p_{\vtheta}(\vx|\vz)
&= \calN(\vx \;|\; \sigma\vz + \vmu, \; 0) \notag \\
&= \delta(\vx - (\sigma\vz + \vmu)).
\end{align}
}

\textbf{Limitation of ELBO}. ELBO is practically useful, but it is \emph{not} the same as the true likelihood $\log p(\vx)$. As we mentioned, ELBO is exactly equal to $\log p(\vx)$ if and only if $\mathbb{D}_{\text{KL}}(q_{\vphi}(\vz|\vx)\|p(\vz|\vx)) = 0$ which happens when $q_{\vphi}(\vz|\vx) = p(\vz|\vx)$. In the following example, we will show a case where the $q_{\vphi}(\vz|\vx)$ obtained from maximizing ELBO is not the same as $p(\vz|\vx)$.

\boxedeg{
(\textbf{Limitation of ELBO}). In the previous example, if we have no idea about $p(\vz|\vx)$, we need to train the VAE by maximizing ELBO. However, since ELBO is only a lower bound of the true distribution $\log p(\vx)$, maximizing ELBO will not return us the delta functions as we hope. Instead, we will obtain something that is quite meaningful but not exactly the delta functions.

For simplicity, let's consider the distributions that will return us unbiased estimates of the mean but with unknown variances:
\begin{align*}
q_{\vphi}(\vz|\vx)   &= \calN(\vz\;|\; \tfrac{\vx-\vmu}{\sigma}, t^2\mI),\\
p_{\vtheta}(\vx|\vz) &= \calN(\vx\;|\; \sigma\vz+\vmu, s^2\mI).
\end{align*}
This is partially ``cheating'' because in theory we should not assume anything about the estimates of the means. But from an intuitive angle, since $q_{\vphi}(\vz|\vx)$ and $p_{\vtheta}(\vx|\vz)$ are proxies to $p(\vz|\vx)$ and $p(\vx|\vz)$, they must resemble some properties of the delta functions. The closest choice is to define $q_{\vphi}(\vz|\vx)$ and $p_{\vtheta}(\vx|\vz)$ as Gaussians with means consistent with those of the two delta functions. The variances are unknown, and they are the subject of interest in this example.

Our focus here is to maximize ELBO which consists of the prior matching term and the reconstruction term. For the prior matching error, we want to minimize the KL-divergence:
\begin{align*}
\mathbb{D}_{\text{KL}}(q_{\vphi}(\vz|\vx)\|p(\vz)) = \mathbb{D}_{\text{KL}}\left( \calN(\vz\;|\; \tfrac{\vx-\vmu}{\sigma}, t^2\mI) \; \| \; \calN(\vz\;|\; 0, \mI) \right).
\end{align*}
The KL-divergence of two multivariate Gaussians $\calN(\vz|\vmu_0,\mSigma_0)$ and $\calN(\vz|\vmu_1,\mSigma_1)$  has a closed form expression which can be found in Wikipedia:
\begin{align*}
&\mathbb{D}_{\text{KL}}(\calN(\vmu_0,\mSigma_0) \| \calN(\vmu_1,\mSigma_1)) \notag \\
&\qquad\qquad =
\frac{1}{2}\left( \text{Tr}(\mSigma_1^{-1}\mSigma_0) - d + (\vmu_1 - \vmu_0)^T\mSigma_1^{-1}(\vmu_1-\vmu_0) + \log\frac{\text{det}\mSigma_1}{\text{det}\mSigma_0}\right).
\end{align*}
Using this result (and with some algebra), we can show that
\begin{align*}
\mathbb{D}_{\text{KL}}\left( \calN(\vz\;|\; \tfrac{\vx-\vmu}{\sigma}, t^2\mI) \; \| \; \calN(\vz\;|\; 0, \mI) \right)
&= \frac{1}{2}\left[ t^2 d - d + \|\tfrac{\vx-\vmu}{\sigma}\|^2 - 2 d \log t \right],
\end{align*}
where $d$ is the dimension of $\vx$ and $\vz$. To minimize the KL-divergence, we take derivative with respect to $t$ and show that
\begin{align*}
\frac{\partial}{\partial t}\left\{\frac{1}{2}\left[ t^2 d - d + \|\tfrac{\vx-\vmu}{\sigma}\|^2 - 2 d \log t \right]\right\} = t\cdot d - \frac{d}{t}.
\end{align*}
Setting this to zero will give us $t = 1$. Therefore, we can show that
\begin{equation*}
q_{\vphi}(\vz|\vx) = \calN(\vz\;|\; \tfrac{\vx-\vmu}{\sigma}, \mI).
\end{equation*}

For the reconstruction term, we can show that
\begin{align*}
\E_{q_{\vphi}(\vz|\vx)}[\log p_{\vtheta}(\vx|\vz)]
&= \E_{q_{\vphi}(\vz|\vx)}\left[ \log \frac{1}{(\sqrt{2\pi s^2})^d} \exp\left\{-\frac{\|\vx-(\sigma\vz+\vmu)\|^2}{2s^2}\right\} \right]\\
&= \E_{q_{\vphi}(\vz|\vx)}\left[ -\frac{d}{2}\log 2\pi - d \log s - \frac{\|\vx-(\sigma\vz+\vmu)\|^2}{2s^2}\right]\\
&= -\tfrac{d}{2}\log 2\pi - d\log s - \frac{\sigma^2}{2s^2} \E_{q_{\vphi}(\vz|\vx)}\left[\left\|\vz-\tfrac{\vx-\vmu}{\sigma}\right\|^2\right]\\
&= -\tfrac{d}{2}\log 2\pi - d\log s - \frac{\sigma^2}{2s^2} \text{Trace} \left\{\E_{q_{\vphi}(\vz|\vx)}\left[\left(\vz-\tfrac{\vx-\vmu}{\sigma}\right)\left(\vz-\tfrac{\vx-\vmu}{\sigma}\right)^T\right]\right\}\\
&= -\frac{d}{2}\log 2\pi - d\log s - \frac{\sigma^2}{2s^2} \cdot d,
\end{align*}
because the covariance of $\vz \sim q_{\vphi}(\vz|\vx)$ is $\mI$ and so the trace will give us $d$. Taking derivatives with respect to $s$ will give us
\begin{align*}
\frac{d}{ds}\left\{-\tfrac{d}{2}\log 2\pi - d\log s - \frac{d \sigma^2}{2s^2}\right\}
= -\frac{d}{s} + \frac{d \sigma^2}{s^3} = 0.
\end{align*}
Equating this to zero will give us $s = \sigma$. Therefore,
\begin{equation*}
p_{\vtheta}(\vx|\vz) = \calN(\vx\;|\; \sigma \vz + \vmu, \sigma^2\mI).
\end{equation*}

As we can see in this example and the previous example, while the ideal distributions are delta functions, the proxy distributions we obtain have a finite variance. This finite variance adds additional randomness to the samples generated by the VAE. There is nothing wrong with this VAE --- we do it correctly by maximizing ELBO. It is just that maximizing the ELBO is not the same as maximizing $\log p(\vx)$.
}

\subsection{Optimization in VAE}
In the previous two subsections we introduced the building blocks of VAE and ELBO. The goal of this subsection is to discuss how to train a VAE and how to do inference.

VAE is a model that aims to approximate the true distribution $p(\vx)$ so that we can draw samples. A VAE is parameterized by $(\vphi, \vtheta)$. Therefore, training a VAE is equivalent to solving an optimization problem that encapsulates the essence of $p(\vx)$ while being tractable. However, since $p(\vx)$ is not accessible, the natural alternative is to optimize the ELBO which is the lower bound of $\log p(\vx)$. That means, the learning goal of VAE is to solve the following problem.
\boxeddef{
The optimization objective of VAE is to maximize the ELBO:
\begin{equation}
(\vphi, \vtheta) = \argmax{\vphi, \vtheta} \;\;\;  \sum_{\vx \in \calX} \text{ELBO}(\vx),
\end{equation}
where $\calX = \{\vx^{(\ell)} \;|\; \ell = 1,\ldots,L\}$ is the training dataset.
}

\textbf{Intractability of ELBO's Gradient}. The challenge associated with the above optimization is that the gradient of ELBO with respect to $(\vphi,\vtheta)$ is intractable. Since the majority of today's neural network optimizers use first-order methods and backpropagate the gradient to update the network weights, an intractable gradient will pose difficulties in training the VAE.

Let's elaborate more about the intractability of the gradient. We first substitute Definition~\ref{def: ELBO} into the above objective function. The gradient of ELBO is: \footnote{The original definition of ELBO uses the true joint distribution $p(\vx,\vz)$. In practice, since $p(\vx,\vz)$ is not accessible, we replace it by its proxy $p_{\vtheta}(\vx,\vz)$ which is a computable distribution.}
\begin{align}
\nabla_{\vtheta,\vphi} \;\; \text{ELBO}(\vx)
&= \nabla_{\vtheta,\vphi} \left\{\E_{q_{\phi}(\vz|\vx)}\left[ \log \frac{p_{\vtheta}(\vx,\vz)}{q_{\vphi}(\vz|\vx)} \right]\right\} \notag \\
&= \nabla_{\vtheta,\vphi} \Big\{\E_{q_{\phi}(\vz|\vx)}\Big[ \log p_{\vtheta}(\vx,\vz) - \log q_{\vphi}(\vz|\vx) \Big]  \Big\}.
\end{align}

The gradient contains two parameters. Let's first look at $\vtheta$. We can show that
\begin{alignat}{2}
\nabla_{\vtheta} \;\; \text{ELBO}(\vx)
&= \nabla_{\vtheta} \Big\{\E_{q_{\phi}(\vz|\vx)}\Big[ \log p_{\vtheta}(\vx,\vz) - \log q_{\vphi}(\vz|\vx) \Big]  \Big\} \notag \\
&= \nabla_{\vtheta} \left\{\int \Big[ \log p_{\vtheta}(\vx,\vz) - \log q_{\vphi}(\vz|\vx) \Big]  \cdot q_{\phi}(\vz|\vx) d\vz\right\} \notag \\
&= \int \nabla_{\vtheta} \Big\{ \log p_{\vtheta}(\vx,\vz) - \log q_{\vphi}(\vz|\vx) \Big\}  \cdot q_{\phi}(\vz|\vx) \; d\vz \notag \\
&= \E_{q_{\phi}(\vz|\vx)} \Big[ \nabla_{\vtheta}\Big\{ \log p_{\vtheta}(\vx,\vz) - \log q_{\vphi}(\vz|\vx) \Big\} \Big] \notag \\
&= \E_{q_{\phi}(\vz|\vx)} \Big[ \nabla_{\vtheta}\Big\{ \log p_{\vtheta}(\vx,\vz) \Big\} \Big] \notag \\
&\approx \frac{1}{L}\sum_{\ell=1}^L \nabla_{\vtheta}\Big\{ \log p_{\vtheta}(\vx,\vz^{(\ell)}) \Big\},
&&\qquad \textcolor{purple}{(\text{where}\quad \vz^{(\ell)} \sim q_{\vphi}(\vz|\vx))}
\end{alignat}
where the last equality is the Monte Carlo approximation of the expectation.

In the above equation, if $p_{\vtheta}(\vx,\vz)$ is realized by a computable model such as a neural network, then its gradient $\nabla_{\vtheta}\{ \log p_{\vtheta}(\vx,\vz)\}$ can be computed via automatic differentiation. Thus, the maximization can be achieved by backpropagating the gradient.

The gradient with respect to $\vphi$ is more difficult. We can show that
\begin{alignat}{2}
\nabla_{\vphi} \;\; \text{ELBO}(\vx)
&= \nabla_{\vphi} \Big\{\E_{q_{\phi}(\vz|\vx)}\Big[ \log p_{\vtheta}(\vx,\vz) - \log q_{\vphi}(\vz|\vx) \Big]  \Big\}  \notag \\
&= \nabla_{\vphi} \left\{\int \Big[ \log p_{\vtheta}(\vx,\vz) - \log q_{\vphi}(\vz|\vx) \Big]  \cdot q_{\phi}(\vz|\vx) d\vz\right\}  \notag \\
&= \int \nabla_{\vphi} \Big\{ [\log p_{\vtheta}(\vx,\vz) - \log q_{\vphi}(\vz|\vx)]  \cdot q_{\phi}(\vz|\vx) \Big\}  \; d\vz \notag \\
&\not= \int \nabla_{\vphi} \Big\{ \log p_{\vtheta}(\vx,\vz) - \log q_{\vphi}(\vz|\vx)  \Big\} \cdot q_{\phi}(\vz|\vx) \; d\vz \notag \\
&= \E_{q_{\phi}(\vz|\vx)} \Big[ \nabla_{\vphi}\Big\{ \log p_{\vtheta}(\vx,\vz) - \log q_{\vphi}(\vz|\vx) \Big\} \Big] \notag \\
&= \E_{q_{\phi}(\vz|\vx)} \Big[ \nabla_{\vphi}\Big\{ - \log q_{\vphi}(\vz|\vx) \Big\} \Big] \notag \\
&\approx \frac{1}{L}\sum_{\ell=1}^L \nabla_{\vphi}\Big\{ - \log q_{\vphi}(\vz^{(\ell)}|\vx) \Big\},
&&\qquad\textcolor{purple}{(\text{where}\quad \vz^{(\ell)} \sim q_{\vphi}(\vz|\vx))}.
\end{alignat}
As we can see, even though we \emph{wish} to maintain a similar structure as we did for $\vtheta$, the expectation and the gradient operators in the above derivations cannot be switched. This forbids us from doing any backpropagation of the gradient to maximize ELBO.

\vspace{2ex}
\textbf{Reparameterization Trick}. The intractability of ELBO's gradient is inherited from the fact that we need to draw samples $\vz$ from a distribution $q_{\vphi}(\vz|\vx)$ which itself is a function of $\vphi$. As noted by Kingma and Welling \cite{Kingma_2014_ICLR}, for continuous latent variables, it is possible to compute an unbiased estimate of $\nabla_{\vtheta,\vphi} \;\; \text{ELBO}(\vx)$ so that we can approximately calculate the gradient and hence maximize ELBO. The idea is to employ an technique known as the \emph{reparameterization trick} \cite{Kingma_2014_ICLR}.

Recall that the latent variable $\vz$ is a sample drawn from the distribution $q_{\vphi}(\vz|\vx)$. The idea of reparameterization trick is to express $\vz$ as some differentiable and invertible transformation of another random variable $\vepsilon$ whose distribution is independent of $\vx$ and $\vphi$. That is, we define a differentiable and invertible function $\vg$ such that
\begin{equation}
\vz = \vg(\vepsilon, \vphi, \vx),
\end{equation}
for some random variable $\vepsilon \sim p(\vepsilon)$. To make our discussions easier, we pose an additional requirement that
\begin{equation}
q_{\vphi}(\vz|\vx) \cdot \left|\det \left(\frac{\partial \vz}{\partial \vepsilon}\right)\right| = p(\vepsilon),
\label{eq: VAE change of variable Jacobian}
\end{equation}
where $\frac{\partial \vz}{\partial \vepsilon}$ is the Jacobian, and $\text{det}(\cdot)$ is the matrix determinant. This requirement is related to change of variables in multivariate calculus. The following example will make it clear.

\boxedeg{
Suppose $\vz \sim q_{\vphi}(\vz|\vx) \bydef \calN(\vz \;|\; \vmu, \text{diag}(\vsigma^2))$. We can define
\begin{equation}
\vz = \vg(\vepsilon,\vphi,\vx) \bydef \vepsilon \odot \vsigma + \vmu,
\end{equation}
where $\vepsilon \sim \calN(0,\mI)$ and ``$\odot$'' means elementwise multiplication. The parameter $\vphi$ is $\vphi = (\vmu,\vsigma^2)$. For this choice of the distribution, we can show that by letting $\vepsilon = \frac{\vz-\vmu}{\vsigma}$:
\begin{align*}
q_{\vphi}(\vz|\vx) \cdot \left|\det \left(\frac{\partial \vz}{\partial \vepsilon}\right)\right|
&= \prod_{i=1}^d \frac{1}{\sqrt{2\pi \sigma_i^2}} \exp\left\{-\frac{(z_i-\mu_i)^2}{2\sigma_i^2}\right\} \cdot \prod_{i=1}^d \sigma_i\\
&= \frac{1}{(\sqrt{2\pi})^d} \exp\left\{-\frac{\|\vepsilon\|^2}{2}\right\} = \calN(0,\mI) = p(\vepsilon).
\end{align*}
}

With this re-parameterization of $\vz$ by expressing it in terms of $\vepsilon$, we can look at $\nabla_{\vphi} \E_{q_{\vphi}(\vz|\vx)}[f(\vz)]$ for some general function $f(\vz)$. (Later we will consider $f(\vz) = -\log q_{\vphi}(\vz|\vx)$.) For notational simplicity, we write $\vg(\vepsilon)$ instead of $\vg(\vepsilon,\vphi,\vx)$ although we understand that $\vg$ has three inputs. By change of variables, we can show that
\begin{alignat}{2}
\E_{q_{\vphi}(\vz|\vx)}[f(\vz)]
&= \int f(\vz) \cdot q_{\vphi}(\vz|\vx) \; d\vz & & \notag \\
&= \int f(\vg(\vepsilon)) \cdot q_{\vphi}(\vg(\vepsilon)|\vx) \; d \vg(\vepsilon), & & \qquad \textcolor{purple}{(\vz = \vg(\vepsilon))} \notag\\
&= \int f \Big( \vg(\vepsilon) \Big) \cdot
q_{\vphi}(\vg(\vepsilon)|\vx) \cdot \left|\det \left(\frac{\partial \vg(\vepsilon)}{\partial \vepsilon}\right)\right| \; d\vepsilon & &  \qquad \text{\textcolor{purple}{(Jacobian due to change of variable)}}\notag\\
&= \int f (\vz) \cdot p(\vepsilon) \; d\vepsilon & & \qquad \text{\textcolor{purple}{(use \eref{eq: VAE change of variable Jacobian})}}\notag \\
&= \E_{p(\vepsilon)} \left[ f (\vz)  \right].
\end{alignat}
So, if we want to take the gradient with respect to $\vphi$, we can show that
\begin{align}
\nabla_{\vphi} \E_{q_{\vphi}(\vz|\vx)}[f(\vz)]
= \nabla_{\vphi}  \E_{p(\vepsilon)} \left[ f (\vz)  \right]
&= \nabla_{\vphi} \left\{ \int f (\vz) \cdot p(\vepsilon) \; d\vepsilon \right\}\notag\\
&= \int \nabla_{\vphi} \left\{ f (\vz) \cdot p(\vepsilon) \right\} \; d\vepsilon\notag\\
&= \int \left\{ \nabla_{\vphi}  f (\vz) \right\} \cdot p(\vepsilon) \; d\vepsilon \notag\\
&= \E_{p(\vepsilon)} \left[ \nabla_{\vphi} f (\vz) \right],
\end{align}
which can be approximated by Monte Carlo. Substituting $f(\vz) = -\log q_{\vphi}(\vz|\vx)$, we can show that
\begin{align*}
\nabla_{\vphi} \E_{q_{\vphi}(\vz|\vx)}[-\log q_{\vphi}(\vz|\vx)]
&= \E_{p(\vepsilon)} \left[ -\nabla_{\vphi} \log q_{\vphi}(\vz|\vx) \right]\\
&\approx - \frac{1}{L}\sum_{\ell=1}^L \nabla_{\vphi} \log q_{\vphi}(\vz^{(\ell)}|\vx), \qquad \textcolor{purple}{(\text{where}\quad \vz^{(\ell)} = \vg(\vepsilon^{(\ell)},\vphi,\vx))}\\
&= -\frac{1}{L}\sum_{\ell=1}^L \nabla_{\vphi}\left[ \log p(\vepsilon^{(\ell)}) - \log \left|\det \frac{\partial \vz^{(\ell)}}{\partial \vepsilon^{(\ell)}}\right| \right]\\
&= \frac{1}{L} \sum_{\ell=1}^L \nabla_{\vphi}\left[ \log \left|\det \frac{\partial \vz^{(\ell)}}{\partial \vepsilon^{(\ell)}}\right| \right].
\end{align*}
So, as long as the determinant is differentiable with respect to $\vphi$, the Monte Carlo approximation can be numerically computed.

\boxedeg{
Suppose that the parameters and the distribution $q_{\vphi}$ are defined as follows:
\begin{align*}
(\vmu,\vsigma^2) &= \text{EncoderNetwork}_{\vphi}(\vx)\\
q_{\vphi}(\vz|\vx) &= \calN(\vz \;|\; \vmu, \text{diag}(\vsigma^2)).
\end{align*}
We can define $\vz = \vmu + \vsigma \odot \vepsilon$, with $\vepsilon \sim \calN(0,\mI)$. Then, we can show that
\begin{align*}
\log \left|\det \frac{\partial \vz}{\partial \vepsilon}\right|
&= \log \left| \text{det} \left(\frac{\partial (\vmu + \vsigma\odot\vepsilon)}{\partial \vepsilon}\right)\right|\\
&= \log \left| \text{det} \Big( \diag{\vsigma} \Big) \right| \\
&= \log \prod_{i=1}^d \sigma_i = \sum_{i=1}^d \log \sigma_i.
\end{align*}
Therefore, we can show that
\begin{align*}
\nabla_{\vphi} \E_{q_{\vphi}(\vz|\vx)}[-\log q_{\vphi}(\vz|\vx)]
&\approx \frac{1}{L} \sum_{\ell=1}^L \nabla_{\vphi}\left[ \log \left|\det \frac{\partial \vz^{(\ell)}}{\partial \vepsilon^{(\ell)}}\right| \right]\\
&= \frac{1}{L} \sum_{\ell=1}^L \nabla_{\vphi}\left[ \sum_{i=1}^d \log  \sigma_i \right] \\
&= \nabla_{\vphi}\left[ \sum_{i=1}^d \log  \sigma_i \right] \\
&= \frac{1}{\vsigma} \odot \nabla_{\vphi}\Big\{\vsigma_{\vphi}(\vx)\Big\},
\end{align*}
where we emphasize that $\vsigma_{\vphi}(\vx)$ is the output of the encoder which is a neural network.
}
As we can see in the above example, for some specific choices of the distributions (e.g., Gaussian), the gradient of ELBO can be significantly easier to derive.

\vspace{2ex}
\textbf{VAE Encoder}. After discussing the reparameterizing trick, we can now discuss the specific structure of the encoder in VAE. To make our discussions focused, we assume a relatively common choice of the encoder:
\begin{align*}
(\vmu,\sigma^2) &= \text{EncoderNetwork}_{\vphi}(\vx)\\
q_{\vphi}(\vz|\vx) &= \calN(\vz \;|\; \vmu, \sigma^2\mI).
\end{align*}
The parameters $\vmu$ and $\sigma$ are technically \emph{neural networks} because they are the outputs of $\text{EncoderNetwork}_{\vphi}(\cdot)$. Therefore, it will be helpful if we denote them as
\begin{align*}
\vmu      &= \underset{\text{neural network}}{\underbrace{\vmu_{\vphi}}}(\vx),\\
\sigma^2   &= \underset{\text{neural network}}{\underbrace{\sigma_{\vphi}^2}}(\vx),
\end{align*}
Our notation is slightly more complicated because we want to emphasize that $\vmu$ is a function of $\vx$; You give us an image $\vx$, our job is to return you the parameters of the Gaussian (i.e., mean and variance). If you give us a different $\vx$, then the parameters of the Gaussian should also be different. The parameter $\vphi$ specifies that $\vmu$ is controlled (or parameterized) by $\vphi$.

\begin{figure}[h]
\centering
\includegraphics[width=0.6\linewidth]{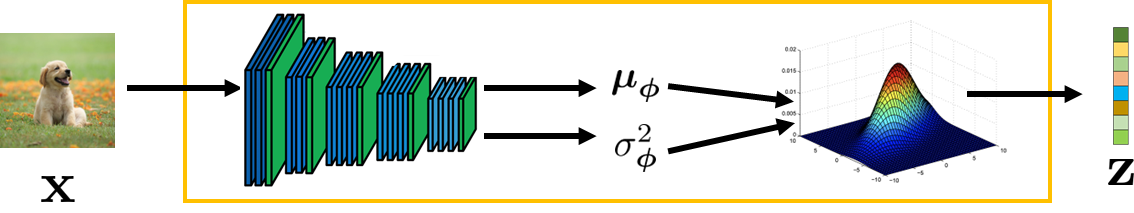}
\caption{Implementation of a VAE encoder. We use a neural network to take the image $\vx$ and estimate the mean $\vmu_{\vphi}$ and variance $\sigma^2_{\vphi}$ of the Gaussian distribution.}
\label{fig: VAE encoder}
\end{figure}

Suppose that we are given the $\ell$-th training sample $\vx^{(\ell)}$. From this $\vx^{(\ell)}$ we want to generate a latent variable $\vz^{(\ell)}$ which is a sample from $q_{\vphi}(\vz|\vx)$. Because of the Gaussian structure, it is equivalent to say that
\begin{align}
\vz^{(\ell)} \sim \calN\Big( \vz \;\Big| \; \vmu_{\vphi}(\vx^{(\ell)}), \quad \sigma_{\vphi}^2(\vx^{(\ell)})\mI\Big). \label{eq: VAE sampling z}
\end{align}
The interesting thing about this equation is that we use a neural network $\text{EncoderNetwork}_{\vphi}(\cdot)$ to estimate the mean and variance of the Gaussian. Then, from this Gaussian we draw a sample $\vz^{(\ell)}$, as illustrated in \fref{fig: VAE encoder}.

A more convenient way of expressing \eref{eq: VAE sampling z} is to realize that the sampling operation $\vz \sim \calN(\vmu,\sigma^2\mI)$ can be done using the reparameterization trick.
\boxedmsg{
\textbf{Reparameterization Trick for High-dimensional Gaussian}:
\begin{equation}
\vz \sim \calN(\vmu,\sigma^2\mI) \qquad \Longleftrightarrow \qquad \vz = \vmu + \sigma \vepsilon, \quad \vepsilon \sim \calN(0,\mI).
\end{equation}
}

Using the reparameterization trick, \eref{eq: VAE sampling z}  can be written as
\begin{equation*}
\vz^{(\ell)} = \vmu_{\vphi}(\vx^{(\ell)}) + \sigma_{\vphi}(\vx^{(\ell)}) \vepsilon, \qquad \vepsilon \sim \calN(0,\mI).
\end{equation*}

\boxedproof{
\textbf{Proof}. We will prove a general case for an arbitrary covariance matrix $\mSigma$ instead of a diagonal matrix $\sigma^2\mI$.

For any high-dimensional Gaussian $\vz \sim \calN(\vz|\vmu,\mSigma)$, the sampling process can be done via the transformation of white noise
\begin{equation}
\vz = \vmu + \mSigma^{\frac{1}{2}}\vepsilon,
\end{equation}
where $\vepsilon \sim \calN(0,\mI)$. The half matrix $\mSigma^{\frac{1}{2}}$ can be obtained through eigen-decomposition or Cholesky factorization. If $\mSigma$ has an eigen-decomposition $\mSigma = \mU\mS\mU^T$, then $\mSigma^{\frac{1}{2}} = \mU\mS^{\frac{1}{2}}\mU^T$. The square root of the eigenvalue matrix $\mS$ is well-defined because $\mSigma$ is a positive semi-definite matrix.

We can calculate the expectation and covariance of $\vx$:
\begin{align*}
\E[\vz]         &= \E[\vmu + \mSigma^{\frac{1}{2}}\vepsilon] = \vmu + \mSigma^{\frac{1}{2}} \underset{=0}{\underbrace{\E[\vepsilon]}} = \vmu,\\
\text{Cov}(\vz) &= \E[(\vz-\vmu)(\vz-\vmu)^T] = \E\left[\mSigma^{\frac{1}{2}} \vepsilon \vepsilon^T (\mSigma^{\frac{1}{2}})^T\right]
                = \mSigma^{\frac{1}{2}} \underset{=\mI}{\underbrace{\E[\vepsilon\vepsilon^T]}} (\mSigma^{\frac{1}{2}})^T = \mSigma.
\end{align*}
Therefore, for diagonal matrices $\mSigma = \sigma^2\mI$, the above is reduced to
\begin{equation}
\vz = \vmu + \sigma\vepsilon, \qquad \text{where} \; \vepsilon \sim \calN(0,\mI).
\end{equation}
}

Given the VAE encoder structure and $q_{\vphi}(\vz|\vx)$, we can go back to ELBO. Recall that ELBO consists of the prior matching term and the reconstruction term. The prior matching term is measured in terms of the KL divergence $\mathbb{D}_{\text{KL}}\left( q_{\vphi}(\vz|\vx) \| p(\vz) \right)$. Let's evaluate this KL divergence.

To evaluate the KL divergence, we (re)use a result which we summarize below:
\boxedthm{
\textbf{KL-Divergence of Two Gaussian}.

The KL divergence for two $d$-dimensional Gaussian distributions $\calN(\vmu_0,\mSigma_0)$ and $\calN(\vmu_1,\mSigma_1)$ is
\begin{align}
&\mathbb{D}_{\text{KL}}\Big(\calN(\vmu_0,\mSigma_0) \; \| \; \calN(\vmu_1,\mSigma_1)\Big) \notag \\
&\qquad\qquad =
\frac{1}{2}\left( \text{Tr}(\mSigma_1^{-1}\mSigma_0) - d + (\vmu_1 - \vmu_0)^T\mSigma_1^{-1}(\vmu_1-\vmu_0) + \log\frac{\text{det}\mSigma_1}{\text{det}\mSigma_0}\right).
\end{align}
}
Substituting our distributions by considering
\begin{alignat*}{2}
\vmu_0 &= \vmu_{\vphi}(\vx),  &\quad& \mSigma_0 = \sigma^2_{\vphi}(\vx)\mI\\
\vmu_1 &= 0,                  &\quad& \mSigma_1 = \mI,
\end{alignat*}
we can show that the KL divergence has an analytic expression
\begin{equation}
\mathbb{D}_{\text{KL}}\Big( q_{\vphi}(\vz|\vx) \; \| \;  p(\vz) \Big) = \frac{1}{2} \Big( \sigma_{\vphi}^{2}(\vx)d - d + \|\vmu_{\vphi}(\vx)\|^2 - 2 d \log \sigma_{\vphi}(\vx)\Big),
\end{equation}
where $d$ is the dimension of the vector $\vz$. The gradient of the KL-divergence with respect to $\vphi$ does not have a closed form but they can be calculated numerically:
\begin{equation}
\nabla_{\vphi} \mathbb{D}_{\text{KL}}\Big( q_{\vphi}(\vz|\vx) \; \| \;  p(\vz) \Big) = \frac{1}{2} \nabla_{\vphi} \Big( \sigma_{\vphi}^{2}(\vx)d - d + \|\vmu_{\vphi}(\vx)\|^2 - 2 d \log \sigma_{\vphi}(\vx)\Big).
\end{equation}
The gradient with respect to $\vtheta$ is zero because there is nothing dependent on $\vtheta$.

\vspace{2ex}
\textbf{VAE Decoder}. The decoder is implemented through a neural network. For notation simplicity, let's define it as $\text{DecoderNetwork}_{\vtheta}(\cdot)$ where $\vtheta$ denotes the network parameters. The job of the decoder network is to take a latent variable $\vz$ and generate an image $f_{\vtheta}(\vz)$:
\begin{equation}
f_{\vtheta}(\vz) = \text{DecoderNetwork}_{\vtheta}(\vz).
\end{equation}
The distribution $p_{\vtheta}(\vx|\vz)$ can be defined as
\begin{equation}
p_{\vtheta}(\vx|\vz) = \calN(\vx \;|\; f_{\vtheta}(\vz), \sigma_{\text{dec}}^2 \mI), \qquad \text{for some hyperparameter $\sigma_{\text{dec}}$.}
\end{equation}
The interpretation of $p_{\vtheta}(\vx|\vz)$ is that we estimate $f_{\vtheta}(\vz)$ through a network and put it as the mean of the Gaussian. If we draw a sample $\vx$ from $p_{\vtheta}(\vx|\vz)$, then by the reparameterization trick we can write the generated image $\widehat{\vx}$ as
\begin{equation*}
\widehat{\vx} = f_{\vtheta}(\vz) + \sigma_{\text{dec}}\vepsilon, \qquad \vepsilon \sim \calN(0,\mI).
\end{equation*}
Moreover, if we take the log of the likelihood, we can show that
\begin{align}
\log p_{\vtheta}(\vx|\vz)
&= \log \calN(\vx \,|\,  f_{\vtheta}(\vz), \sigma_{\text{dec}}^2\mI) \notag \\
&= \log \frac{1}{\sqrt{(2\pi \sigma_{\text{dec}}^2)^d}} \exp\left\{-\frac{\|\vx - f_{\vtheta}(\vz)\|^2}{2\sigma_{\text{dec}}^2}\right\} \notag \\
&= -\frac{\|\vx - f_{\vtheta}(\vz)\|^2}{2\sigma_{\text{dec}}^2} \;\; - \;\;
\underset{\text{independent of $\vtheta$ so we can drop it}}{\underbrace{\log \sqrt{(2\pi \sigma_{\text{dec}}^2)^d}}}.
\label{eq: VAE decoder log p}
\end{align}

Going back to ELBO, we want to compute $\E_{q_{\vphi}(\vz|\vx)}[\log p_{\vtheta}(\vx|\vz)]$. If we straightly calculate the expectation, we will need to compute an integration
\begin{align*}
\E_{q_{\vphi}(\vz|\vx)}[\log p_{\vtheta}(\vx|\vz)]
&= \int \log \left[ \calN(\vx \;|\; f_{\vtheta}(\vz), \sigma_{\text{dec}}^2 \mI)\right] \cdot \calN(\vz \;|\; \vmu_{\vphi}(\vx), \sigma_{\vphi}^2(\vx)) d\vz\\
&= - \int \frac{\|\vx - f_{\vtheta}(\vz)\|^2}{2\sigma_{\text{dec}}^2} \cdot \calN\Big(\vz \;\Big|\; \vmu_{\vphi}(\vx), \sigma_{\vphi}^2(\vx)\Big) d\vz + C,
\end{align*}
where the constant $C$ coming out of the log of the Gaussian can be dropped. By using the reparameterization trick, we write $\vz = \vmu_{\vphi}(\vx) + \sigma_{\vphi}(\vx)\vepsilon$ and substitute it into the above equation. This will give us\footnote{The negative sign here is not a mistake. We want to \emph{maximize} $\E_{q_{\vphi}(\vz|\vx)}[\log p_{\vtheta}(\vx|\vz)]$, which is equivalent to \emph{minimize} the negative of the $\ell_2$ norm.}
\begin{align}
\E_{q_{\vphi}(\vz|\vx)}[\log p_{\vtheta}(\vx|\vz)]
&= - \int \frac{\|\vx - f_{\vtheta}(\vz)\|^2}{2\sigma_{\text{dec}}^2} \cdot \calN\Big(\vz \;\Big|\; \vmu_{\vphi}(\vx), \sigma_{\vphi}^2(\vx)\Big) d\vz\notag \\
&\approx -\frac{1}{M}\sum_{m=1}^M \frac{\|\vx - f_{\vtheta}(\vz^{(m)})\|^2}{2\sigma_{\text{dec}}^2}\label{eq: E_q(log p) L2 norm}\\
&= -\frac{1}{M}\sum_{m=1}^M \frac{\|\vx - f_{\vtheta}\Big(\vmu_{\vphi}(\vx) + \sigma_{\vphi}(\vx)\vepsilon^{(m)} \Big)\|^2}{2\sigma_{\text{dec}}^2}.
\notag
\end{align}
The approximation above is due to Monte Carlo where the randomness is based on the sampling of the $\vepsilon \sim \calN(\vepsilon \;|\; 0, \mI)$. The index $M$ specifies the number of Monte Carlo samples we want to use to approximate the expectation. Note that the input image $\vx$ is fixed because $\E_{q_{\vphi}(\vz|\vx)}[\log p_{\vtheta}(\vx|\vz)]$ is a function of $\vx$.

The gradient of $\E_{q_{\vphi}(\vz|\vx)}[\log p_{\vtheta}(\vx|\vz)]$ with respect to $\vtheta$ is relatively easy to compute. Since only $f_{\vtheta}$ depends on $\vtheta$, we can do automatic differentiation. The gradient with respect to $\vphi$ is slightly harder, but it is still computable because we use chain rule and go into $\vmu_{\vphi}(\vx)$ and $\vphi_{\vphi}(\vx)$.

\begin{figure}[h]
\centering
\includegraphics[width=0.7\linewidth]{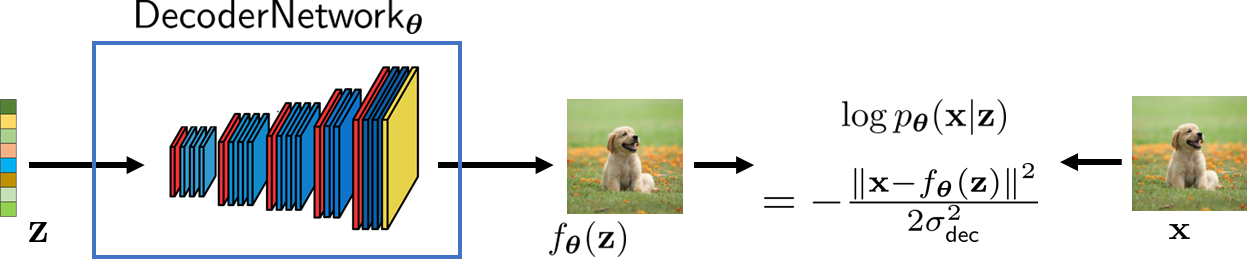}
\caption{Implementation of a VAE decoder. We use a neural network to take the latent vector $\vz$ and generate an image $f_{\vtheta}(\vz)$. The log likelihood will give us a quadratic equation if we assume a Gaussian distribution.}
\label{fig: VAE decoder}
\end{figure}

Inspecting \eref{eq: E_q(log p) L2 norm}, we notice one interesting thing that the loss function is simply the $\ell_2$ norm between the reconstructed image $f_{\vtheta}(\vz)$ and the ground truth image $\vx$. This means that if we have the generated image $f_{\vtheta}(\vz)$, we can do a direct comparison with the ground truth $\vx$ via the usual $\ell_2$ loss as illustrated in \fref{fig: VAE decoder}.

\vspace{2ex}
\textbf{Training the VAE}. Given a training dataset $\calX = \{(\vx^{(\ell)})\}_{\ell=1}^L$ of clean images, the training objective of VAE is to maximize the ELBO
\begin{align*}
\argmax{\vtheta,\vphi} \;\; \sum_{\vx \in \calX} \text{ELBO}_{\vphi,\vtheta}(\vx),
\end{align*}
where the summation is taken with respect to the entire training dataset. The individual ELBO is based on the sum of the terms we derived above
\begin{equation}
\text{ELBO}_{\vphi,\vtheta}(\vx) = \E_{q_{\vphi}(\vz|\vx)}[\log p_{\vtheta}(\vx|\vz)] - \mathbb{D}_{\text{KL}}\Big( q_{\vphi}(\vz|\vx) \; \| \;  p(\vz) \Big).
\end{equation}
Here, the reconstruction term is:
\begin{align}
\E_{q_{\vphi}(\vz|\vx)}[\log p_{\vtheta}(\vx|\vz)] &\approx -\frac{1}{M}\sum_{m=1}^M \frac{\|\vx - f_{\vtheta}\Big(\vmu_{\vphi}(\vx) + \sigma_{\vphi}(\vx)\vepsilon^{(m)} \Big)\|^2}{2\sigma_{\text{dec}}^2},
\end{align}
whereas the prior matching term is
\begin{align}
\mathbb{D}_{\text{KL}}\Big( q_{\vphi}(\vz|\vx) \; \| \;  p(\vz) \Big) = \frac{1}{2} \Big( \sigma_{\vphi}^{2}(\vx)d - d + \|\vmu_{\vphi}(\vx)\|^2 - 2 \log \sigma_{\vphi}(\vx)\Big).
\end{align}
To optimize for $\vtheta$ and $\vphi$, we can run stochastic gradient descent. The gradients can be taken based on the tensor graphs of the neural networks. On computers, this is done automatically by the automatic differentiation.

Let's summarize these.
\boxedthm{
(\textbf{VAE Training}). To train a VAE, we need to solve the optimization problem
\begin{align*}
\argmax{\vtheta,\vphi} \;\;  \sum_{\vx \in \calX} \text{ELBO}_{\vphi,\vtheta}(\vx),
\end{align*}
where
\begin{align}
\text{ELBO}_{\vphi,\vtheta}(\vx)
&= -\frac{1}{M}\sum_{m=1}^M \frac{\|\vx - f_{\vtheta}\Big(\vmu_{\vphi}(\vx) + \sigma_{\vphi}(\vx)\vepsilon^{(m)} \Big)\|^2}{2\sigma_{\text{dec}}^2} \notag \\
&\qquad\qquad + \frac{1}{2} \Big( \sigma_{\vphi}^{2}(\vx)d - d + \|\vmu_{\vphi}(\vx)\|^2 - 2 d \log \sigma_{\vphi}(\vx)\Big).
\end{align}
}

\begin{figure}[h]
\centering
\includegraphics[width=\linewidth]{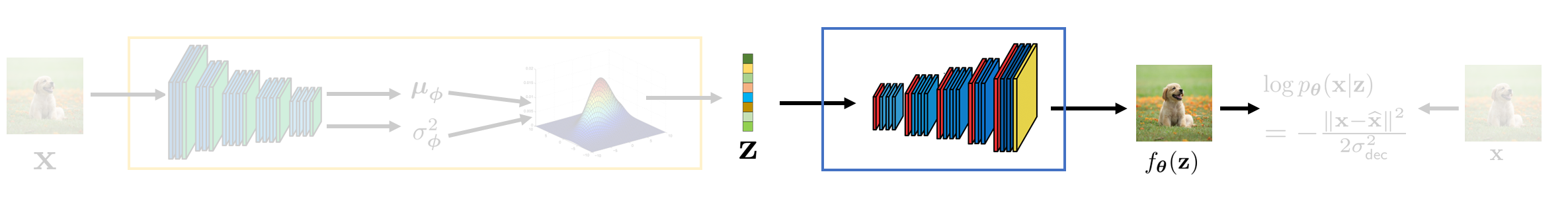}
\vspace{-2ex}
\caption{Using VAE to generate image is as simple as sending a latent noise code $\vz$ through the decoder.}
\label{fig: VAE inference}
\end{figure}

\vspace{2ex}
\textbf{VAE Inference}. The inference of an VAE is relatively simple. Once the VAE is trained, we can drop the encoder and only keep the decoder, as shown in \fref{fig: VAE inference}. To generate a new image from the model, we pick a random latent vector $\vz \in \R^d$. By sending this $\vz$ through the decoder $f_{\vtheta}$, we will be able to generate a new image $\widehat{\vx} = f_{\vtheta}(\vz)$.

\subsection{Concluding Remark}
For readers who are looking for additional references, we highly recommend the tutorial by Kingma and Welling \cite{Kingma_2019_VAE} which is based on their original VAE paper \cite{Kingma_2014_ICLR}. A shorter tutorial by Doersch et al \cite{Doersch_2016_VAE} can also be helpful. \cite{Kingma_2019_VAE} includes a long list of good papers including a paper by Rezende and Mohamed \cite{Rezende_2015_normalizingflow} on normalizing flow which was published around the same time as Kingma and Welling's VAE paper.

VAE has many linkages to the classical variational inference and graphical models \cite{Wainwright_2008_tutorial}. VAE is also relevant to the generative adversarial networks (GAN) by Goodfellow et al. \cite{Goodfellow_2014_GAN}. Kingma and Welling commented in \cite{Kingma_2019_VAE} that VAE and GAN have complementary properties; while GAN produces better perceptual quality images, there is a weaker linkage with the data likelihood. VAE can meet the data likelihood criterion better but the samples are at times not perceptually as good.

\newpage
\section{Denoising Diffusion Probabilistic Model (DDPM)}
\setcounter{figure}{0}
\setcounter{equation}{0}

In this section, we discuss the diffusion models. There are many different perspectives on how the diffusion models can be derived, e.g., score matching, differential equation, etc. We will follow the approach outlined by the original paper on denoising diffusion probability model by Ho et al. \cite{Ho_2020_DDPM}.

Before we discuss the mathematical details, let's summarize DDPM from the perspective of VAE's extension:
\boxedproof{
\begin{center}
Diffusion models are \emph{incremental} updates where the assembly of the whole gives us the encoder-decoder structure.
\end{center}
}

Why increment? It's like turning the direction of a giant ship. You need to turn the ship slowly towards your desired direction or otherwise you will lose control. The same principle applies to your company HR and your university administration.

\begin{center}
Bend one inch at a time.
\end{center}
Okay. Enough philosophy. Let's get back to our business.

\vspace{2ex}
DDPM has a lot of linkage to a piece of earlier work by Sohl-Dickstein et al in 2015 \cite{Sohl-Dickstein_2015_ICML}. Sohl-Dickstein et al asked the question of how to convert from one distribution to another distribution. VAE provides one approach: Referring to the previous section, we can think of the source distribution being the latent variable $\vz \sim p(\vz)$ and the target distribution being the input variable $\vx \sim p(\vx)$. Then by setting up the proxy distributions $p_{\vtheta}(\vx|\vz)$ and $q_{\vphi}(\vz|\vx)$, we can train the encoder and decoder so that the decoder will serve the goal of generating images. But VAE is largely a \emph{one-step} generation --- if you give us a latent code $\vz$, we ask the neural network $f_{\vtheta}(\cdot)$ to immediately return us the generated signal $\vx \sim \calN(\vx \;|\; f_{\vtheta}(\vz), \sigma_{\text{dec}}^2\mI)$. In some sense, this is asking a lot from the neural network. We are asking it to use a few layers of neurons to immediately convert from one distribution $p(\vz)$ to another distribution $p(\vx)$. This is too much.

The idea Sohl-Dickstein et al proposed was to construct a chain of conversions instead of a one-step process. To this end they defined two processes analogous to the encoder and decoder in a VAE. They call the encoder as the forward process, and the decoder as the reverse process. In both processes, they consider a sequence of variables $\vx_0,\ldots,\vx_T$ whose joint distribution is denoted as $q_{\vphi}(\vx_{0:T})$ and $p_{\vtheta}(\vx_{0:T})$ respectively for the forward and reverse processes. To make both processes tractable (and also flexible), they impose a Markov chain structure (i.e., memoryless) where
\begin{alignat*}{2}
\text{forward from $\vx_0$ to $\vx_T$}: & \qquad \qquad & q_{\vphi}(\vx_{0:T}) &= q(\vx_0) \prod_{t=1}^T q_{\vphi}(\vx_t \;|\; \vx_{t-1}),\\
\text{reverse from $\vx_T$ to $\vx_0$}: & \qquad \qquad & p_{\vtheta}(\vx_{0:T}) &= p(\vx_T) \prod_{t=1}^T p_{\vtheta}(\vx_{t-1} \;|\; \vx_{t}).
\end{alignat*}
In both equations, the transition distributions are only dependent on its immediate previous stage. Therefore, if each transition is realized through some form of neural networks, the overall generation process is broken down into many smaller tasks. It does not mean that we will need $T$ times more neural networks. We are just re-using one network for $T$ times.

Breaking the overall process into smaller steps allows us to use simple distributions at each step. As will be discussed in the following subsections, we can use Gaussian distributions for the transitions. Thanks to the properties of a Gaussian, the posterior will remain a Gaussian if the likelihood and the prior are both Gaussians. Therefore, if each transitional distribution above is a Gaussian, the joint distribution is also a Gaussian. Since a Gaussian is fully characterized by the first two moments (mean and variance), the computation is highly tractable. In the original paper of Sohl-Dickstein et al, there is also a case study of binomial diffusion processes.

\begin{figure}[h]
\centering
\includegraphics[width=0.7\linewidth]{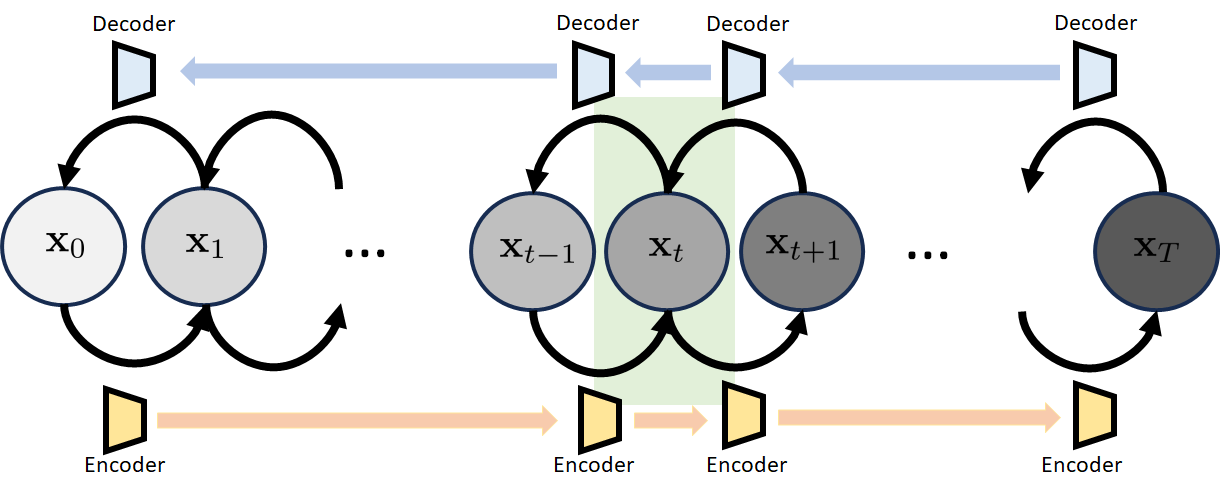}
\caption{Variational diffusion model by Kingma et al \cite{Kingma_2021_VDM}. In this model, the input image is $\vx_0$ and the white noise is $\vx_T$. The intermediate variables (or states) $\vx_1,\ldots,\vx_{T-1}$ are latent variables. The transition from $\vx_{t-1}$ to $\vx_t$ is analogous to the forward step (encoder) in VAE, whereas the transition from $\vx_t$ to $\vx_{t-1}$ is analogous to the reverse step (decoder) in VAE. In variational diffusion models, the input dimension and the output dimension of the encoders/decoders are identical. }
\label{fig: VDM block diagram}
\end{figure}

\vspace{4ex}
After providing a high-level overview of the concepts, let's talk about some details. The starting point of the diffusion model is to consider the VAE structure and make it a chain of incremental updates as shown in \fref{fig: VDM block diagram}. This particular structure is called the \textbf{variational diffusion model}, a name given by Kingma et al in 2021 \cite{Kingma_2021_VDM}. The variational diffusion model has a sequence of states $\vx_0,\vx_1,\ldots,\vx_T$ with the following interpretations:
\begin{itemize}
\setlength\itemsep{-1ex}
\item $\vx_0$: It is the original image, which is the same as $\vx$ in VAE.
\item $\vx_T$: It is the latent variable, which is the same as $\vz$ in VAE. As explained above, we choose $\vx_T \sim \calN(0,\mI)$ for simplicity, tractability, and computational efficiency.
\item $\vx_1,\ldots,\vx_{T-1}$: They are the intermediate states. They are also the latent variables, but they are not white Gaussian.
\end{itemize}
The structure of the variational diffusion model consists of two paths. The forward and the reverse paths are analogous to the paths of a single-step variational autoencoder. The difference is that the encoders and decoders have identical input-output dimensions. The assembly of all the forward building blocks will give us the encoder, and the assembly of all the reverse building blocks will give us the decoder.

\subsection{Building Blocks}
Let's talk about the building blocks of the variational diffusion model. There are three classes of building blocks: the transition block, the initial block, and the final block.

\begin{figure}[h]
\centering
\includegraphics[height=3.5cm]{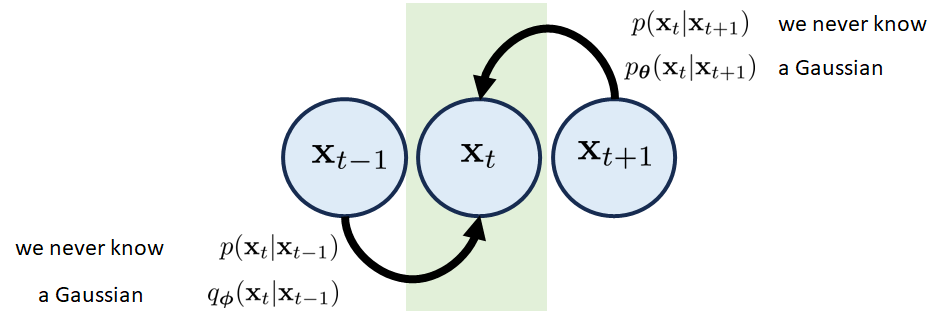}
\caption{The transition block of a variational diffusion model consists of three nodes. The transition distributions $p(\vx_t|\vx_{t+1})$ and $p(\vx_t|\vx_{t-1})$ are not accessible, but we can approximate them by Gaussians. }
\label{fig: VDM transition block}
\end{figure}

\textbf{Transition Block} The $t$-th transition block consists of three states $\vx_{t-1}$, $\vx_{t}$, and $\vx_{t+1}$. There are two possible paths to get to state $\vx_t$, as illustrated in \fref{fig: VDM transition block}.

\begin{itemize}
\setlength\itemsep{-1ex}
\item The first path is the forward transition going from $\vx_{t-1}$ to $\vx_t$. The associated transition distribution is $p(\vx_{t}|\vx_{t-1})$. In plain words, if you tell us $\vx_{t-1}$, we can draw a sample $\vx_t$ according to $p(\vx_{t}|\vx_{t-1})$. However, just like a VAE, the transition distribution $p(\vx_{t}|\vx_{t-1})$ is not accessible. We can approximate it by some simple distributions such as a Gaussian. The approximated distribution is denoted as $q_{\vphi}(\vx_t|\vx_{t-1})$. We will discuss the exact form of $q_{\vphi}$ later.
\item The second path is the reverse transition going from $\vx_{t+1}$ to $\vx_t$. Again, we do not know $p(\vx_{t}|\vx_{t+1})$ and so we have another proxy distribution, e.g. a Gaussian, to approximate the true distribution. This proxy distribution is denoted as $p_{\vtheta}(\vx_{t}|\vx_{t+1})$.
\end{itemize}

\begin{figure}[h]
\centering
\includegraphics[height=3cm]{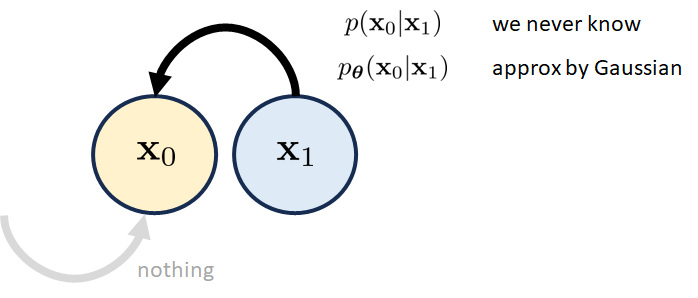}
\caption{The initial block of a variational diffusion model focuses on the node $\vx_0$. Since there is no state before time $t = 0$, we only have a reverse transition from $\vx_1$ to $\vx_0$.}
\label{fig: VDM beginning node}
\end{figure}

\textbf{Initial Block} The initial block of the variational diffusion model focuses on the state $\vx_0$. Since we start at $\vx_0$, we only need the reverse transition from $\vx_1$ to $\vx_0$. The forward transition from $\vx_{-1}$ to $\vx_0$ can be dropped. Therefore, we only need to consider $p(\vx_0|\vx_1)$. But since $p(\vx_0|\vx_1)$ is never accessible, we approximate it by a Gaussian  $p_{\vtheta}(\vx_0|\vx_1)$ where the mean is computed through a neural network. See \fref{fig: VDM beginning node} for illustration.

\begin{figure}[h]
\centering
\includegraphics[height=3cm]{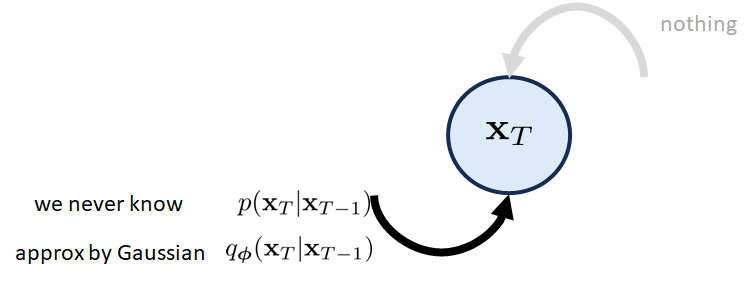}
\caption{The final block of a variational diffusion model focuses on the node $\vx_T$. Since there is no state after time $t = T$, we only have a forward transition from $\vx_{T-1}$ to $\vx_{T}$.}
\label{fig: VDM final node}
\end{figure}

\textbf{Final Block}. The final block focuses on the state $\vx_T$. Remember that $\vx_T$ is supposed to be our final latent variable which is a white Gaussian noise vector. Because it is the final block, we only need a forward transition from $\vx_{T-1}$ to $\vx_T$, and nothing such as $\vx_{T+1}$ to $\vx_T$. The forward transition is approximated by $q_{\vphi}(\vx_T|\vx_{T-1})$ which is a Gaussian. See \fref{fig: VDM final node} for illustration.

\vspace{2ex}
\textbf{Understanding the Transition Distribution}. Before we proceed further, we need to explain the transition distribution $q_{\vphi}(\vx_t|\vx_{t-1})$. We know that it is a Gaussian. But what is the mean and variance of this Gaussian?

\boxeddef{
\textbf{Transition Distribution $q_{\vphi}(\vx_t|\vx_{t-1})$}. In a variational diffusion model (and also DDPM which we will discuss later), the transition distribution $q_{\vphi}(\vx_t|\vx_{t-1})$ is defined as
\begin{equation}
q_{\vphi}(\vx_t|\vx_{t-1}) \bydef \calN(\vx_t \,|\, \sqrt{\alpha_t} \vx_{t-1},(1-\alpha_t)\mI).
\label{eq: transition q from t-1 to t}
\end{equation}
}
In other words, $q_{\vphi}(\vx_t|\vx_{t-1})$ is a Gaussian. The mean is $\sqrt{\alpha_t} \vx_{t-1}$ and the variance is $1-\alpha_t$. The choice of the scaling factor $\sqrt{\alpha_t}$ is to make sure that the variance magnitude is preserved so that it will not explode and vanish after many iterations.

\boxedeg{
\label{example: DDPM GMM 1}
Let's consider a Gaussian mixture model
\begin{equation*}
\vx_0 \sim p_0(\vx) = \pi_1 \calN(\vx| \mu_1, \sigma_1^2) + \pi_2 \calN(\vx| \mu_2, \sigma_2^2).
\end{equation*}
Given the transition probability, we know that if $\vx_t \sim q_{\vphi}(\vx_t|\vx_{t-1})$ then
\begin{equation*}
\vx_t = \sqrt{\alpha_t} \vx_{t-1} + \sqrt{(1-\alpha_t)}\vepsilon, \qquad \text{where}\;\; \vepsilon \sim \calN(0,\mI).
\end{equation*}
Our goal is to see whether this iterative procedure (using the above transition probability) will give us a white Gaussian in the equilibrium state (i.e., when $t \rightarrow \infty$).

For a mixture model, it is not difficult to show that the probability distribution of $\vx_t$ can be calculated recursively via the algorithm for $t = 1, 2, \ldots,T$: (the proof will be shown later)
\begin{align}
\vx_t \sim p_t(\vx) =
&\pi_1 \calN(\vx| \sqrt{\alpha_t} \mu_{1,t-1}, \;\; \alpha_t \sigma_{1,t-1}^2 + (1-\alpha_t) ) \notag \\
&\qquad + \pi_2 \calN(\vx| \sqrt{\alpha_t} \mu_{2,t-1}, \;\; \alpha_t \sigma_{2,t-1}^2 + (1-\alpha_t) ), \label{eq: Sec 2 GMM transition example}
\end{align}
where $\mu_{1,t-1}$ is the mean for class $1$ at time $t - 1$, with $\mu_{1,0} = \mu_1$ being the initial mean. Similarly, $\sigma_{1,t-1}^2$ is the variance for class $1$ at time $t-1$, with $\sigma_{1,0}^2 = \sigma_1^2$ being the initial variance.

In the figure below, we show a numerical example where $\pi_1 = 0.3$, $\pi_2 = 0.7$, $\mu_1 = -2$, $\mu_2 = 2$, $\sigma_1 = 0.2$, and $\sigma_2 = 1$. The rate is defined as $\alpha_t = 0.97$ for all $t$. We plot the probability distribution function for different $t$.

\begin{center}
\begin{tabular}{cccc}
\hspace{-2ex}\includegraphics[width=0.23\linewidth]{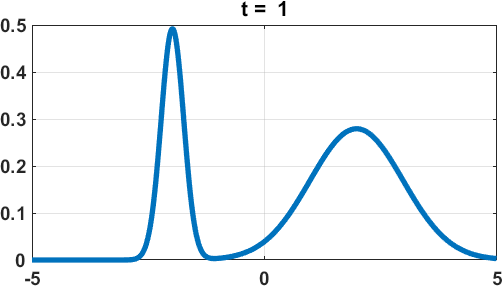}&
\hspace{-2ex}\includegraphics[width=0.23\linewidth]{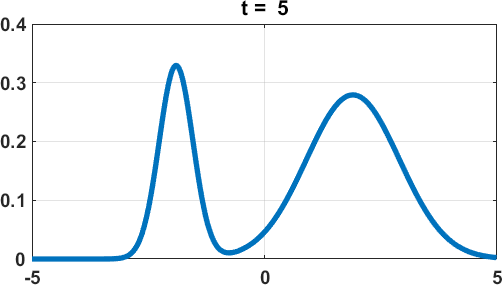}&
\hspace{-2ex}\includegraphics[width=0.23\linewidth]{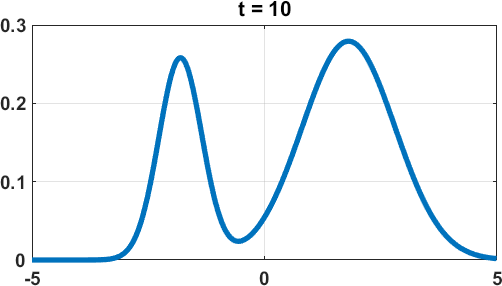}&
\hspace{-2ex}\includegraphics[width=0.23\linewidth]{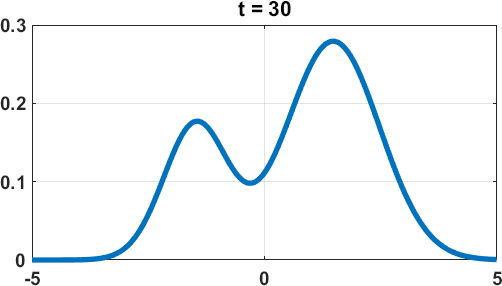}\\
\hspace{-2ex}\includegraphics[width=0.23\linewidth]{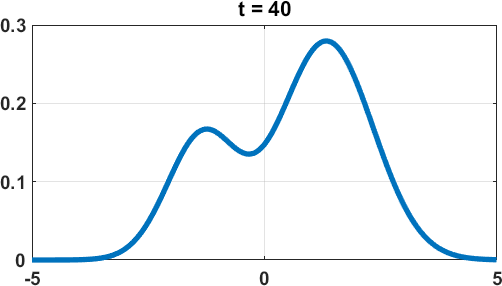}&
\hspace{-2ex}\includegraphics[width=0.23\linewidth]{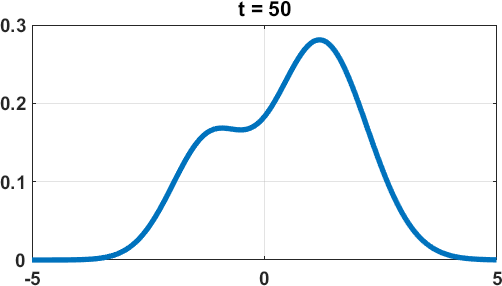}&
\hspace{-2ex}\includegraphics[width=0.23\linewidth]{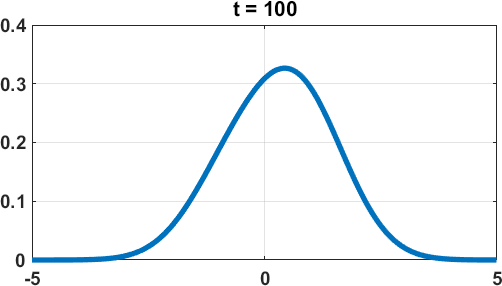}&
\hspace{-2ex}\includegraphics[width=0.23\linewidth]{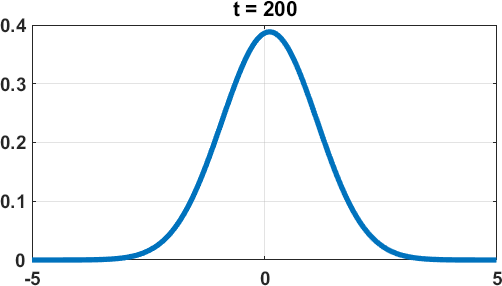}
\end{tabular}
\captionof{figure}{Evolution of the distribution $p_t(\vx)$. As time $t$ progresses, the bimodal distribution gradually becomes a Gaussian.}
\end{center}
}

\boxedproof{
\textbf{Proof of \eref{eq: Sec 2 GMM transition example}}. For those who would like to understand how we derive the probability density of a mixture model in \eref{eq: Sec 2 GMM transition example}, we can show a simple derivation. Consider a mixture model
\begin{align*}
p(\vx) = \sum_{k=1}^K \pi_k
\underset{p(\vx|k)}{\underbrace{\calN(\vx|\mu_k,\sigma_k^2\mI)}}.
\end{align*}
If we consider a new variable $\vy = \sqrt{\alpha}\vx + \sqrt{1-\alpha}\vepsilon$ where $\vepsilon \sim \calN(0,\mI)$, then the distribution of $\vy$ can be derived by using the law of total probability:
\begin{align*}
p(\vy)
&= \sum_{k=1}^K p(\vy|k)p(k) = \sum_{k=1}^K \pi_k p(\vy|k).
\end{align*}
Since $\vy|k = \sqrt{\alpha}\vx|k + \sqrt{1-\alpha}\vepsilon$ is a linear combination of a (conditioned) Gaussian random variable $\vx|k$ and another Gaussian random variable $\vepsilon$, the sum $\vy|k$ will remain as a Gaussian. The mean and variance are
\begin{align*}
\E[\vy|k]   &= \sqrt{\alpha}\E[\vx|k] + \sqrt{1-\alpha}\E[\vepsilon] = \sqrt{\alpha}\mu_k, \\
\Var[\vy|k] &= \alpha\Var[\vx|k] + (1-\alpha)\Var[\vepsilon] = \alpha \sigma_k^2 + (1-\alpha),
\end{align*}
where we used the fact that $\E[\vepsilon] = 0$, and $\Var[\vepsilon] = 1$. Since we just argued that $\vy|k$ is a Gaussian, the distribution of $\vy|k$ is completely specified once we know the mean and variance. Substituting the above derived results, we know that $p(\vy|k) = \calN(\vy|\sqrt{\alpha}\mu_k,  \alpha \sigma_k^2 + (1-\alpha))$. This completes the derivation.
}

\textbf{The magical scalars $\sqrt{\alpha_t}$ and $1-\alpha_t$}. You may wonder how the genius people (the authors of the denoising diffusion papers) come up with the magical scalars $\sqrt{\alpha}_t$ and $(1-\alpha_t)$ for the above transition probability. To demystify this, let's consider two unrelated scalars $a \in \R$ and $b \in \R$, and define the transition distribution as
\begin{equation}
q_{\vphi}(\vx_t|\vx_{t-1}) = \calN(\vx_t \,|\, a \vx_{t-1}, b^2 \mI).
\label{eq: transition q from t-1 to t, a b}
\end{equation}
Here is the finding:
\boxedthm{
\textbf{(Why $\sqrt{\alpha}$ and $1-\alpha$?)}
Suppose that $q_{\vphi}(\vx_t|\vx_{t-1}) = \calN(\vx_t \,|\, a \vx_{t-1}, b^2 \mI)$ for some constants $a$ and $b$. If we want to choose $a$ and $b$ such that the distribution of $\vx_t$ will become $\calN(0,\mI)$, then it is necessary that
\begin{equation*}
a = \sqrt{\alpha} \qquad \text{and} \qquad b = \sqrt{1-\alpha}.
\end{equation*}
Therefore, the transition distribution is
\begin{equation}
q_{\vphi}(\vx_t|\vx_{t-1}) \bydef \calN(\vx_t \,|\, \sqrt{\alpha} \vx_{t-1},(1-\alpha)\mI).
\end{equation}
}
Remark: You can replace $\alpha$ by $\alpha_t$, if you prefer a noise schedule.

\boxedproof{
\textbf{Proof}. We want to show that $a = \sqrt{\alpha}$ and $b = \sqrt{1-\alpha}$. For the distribution shown in \eref{eq: transition q from t-1 to t, a b}, the equivalent sampling step is:
\begin{equation}
\vx_t = a \vx_{t-1} + b\vepsilon_{t-1}, \qquad\text{where}\qquad \vepsilon_{t-1} \sim \calN(0,\mI).
\end{equation}

We can carry on the recursion to show that
\begin{alignat}{2}
\vx_t
&= a \vx_{t-1} + b\vepsilon_{t-1} \notag \\
&= a( a\vx_{t-2} + b \vepsilon_{t-2} ) + b\vepsilon_{t-1}
&&\qquad \textcolor{purple}{(\text{substitute } \vx_{t-1} = a\vx_{t-2} + b \vepsilon_{t-2})} \notag \\
&= a^2 \vx_{t-2} + ab \vepsilon_{t-2} + b\vepsilon_{t-1}
&&\qquad \textcolor{purple}{(\text{regroup terms })}\notag \\
&= \vdots\notag \\
&= a^t \vx_0 + b
\underset{\bydef \vw_t}{\underbrace{\left[ \vepsilon_{t-1} + a \vepsilon_{t-2} + a^2 \vepsilon_{t-3} + \ldots + a^{t-1}\vepsilon_{0}\right]}}.
\end{alignat}
The finite sum above is a sum of independent Gaussian random variables. The mean vector $\E[\vw_t]$ remains zero because everyone has a zero mean. The covariance matrix (for a zero-mean vector) is
\begin{align*}
\text{Cov}[\vw_t] &\bydef \E[\vw_t\vw_t^T] \\
&= b^2(\text{Cov}(\vepsilon_{t-1}) + a^2\text{Cov}(\vepsilon_{t-2}) + \ldots + (a^{t-1})^2\text{Cov}(\vepsilon_{0}))\\
&= b^2(1 + a^2 + a^4 + \ldots + a^{2(t-1)})\mI \\
&= b^2 \cdot \frac{1-a^{2t}}{1-a^2}\mI.
\end{align*}
As $t \rightarrow \infty$, $a^t \rightarrow 0$ for any $0 < a < 1$. Therefore, at the limit when $t = \infty$,
\begin{align*}
\lim_{t\rightarrow \infty} \text{Cov}[\vw_t] = \frac{b^2}{1-a^2}\mI.
\end{align*}
So, if we want $\lim_{t\rightarrow \infty} \text{Cov}[\vw_t] = \mI$ (so that the distribution of $\vx_t$ will approach $\calN(0,\mI)$), then we need
\begin{equation*}
1 = \frac{b^2}{1-a^2},
\end{equation*}
or equivalently $b = \sqrt{1-a^2}$. Now, if we let $a = \sqrt{\alpha}$, then $b = \sqrt{1-\alpha}$. This will give us
\begin{equation}
\vx_t = \sqrt{\alpha} \vx_{t-1} + \sqrt{1-\alpha} \vepsilon_{t-1}.
\end{equation}
}

\textbf{Distribution $q_{\vphi}(\vx_t|\vx_{0})$}. With the understanding of the magical scalars, we can talk about the distribution $q_{\vphi}(\vx_t|\vx_{0})$. That is, we want to know how $\vx_t$ will be distributed if we are given $\vx_0$.

\boxedthm{
\textbf{(Conditional Distribution $q_{\vphi}(\vx_t|\vx_0)$)}. The conditional distribution $q_{\vphi}(\vx_t|\vx_0)$ is given by
\begin{equation}
q_{\vphi}(\vx_t|\vx_0) = \calN(\vx_t \,|\, \sqrt{\overline{\alpha}}_t \vx_0, \;\; (1-\overline{\alpha}_t)\mI),
\label{eq: q t | 0}
\end{equation}
where $\overline{\alpha}_t = \prod_{i=1}^{t} \alpha_i$.
}

\boxedproof{
\textbf{Proof}. To see how \eref{eq: q t | 0} is derived, we can re-do the recursion but this time we use $\sqrt{\alpha_t}\vx_{t-1}$ and $(1-\alpha_t)\mI$ as the mean and covariance, respectively. This will give us
\begin{align}
\vx_t
&= \sqrt{\alpha_{t}} \vx_{t-1} + \sqrt{1-\alpha_t} \vepsilon_{t-1} \notag \\
&= \sqrt{\alpha_{t}} ( \sqrt{\alpha_{t-1}} \vx_{t-2} + \sqrt{1-\alpha_{t-1}} \vepsilon_{t-2}) + \sqrt{1-\alpha_t} \vepsilon_{t-1} \notag\\
&= \sqrt{\alpha_{t} \alpha_{t-1}}  \vx_{t-2} +
\underset{\vw_{1}}{\underbrace{\sqrt{\alpha_{t}} \sqrt{1-\alpha_{t-1}} \vepsilon_{t-2} + \sqrt{1-\alpha_t} \vepsilon_{t-1}}}.
\label{eq: recursion for xt, using alpha_t}
\end{align}
Therefore, we have a sum of two Gaussians. But since sum of two Gaussians remains a Gaussian, we can just calculate its new covariance (because the mean remains zero). The new covariance is
\begin{align*}
\E[\vw_1\vw_1^T]
&= [(\sqrt{\alpha_{t}} \sqrt{1-\alpha_{t-1}})^2 + (\sqrt{1-\alpha_t})^2]\mI\\
&= [\alpha_{t} (1-\alpha_{t-1}) + 1-\alpha_t]\mI = [1-\alpha_{t}\alpha_{t-1}]\mI.
\end{align*}
Returning to \eref{eq: recursion for xt, using alpha_t}, we can show that the recursion becomes a linear combination of $\vx_{t-2}$ and a noise vector $\vepsilon_{t-2}$:
\begin{align}
\vx_t
&= \sqrt{\alpha_{t} \alpha_{t-1}}  \vx_{t-2} + \sqrt{1-\alpha_{t}\alpha_{t-1}}\vepsilon_{t-2} \notag \\
&=\sqrt{\alpha_{t} \alpha_{t-1} \alpha_{t-2}}  \vx_{t-3} + \sqrt{1-\alpha_{t}\alpha_{t-1} \alpha_{t-2}}\vepsilon_{t-3} \notag \\
&= \vdots \notag \\
&= \left(\sqrt{\prod_{i=1}^{t} \alpha_i}\right) \vx_{0} + \left(\sqrt{1- \prod_{i=1}^{t} \alpha_i}\right) \vepsilon_{0}.
\end{align}
So, if we define $\overline{\alpha}_t = \prod_{i=1}^{t} \alpha_i$, we can show that
\begin{align}
\vx_t = \sqrt{ \overline{\alpha}_t } \vx_0 + \sqrt{1-\overline{\alpha}_t} \vepsilon_0.
\label{eq: recursion for xt, using x0 and epsilon0}
\end{align}
In other words, the distribution $q_{\vphi}(\vx_t|\vx_0)$ is
\begin{align}
\vx_t \sim q_{\vphi}(\vx_t|\vx_0) = \calN(\vx_t \,|\, \sqrt{\overline{\alpha}_t} \vx_0, \;\; (1-\overline{\alpha}_t)\mI).
\label{eq: recursion for xt, using alpha_t final}
\end{align}
}

\begin{figure}[h]
\centering
\includegraphics[width=0.8\linewidth]{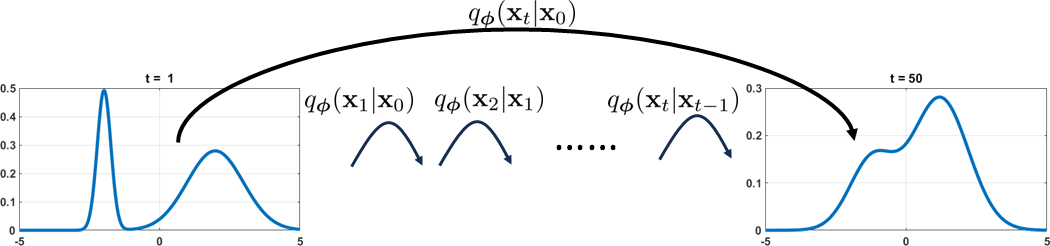}
\caption{The difference between $q_{\vphi}(\vx_t|\vx_{t-1})$ and $q_{\vphi}(\vx_t|\vx_0)$.}
\label{fig: one-step vs multi-step}
\end{figure}

The utility of the new distribution $q_{\vphi}(\vx_t|\vx_0)$ is its one-shot forward diffusion step compared to the chain $\vx_0 \rightarrow \vx_1 \rightarrow \ldots \rightarrow \vx_{T-1} \rightarrow \vx_{T}$. In every step of the forward diffusion model, since we already know $\vx_0$ and we assume that all subsequence transitions are Gaussian, we will know $\vx_t$ for any $t$. The situation can be understood from \fref{fig: one-step vs multi-step}.

\boxedeg{
\label{example: DDPM GMM 2}
For a Gaussian mixture model such that $\vx_0 \sim p_0(\vx) = \sum_{k=1}^K \pi_k \calN(\vx|\vmu_k,\sigma_k^2\mI)$, we can show that the distribution at time $t$ is
\begin{align}
\vx_t \sim p_t(\vx)
&= \sum_{k=1}^K \pi_k \calN(\vx \;|\; \sqrt{\overline{\alpha}_t} \vmu_k, (1-\overline{\alpha}_t)\mI + \overline{\alpha}_t \sigma_k^2\mI ) \label{eq: ddpm one shot eqn} \\
&= \sum_{k=1}^K \pi_k \calN(\vx \;|\; \sqrt{ \alpha^t } \vmu_k, (1- \alpha^t )\mI +  \alpha^t  \sigma_k^2\mI ), \qquad \text{if } \; \alpha_t = \alpha \;\; \text{so that }\; \overline{\alpha}_t = \prod_{i=1}^t \alpha = \alpha^t.\notag
\end{align}

If you are curious about how the probability distribution $p_t$ evolves over time $t$, we can visualize the trajectory of a Gaussian mixture distribution we discussed in Example~\ref{example: DDPM GMM 1}. We use \eref{eq: ddpm one shot eqn} to plot the heatmap. You can see that when $t = 0$, the initial distribution is a mixture of two Gaussians. As we progress by following the transition defined in \eref{eq: ddpm one shot eqn}, we can see that the distribution gradually becomes the single Gaussian $\calN(0,\mI)$.

\begin{center}
\includegraphics[width=0.45\linewidth]{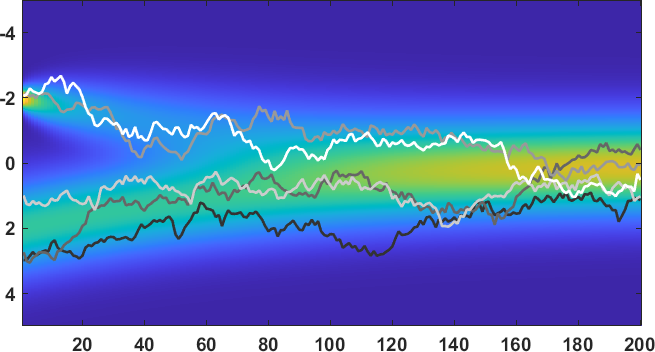}
\captionof{figure}{Realizations of random trajectories made by $\vx_t$. The color map in the background indicates the probability distribution $p_t(\vx)$.}
\end{center}

In the same plot, we overlay and show a few instantaneous trajectories of the random samples $\vx_t$ as a function of time $t$. The equation we used to generate the samples is
\begin{equation*}
\vx_t = \sqrt{\alpha_t} \vx_{t-1} + \sqrt{1-\alpha_t}\vepsilon, \qquad \vepsilon \sim \calN(0,\mI).
\end{equation*}
As you can see, the trajectories of $\vx_t$ more or less follow the distribution $p_t(\vx)$.
}

A confusing point to many readers is that if the goal is to convert from an image $p(\vx_0)$ to white noise $p(\vx_T)$, what is the point of deriving $q_{\vphi}(\vx_t|\vx_{t-1})$ and $q_{\vphi}(\vx_t \;|\; \vx_0)$? The answer is that so far we have been just talking about the \emph{forward} process. In a diffusion model, the forward process is chosen such that they can be expressed in a closed form. The more interesting part is the \emph{reverse} process. As will be discussed, the reverse process is realized through a chain of denoising operations. Each denoising step should be coupled with the corresponding step in the forward process. $q_{\vphi}(\vx_t \;|\; \vx_0)$ just provides us a slightly more convenient way to implement the forward process.

\subsection{Evidence Lower Bound} Now that we understand the structure of the variational diffusion model, we can write down the ELBO and hence train the model.
\boxedthm{
\label{thm: ELBO for VDM}
\textbf{(ELBO for Variational Diffusion Model)}. The ELBO for the variational diffusion model is
\begin{align}
\text{ELBO}_{\vphi,\vtheta}(\vx)
&=
\E_{q_{\vphi}(\vx_1|\vx_0)}\Big[\log
\underset{\text{how good the initial block is}}{\underbrace{p_{\vtheta}(\vx_0|\vx_1)}} \Big] \notag \\
&\qquad -\E_{q_{\vphi}(\vx_{T-1}|\vx_0)} \Big[
\underset{\text{how good the final block is}}{\underbrace{\mathbb{D}_{\text{KL}}\Big(q_{\vphi}(\vx_T|\vx_{T-1}) \| p(\vx_T) \Big)}} \Big] \notag \\
&\qquad -\sum_{t=1}^{T-1} \E_{q_{\vphi}(\vx_{t-1},\vx_{t+1}|\vx_0)} \Big[
\underset{\text{how good the transition blocks are}}{\underbrace{\mathbb{D}_{\text{KL}}\Big(q_{\vphi}(\vx_t|\vx_{t-1}) \| p_{\vtheta}(\vx_t|\vx_{t+1}) \Big) }}
\Big],
\label{eq: ELBO VDM}
\end{align}
where $\vx_0 = \vx$, and $\vx_T \sim \calN(0,\mI)$.
}
If you are a casual reader of this tutorial, we hope that this equation does not throw you off. While it appears a monster, it does have structures. We just need to be patient when we try to understand it.

\textbf{Reconstruction} (Initial Block). Let's first look at the term
\begin{equation*}
\E_{q_{\vphi}(\vx_1|\vx_0)}\Big[\log \;\; p_{\vtheta}(\vx_0|\vx_1) \Big].
\end{equation*}
This term is based on the initial block and it is analogous to \eref{eq: ELBO for VAE}. The subject inside the expectation is the log-likelihood $\log p_{\vtheta}(\vx_0|\vx_1)$. This log-likelihood measures how good the neural network (associated with $p_{\vtheta}$) can recover $\vx_0$ from the latent variable $\vx_1$.

The expectation is taken with respect to the samples drawn from $q_{\vphi}(\vx_1|\vx_0)$. Recall that $q_{\vphi}(\vx_1|\vx_0)$ is the distribution that generates $\vx_1$. We require $\vx_1$ to be drawn from this distribution because $\vx_1$ do not come from the sky but \emph{created} by the forward transition $q_{\vphi}(\vx_1|\vx_0)$. The conditioning on $\vx_0$ is needed here because we need to know what the original image is.

The reason why expectation is used here is that $p_{\vtheta}(\vx_0|\vx_1)$ is a function of $\vx_1$ (and so if $\vx_1$ is random then $p_{\vtheta}(\vx_0|\vx_1)$ is random too). For a different intermediate state $\vx_1$, the probability $p_{\vtheta}(\vx_0|\vx_1)$ will be different. The expectation eliminates the dependency on $\vx_1$.

\vspace{2ex}
\textbf{Prior Matching} (Final Block). The prior matching term is
\begin{equation}
-\E_{q_{\vphi}(\vx_{T-1}|\vx_0)} \Big[ \mathbb{D}_{\text{KL}}\Big(q_{\vphi}(\vx_T|\vx_{T-1}) \| p(\vx_T) \Big) \Big],
\end{equation}
and it is based on the final block. We use the KL divergence to measure the difference between $q_{\vphi}(\vx_T|\vx_{T-1})$ and $p(\vx_T)$. The distribution $q_{\vphi}(\vx_T|\vx_{T-1})$ is the forward transition from $\vx_{T-1}$ to $\vx_T$. This describes how $\vx_T$ is generated. The second distribution is $p(\vx_T)$. Because of our laziness, we assume that $p(\vx_T) = \calN(0,\mI)$. We want $q_{\vphi}(\vx_T|\vx_{T-1})$ to be as close to $\calN(0,\mI)$ as possible.

When computing the KL-divergence, the variable $\vx_T$ is a dummy variable. However, since $q_{\vphi}$ is conditioned on $\vx_{T-1}$, the KL-divergence calculated here is a function of the conditioned variable $\vx_{T-1}$. Where does $\vx_{T-1}$ come from? It is generated by $q_{\vphi}(\vx_{T-1}|\vx_0)$. We use a conditional distribution $q_{\vphi}(\vx_{T-1}|\vx_0)$ because $\vx_{T-1}$ depends on what $\vx_0$ we use in the first place. The expectation over $q_{\vphi}(\vx_{T-1}|\vx_0)$ says that for each of the $\vx_{T-1}$ generated by $q_{\vphi}(\vx_{T-1}|\vx_0)$, we will have a value of the KL divergence. We take the expectation over all the possible $\vx_{T-1}$ generated to eliminate the dependency.

\vspace{2ex}
\textbf{Consistency}. (Transition Blocks) The consistency term is
\begin{equation}
-\sum_{t=1}^{T-1} \E_{q_{\vphi}(\vx_{t-1},\vx_{t+1}|\vx_0)}
\Big[\mathbb{D}_{\text{KL}}\Big(q_{\vphi}(\vx_t|\vx_{t-1}) \| p_{\vtheta}(\vx_t|\vx_{t+1}) \Big) \Big],
\end{equation}
and it is based on the transition blocks. There are two directions if you recall \fref{fig: VDM transition block}. The forward transition is determined by the distribution $q_{\vphi}(\vx_t|\vx_{t-1})$ whereas the reverse transition is determined by another distribution $p_{\vtheta}(\vx_{t}|\vx_{t+1})$.
The consistency term uses the KL divergence to measure the deviation.

The expectation is taken with respect to the pair of samples $(\vx_{t-1},\vx_{t+1})$, drawn from $q_{\vphi}(\vx_{t-1},\vx_{t+1}|\vx_0)$. The reason is that the KL divergence above is a function of $\vx_{t-1}$ and $\vx_{t+1}$. (You can ignore $\vx_t$ because it is a dummy variable that will be eliminated during the integration process when we calculate the expectation.) Because of the dependencies on $\vx_{t-1}$ and $\vx_{t+1}$, we need to take the expectation.

\boxedproof{
\textbf{Proof of Theorem~\ref{thm: ELBO for VDM}}. Let's define the following notation: $\vx_{0:T} = \{\vx_0,\ldots,\vx_T\}$ means the collection of all state variables from $t=0$ to $t=T$. We also recall that the prior distribution $p(\vx)$ is the distribution for the image $\vx_0$. So it is equivalent to $p(\vx_0)$. With these in mind, we can show that
\begin{alignat*}{2}
\log p(\vx) &= \log p(\vx_0)                     & & \\
            &= \log \int p(\vx_{0:T}) d\vx_{1:T} & \qquad& \textcolor{purple}{(\text{Marginalize by integrating over $\vx_{1:T}$})}\\
            &= \log \int p(\vx_{0:T}) \frac{\textcolor{blue}{q_{\vphi}(\vx_{1:T}|\vx_0)}}{\textcolor{blue}{q_{\vphi}(\vx_{1:T}|\vx_0)}} d\vx_{1:T}  & \qquad & \textcolor{purple}{(\text{Multiply and divide $q_{\vphi}(\vx_{1:T}|\vx_0)$})}\\
            &= \log \int \textcolor{blue}{q_{\vphi}(\vx_{1:T}|\vx_0)} \left[ \frac{p(\vx_{0:T})}{\textcolor{blue}{q_{\vphi}(\vx_{1:T}|\vx_0)}}\right]  d\vx_{1:T} &\qquad & \textcolor{purple}{(\text{Rearrange terms})}\\
            &= \log \E_{\textcolor{blue}{q_{\vphi}(\vx_{1:T}|\vx_0)}}\left[ \frac{p(\vx_{0:T})}{\textcolor{blue}{q_{\vphi}(\vx_{1:T}|\vx_0)}}\right] &\qquad & \textcolor{purple}{(\text{Definition of expectation})}.
\end{alignat*}
Now, we need to use Jensen's inequality, which states that for any random variable $X$ and any concave function $f$, it holds that $f(\E[X]) \ge \E[f(X)]$. By recognizing that $f(\cdot) = \log(\cdot)$, we can show that
\begin{alignat}{2}
\log p(\vx)
= \log \E_{q_{\vphi}(\vx_{1:T}|\vx_0)}\left[ \frac{p(\vx_{0:T})}{q_{\vphi}(\vx_{1:T}|\vx_0)}\right]
&\ge \E_{q_{\vphi}(\vx_{1:T}|\vx_0)}\left[ \log \frac{p(\vx_{0:T})}{q_{\vphi}(\vx_{1:T}|\vx_0)}\right]
\label{eq: ELBO VDM step inequality}
\end{alignat}
Let's take a closer look at $p(\vx_{0:T})$. Inspecting \fref{fig: VDM transition block}, we notice that if we want to decouple $p(\vx_{0:T})$, we should do conditioning for $\vx_{t-1}|\vx_t$. This leads to:
\begin{align}
p(\vx_{0:T}) = p(\vx_T) \prod_{t=1}^T p(\vx_{t-1}|\vx_t) = p(\vx_T)p(\vx_0|\vx_1)\prod_{t=2}^T p(\vx_{t-1}|\vx_t).
\label{eq: ELBO VDM step p}
\end{align}
As for $q_{\vphi}(\vx_{1:T}|\vx_0)$, \fref{fig: VDM transition block} suggests that we need to do the conditioning for $\vx_{t}|\vx_{t-1}$. However, because of the sequential relationship, we can write
\begin{align}
q_{\vphi}(\vx_{1:T}|\vx_0)
&= \prod_{t=1}^T q_{\vphi}(\vx_{t}|\vx_{t-1}) = q_{\vphi}(\vx_T | \vx_{T-1})  \prod_{t=1}^{T-1} q_{\vphi}(\vx_t | \vx_{t-1}).
\label{eq: ELBO VDM step q}
\end{align}
Substituting \eref{eq: ELBO VDM step p} and \eref{eq: ELBO VDM step q} back to \eref{eq: ELBO VDM step inequality}, we can show that
\begin{alignat*}{2}
\log p(\vx)
&\ge \E_{q_{\vphi}(\vx_{1:T}|\vx_0)}\left[ \log \frac{p(\vx_{0:T})}{q_{\vphi}(\vx_{1:T}|\vx_0)}\right] & \qquad & \\
&= \E_{q_{\vphi}(\vx_{1:T}|\vx_0)} \left[ \log \frac{ \textcolor{blue}{p(\vx_T)p(\vx_0|\vx_1)\prod_{t=2}^T p(\vx_{t-1}|\vx_t)} }{ \textcolor{blue}{q_{\vphi}(\vx_T | \vx_{T-1})  \prod_{t=1}^{T-1} q_{\vphi}(\vx_t | \vx_{t-1})}}  \right] & \qquad & \\
&= \E_{q_{\vphi}(\vx_{1:T}|\vx_0)} \left[ \log \frac{ p(\vx_T)p(\vx_0|\vx_1)
\textcolor{blue}{\prod_{t=1}^{T-1} p(\vx_{t}|\vx_{t+1}) }}{ q_{\vphi}(\vx_T | \vx_{T-1})  \prod_{t=1}^{T-1} q_{\vphi}(\vx_t | \vx_{t-1})}  \right] &\qquad &
\textcolor{purple}{(\text{shift $t$ to $t+1$})}\\
&= \E_{q_{\vphi}(\vx_{1:T}|\vx_0)} \left[ \log \frac{p(\vx_T)p(\vx_0|\vx_1)}{q_{\vphi}(\vx_T | \vx_{T-1}) }  \right]
+ \E_{q_{\vphi}(\vx_{1:T}|\vx_0)} \left[ \log \prod_{t=1}^{T-1} \frac{p(\vx_{t}|\vx_{t+1})}{q_{\vphi}(\vx_t | \vx_{t-1})} \right]
&\qquad & \textcolor{purple}{(\text{split expectation})}
\end{alignat*}

The first term above can be further decomposed into two expectations
\begin{align*}
\E_{q_{\vphi}(\vx_{1:T}|\vx_0)} \left[ \log \frac{p(\vx_T)p(\vx_0|\vx_1)}{q_{\vphi}(\vx_T | \vx_{T-1}) }  \right]
&=
\underset{\text{Reconstruction}}{\underbrace{\E_{q_{\vphi}(\vx_{1:T}|\vx_0)} \bigg[ \log p(\vx_0|\vx_1) \bigg]}} +
\underset{\text{Prior Matching}}{\underbrace{\E_{q_{\vphi}(\vx_{1:T}|\vx_0)} \left[ \log \frac{p(\vx_T)}{q_{\vphi}(\vx_T | \vx_{T-1}) }  \right]}}.
\end{align*}

The Reconstruction term can be simplified as
\begin{align*}
\E_{q_{\vphi}(\vx_{1:T}|\vx_0)} \bigg[ \log p(\vx_0|\vx_1) \bigg]
&= \E_{q_{\vphi}(\vx_{1}|\vx_0)} \bigg[ \log p(\vx_0|\vx_1) \bigg],
\end{align*}
where we used the fact that the conditioning $\vx_{1:T}|\vx_0$ is equivalent to $\vx_{1}|\vx_0$ when the subject of interest (i.e., $\log p(\vx_0|\vx_1)$) only involves $\vx_0$ and $\vx_1$.

The Prior Matching term is
\begin{align*}
   \E_{q_{\vphi}(\vx_{1:T}|\vx_0)}       \left[ \log \frac{p(\vx_T)}{q_{\vphi}(\vx_T | \vx_{T-1})} \right]
&= \E_{q_{\vphi}(\vx_T,\vx_{T-1}|\vx_0)} \left[ \log \frac{p(\vx_T)}{q_{\vphi}(\vx_T | \vx_{T-1})} \right],
\end{align*}
where we note that the conditional expectation can be simplified to samples $\vx_T$ and $\vx_{T-1}$ only, because $\log \frac{p(\vx_T)}{q_{\vphi}(\vx_T | \vx_{T-1})} $ only depends on $\vx_T$ and $\vx_{T-1}$. For the expectation term, chain rule of probability tells us that $q_{\vphi}(\vx_T,\vx_{T-1}|\vx_0) = q_{\vphi}(\vx_T|\vx_{T-1},\vx_0) q_{\vphi}(\vx_{T-1}|\vx_0)$. Since $q_{\vphi}$ is Markovian, we can further write $q_{\vphi}(\vx_T|\vx_{T-1},\vx_0) = q_{\vphi}(\vx_T|\vx_{T-1})$. Therefore, the joint expectation $\E_{q_{\vphi}(\vx_T,\vx_{T-1}|\vx_0)}$ can be written as a product of two expectations $\E_{q_{\vphi}(\vx_{T-1}|\vx_0)}\E_{q_{\vphi}(\vx_{T}|\vx_{T-1})}$. This will give us
\begin{align*}
\E_{q_{\vphi}(\vx_T,\vx_{T-1}|\vx_0)} \left[ \log \frac{p(\vx_T)}{q_{\vphi}(\vx_T | \vx_{T-1})} \right]
&= \E_{q_{\vphi}(\vx_{T-1}|\vx_0)} \left\{ \E_{q_{\vphi}(\vx_{T}|\vx_{T-1})} \left[ \log \frac{p(\vx_T)}{q_{\vphi}(\vx_T | \vx_{T-1})} \right] \right\}\\
&=-\E_{q_{\vphi}(\vx_{T-1}|\vx_0)} \Bigg[ \mathbb{D}_{\text{KL}} \left( q_{\vphi}(\vx_T | \vx_{T-1}) \|  p(\vx_T)  \right) \Bigg].
\end{align*}

Finally, we look at the product term. We can show that
\begin{align*}
\E_{q_{\vphi}(\vx_{1:T}|\vx_0)} \left[ \log \prod_{t=1}^{T-1} \frac{p(\vx_{t}|\vx_{t+1})}{q_{\vphi}(\vx_t | \vx_{t-1})} \right]
&= \sum_{t=1}^{T-1} \E_{q_{\vphi}(\vx_{1:T}|\vx_0)}  \left[ \log \frac{p(\vx_{t}|\vx_{t+1})}{q_{\vphi}(\vx_t | \vx_{t-1})} \right] \\
&= \sum_{t=1}^{T-1} \E_{q_{\vphi}(\vx_{t-1},\vx_t,\vx_{t+1}|\vx_0)}  \left[ \log \frac{p(\vx_{t}|\vx_{t+1})}{q_{\vphi}(\vx_t | \vx_{t-1})} \right],
\end{align*}
where again we use the fact the expectation only needs $\vx_{t-1}$, $\vx_t$, and $\vx_{t+1}$. Then, by using the same conditional independence argument, we can show that
\begin{align*}
\sum_{t=1}^{T-1} \E_{q_{\vphi}(\vx_{t-1},\vx_t,\vx_{t+1}|\vx_0)}  \left[ \log \frac{p(\vx_{t}|\vx_{t+1})}{q_{\vphi}(\vx_t | \vx_{t-1})} \right]
&= \sum_{t=1}^{T-1} \E_{q_{\vphi}(\vx_{t-1},\vx_{t+1}|\vx_0)} \left\{ \E_{q_{\vphi}(\vx_t|\vx_0)} \left[ \log \frac{p(\vx_{t}|\vx_{t+1})}{q_{\vphi}(\vx_t | \vx_{t-1})} \right] \right\}\\
&=
-\sum_{t=1}^{T-1} \E_{q_{\vphi}(\vx_{t-1},\vx_{t+1}|\vx_0)} \Bigg[ \mathbb{D}_{\text{KL}}\left( q_{\vphi}(\vx_{t}|\vx_{t-1})\| p(\vx_t|\vx_{t+1})\right)\Bigg].
\end{align*}
By replacing $p(\vx_0|\vx_1)$ with $p_{\vtheta}(\vx_0|\vx_1)$ and $p(\vx_{t}|\vx_{t+1})$ with $p_{\vtheta}(\vx_t|\vx_{t+1})$, we are done.
}

\textbf{Rewrite the Consistency Term}. The nightmare of the above variational diffusion model is that we need to draw samples $(\vx_{t-1},\vx_{t+1})$ from a joint distribution $q_{\vphi}(\vx_{t-1},\vx_{t+1}|\vx_0)$. We don't know what $q_{\vphi}(\vx_{t-1},\vx_{t+1}|\vx_0)$ is! It is a Gaussian by our choice, but we still need to use future samples $\vx_{t+1}$ to draw the current sample $\vx_t$. This is odd.

Inspecting the consistency term, we notice that $q_{\vphi}(\vx_t|\vx_{t-1})$ and $p_{\vtheta}(\vx_t|\vx_{t+1})$ are moving along two opposite directions. Thus, it is unavoidable that we need to use $\vx_{t-1}$ and $\vx_{t+1}$. The question we need to ask is: Can we come up with something so that we do not need to handle two opposite directions while we are able to check consistency?

\begin{figure}[h]
\centering
\includegraphics[height=4cm]{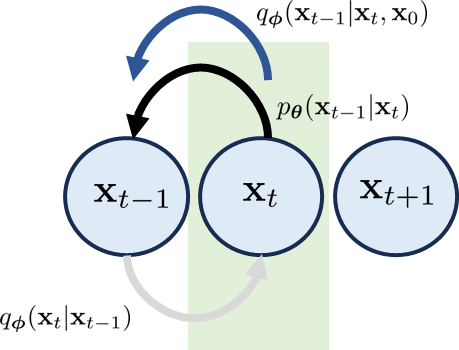}
\caption{If we consider the Bayes theorem in \eref{eq: diffusion rewrite q via Bayes}, we can define a distribution $q_{\vphi}(\vx_{t-1}|\vx_t,\vx_0)$ that has a direction parallel to $p_{\vtheta}(\vx_{t-1}|\vx_t)$.}
\label{fig: VDM q parallel to p}
\end{figure}

So, here is the simple trick called Bayes theorem which will give us
\begin{equation}
q(\vx_t|\vx_{t-1}) = \frac{q(\vx_{t-1}|\vx_t) q(\vx_t)}{q(\vx_{t-1})}
\quad \overset{\text{\scriptsize{condition on $\vx_0$}}}{\Longrightarrow} \quad
q(\vx_t|\vx_{t-1},\textcolor{blue}{\vx_0}) = \frac{q(\vx_{t-1}|\vx_t,\textcolor{blue}{\vx_0}) q(\vx_t|\textcolor{blue}{\vx_0})}{q(\vx_{t-1}|\textcolor{blue}{\vx_0})}.
\label{eq: diffusion rewrite q via Bayes}
\end{equation}
With this change of the conditioning order, we can switch $q(\vx_t|\vx_{t-1},\vx_0)$ to $q(\vx_{t-1}|\vx_t,\vx_0)$ by adding one more condition variable $\vx_0$. (If you do not condition on $\vx_0$, there is no way that we can draw samples from $q(\vx_{t-1})$, for example, because the specific state of $\vx_{t-1}$ depends on the initial image $\vx_0$.) The direction $q(\vx_{t-1}|\vx_t,\vx_0)$ is now parallel to $p_{\vtheta}(\vx_{t-1}|\vx_{t})$ as shown in \fref{fig: VDM q parallel to p}. So, if we want to rewrite the consistency term, a natural option is to calculate the KL divergence between $q_{\vphi}(\vx_{t-1}|\vx_t,\vx_0)$ and $p_{\vtheta}(\vx_{t-1}|\vx_{t})$.

If we manage to go through a few (boring) algebraic derivations, we can show that the ELBO is now:
\boxedthm{
\label{thm: ELBO for VDM 2}
\textbf{(ELBO for Variational Diffusion Model)}. Let $\vx = \vx_0$, and $\vx_T \sim \calN(0,\mI)$. The ELBO for a variational diffusion model in Theorem~\ref{thm: ELBO for VDM} can be equivalently written as
\begin{align}
\text{ELBO}_{\vphi,\vtheta}(\vx)
&=
\E_{q_{\vphi}(\vx_1|\vx_0)}[\log
\underset{\text{same as before}}{\underbrace{p_{\vtheta}(\vx_0|\vx_1)}}
]  - \underset{\text{new prior matching}}{\underbrace{\mathbb{D}_{\text{KL}}\Big(q_{\vphi}(\vx_T|\vx_0) \| p(\vx_T) \Big)}} \notag\\
&\qquad -\sum_{t=2}^{T} \E_{q_{\vphi}(\vx_{t}|\vx_0)} \Big[
\underset{\text{new consistency}}{\underbrace{\mathbb{D}_{\text{KL}}\Big(q_{\vphi}(\vx_{t-1}|\vx_t,\vx_0) \| p_{\vtheta}(\vx_{t-1}|\vx_{t}) \Big) }}
\Big].
\label{eq: ELBO second VDM}
\end{align}
}
Let's quickly make three interpretations:
\begin{itemize}
\setlength\itemsep{-1ex}
\item \textbf{Reconstruction}. The new reconstruction term is the same as before. We are still maximizing the log-likelihood.
\item \textbf{Prior Matching}. The new prior matching is simplified to the KL divergence between $q_{\vphi}(\vx_T|\vx_0)$ and $p(\vx_T)$. The change is due to the fact that we now condition upon $\vx_0$. Thus, there is no need to draw samples from $q_{\vphi}(\vx_{T-1}|\vx_0)$ and take expectation.
\item \textbf{Consistency}. The new consistency term is different from the previous one in two ways. Firstly, the running index $t$ starts at $t=2$ and ends at $t=T$. Previously it was from $t = 1$ to $t = T-1$. Accompanied by this is the distribution matching, which is now between $q_{\vphi}(\vx_{t-1}|\vx_t,\vx_0)$ and $p_{\vtheta}(\vx_{t-1}|\vx_{t})$. So, instead of asking a forward transition to match with a reverse transition, we use $q_{\vphi}$ to construct a reverse transition and use it to match with $p_{\vtheta}$.
\end{itemize}

\boxedproof{
\textbf{Proof of Theorem~\ref{thm: ELBO for VDM 2}}. We begin with \eref{eq: ELBO VDM step inequality} by showing that
\begin{alignat}{2}
\log p(\vx)
&\ge \E_{q_{\vphi}(\vx_{1:T}|\vx_0)}\left[ \log \frac{p(\vx_{0:T})}{q_{\vphi}(\vx_{1:T}|\vx_0)}\right] &\qquad
& \textcolor{purple}{(\text{By \eref{eq: ELBO VDM step inequality}})} \notag \\
&= \E_{q_{\vphi}(\vx_{1:T}|\vx_0)} \left[ \log \frac{p(\vx_T)p(\vx_0|\vx_1) \prod_{t=2}^T p(\vx_{t-1}|\vx_t)}{q_{\vphi}(\vx_1|\vx_0) \prod_{t=2}^T q_{\vphi}(\vx_t|\vx_{t-1},\vx_0) }  \right] &\qquad
& \textcolor{purple}{(\text{split the chain})} \notag \\
&= \E_{q_{\vphi}(\vx_{1:T}|\vx_0)} \left[ \log \frac{p(\vx_T)p(\vx_0|\vx_1) }{q_{\vphi}(\vx_1|\vx_0) } \right]
+
\E_{q_{\vphi}(\vx_{1:T}|\vx_0)} \left[ \log \prod_{t=2}^T  \frac{ p(\vx_{t-1}|\vx_t) }{ q_{\vphi}(\vx_t|\vx_{t-1},\vx_0) } \right]
&\qquad& \label{eq: VDM second proof, split expectation}
\end{alignat}
Let's consider the second term:
\begin{alignat*}{2}
\prod_{t=2}^T  \frac{ p(\vx_{t-1}|\vx_t) }{ q_{\vphi}(\vx_t|\vx_{t-1},\vx_0) }
&= \prod_{t=2}^T  \frac{ p(\vx_{t-1}|\vx_t) }{
\frac{q_{\vphi}(\vx_{t-1}|\vx_{t},\vx_0)q_{\vphi}(\vx_t|\vx_0)}{q_{\vphi}(\vx_{t-1}|\vx_0)} }   &\quad& \textcolor{purple}{(\text{Bayes rule, \eref{eq: diffusion rewrite q via Bayes}})} \\
&= \prod_{t=2}^T  \frac{ p(\vx_{t-1}|\vx_t) }{ q_{\vphi}(\vx_{t-1}|\vx_{t},\vx_0) } \times
\prod_{t=2}^T \frac{q_{\vphi}(\vx_{t-1}|\vx_0)} {q_{\vphi}(\vx_t|\vx_0)}                        &\quad& \textcolor{purple}{(\text{ Rearrange denominator})} \\
&= \prod_{t=2}^T  \frac{ p(\vx_{t-1}|\vx_t) }{ q_{\vphi}(\vx_{t-1}|\vx_{t},\vx_0) } \times
\frac{q_{\vphi}(\vx_{1}|\vx_0)} {q_{\vphi}(\vx_T|\vx_0)},                                        &\quad& \textcolor{purple}{(\text{ Recursion cancels terms})}
\end{alignat*}
where the last equation uses the fact that for any sequence $a_1,\ldots,a_T$, we have $\prod_{t=2}^T \frac{a_{t-1}}{a_{t}} = \frac{a_1}{a_2} \times \frac{a_2}{a_3} \times \ldots \times \frac{a_{T-1}}{a_T} = \frac{a_1}{a_T}$. Going back to the \eref{eq: VDM second proof, split expectation}, we can see that
\begin{alignat*}{2}
&\E_{q_{\vphi}(\vx_{1:T}|\vx_0)} \left[ \log \frac{p(\vx_T)p(\vx_0|\vx_1) }{q_{\vphi}(\vx_1|\vx_0) } \right]
+\E_{q_{\vphi}(\vx_{1:T}|\vx_0)} \left[ \log \prod_{t=2}^T  \frac{ p(\vx_{t-1}|\vx_t) }{ q_{\vphi}(\vx_t|\vx_{t-1},\vx_0) } \right] &\quad & \\
&=\E_{q_{\vphi}(\vx_{1:T}|\vx_0)} \left[ \log \frac{p(\vx_T)p(\vx_0|\vx_1) }{q_{\vphi}(\vx_1|\vx_0) } +
 \log \frac{q_{\vphi}(\vx_{1}|\vx_0)} {q_{\vphi}(\vx_T|\vx_0)}  \right]
+\E_{q_{\vphi}(\vx_{1:T}|\vx_0)} \left[ \log \prod_{t=2}^T  \frac{ p(\vx_{t-1}|\vx_t) }{ q_{\vphi}(\vx_{t-1}|\vx_{t},\vx_0) } \right]
&\quad & \\
&= \E_{q_{\vphi}(\vx_{1:T}|\vx_0)} \left[ \log \frac{p(\vx_T)p(\vx_0|\vx_1) }{q_{\vphi}(\vx_T|\vx_0) } \right]
+\E_{q_{\vphi}(\vx_{1:T}|\vx_0)} \left[ \log \prod_{t=2}^T  \frac{ p(\vx_{t-1}|\vx_t) }{ q_{\vphi}(\vx_{t-1}|\vx_{t},\vx_0) } \right],
&\quad &
\end{alignat*}
where we canceled $q_{\vphi}(\vx_1|\vx_0)$ in the numerator and denominator since $\log \frac{a}{b} + \log \frac{b}{c} = \log \frac{a}{c}$ for any positive constants $a$, $b$, and $c$. This will give us
\begin{align*}
\E_{q_{\vphi}(\vx_{1:T}|\vx_0)} \left[ \log \frac{p(\vx_T)p(\vx_0|\vx_1) }{q_{\vphi}(\vx_T|\vx_0) } \right]
&= \E_{q_{\vphi}(\vx_{1:T}|\vx_0)} \left[ \log p(\vx_0|\vx_1) \right] + \E_{q_{\vphi}(\vx_{1:T}|\vx_0)} \left[\log \frac{p(\vx_T) }{q_{\vphi}(\vx_T|\vx_0) } \right]\\
&=
\underset{\text{reconstruction}}{\underbrace{\E_{q_{\vphi}(\vx_{1}|\vx_0)} \left[ \log p(\vx_0|\vx_1) \right] }}  - \underset{\text{prior matching}}{\underbrace{\mathbb{D}_{\text{KL}}(q_{\vphi}(\vx_T|\vx_0)\|p(\vx_T))}}.
\end{align*}
The last term is
\begin{align*}
\E_{q_{\vphi}(\vx_{1:T}|\vx_0)} \left[ \log \prod_{t=2}^T  \frac{ p(\vx_{t-1}|\vx_t) }{ q_{\vphi}(\vx_{t-1}|\vx_{t},\vx_0) } \right]
&= \sum_{t=2} \E_{q_{\vphi}(\vx_t,\vx_{t-1}|\vx_0)} \log \frac{ p(\vx_{t-1}|\vx_t) }{ q_{\vphi}(\vx_{t-1}|\vx_{t},\vx_0) } \\
&\hspace{-8ex}=
\sum_{t=2} \iint \log \frac{ p(\vx_{t-1}|\vx_t) }{ q_{\vphi}(\vx_{t-1}|\vx_{t},\vx_0) } \cdot q_{\vphi}(\vx_t,\vx_{t-1}|\vx_0) d\vx_{t-1} d\vx_t\\
&\hspace{-8ex}=
\sum_{t=2} \iint \log \frac{ p(\vx_{t-1}|\vx_t) }{ q_{\vphi}(\vx_{t-1}|\vx_{t},\vx_0) } \cdot q_{\vphi}(\vx_{t-1}|\vx_t,\vx_0) q_{\vphi}(\vx_t|\vx_0) d\vx_{t-1} d\vx_t\\
&\hspace{-8ex}=
\sum_{t=2} \int \left\{\int \log \frac{ p(\vx_{t-1}|\vx_t) }{ q_{\vphi}(\vx_{t-1}|\vx_{t},\vx_0) } \cdot q_{\vphi}(\vx_{t-1},\vx_t|\vx_0) d\vx_{t-1} \right\} q_{\vphi}(\vx_t|\vx_0) d\vx_t\\
&\hspace{-8ex}=
-\underset{\text{consistency}}{\underbrace{\sum_{t=2} \E_{q_{\vphi}(\vx_t|\vx_0)} \mathbb{D}_{\text{KL}} ( q_{\vphi}(\vx_{t-1}|\vx_{t},\vx_0) \| p(\vx_{t-1}|\vx_t) )}}.
\end{align*}
Finally, replace $p(\vx_{t-1}|\vx_t)$ by $p_{\vtheta}(\vx_{t-1}|\vx_t)$, and $p(\vx_0|\vx_1)$ by $p_{\vtheta}(\vx_0|\vx_1)$. Done!
}

\subsection{Distribution of the Reverse Process} Now that we know the new ELBO for the variational diffusion model, we should spend some time discussing its core component which is $q_{\vphi}(\vx_{t-1}|\vx_t,\vx_0)$. In a nutshell, what we want to show is that
\begin{itemize}
\setlength\itemsep{-1ex}
\item $q_{\vphi}(\vx_{t-1}|\vx_t,\vx_0)$ is still a Gaussian.
\item Since it is a Gaussian, it is fully characterized by the mean and covariance. It turns out that
\begin{equation}
q_{\vphi}(\vx_{t-1}|\vx_t,\vx_0) = \calN(\vx_{t-1} \,|\, \heartsuit \vx_t + \spadesuit \vx_0, \clubsuit\mI),
\end{equation}
for some magical scalars $\heartsuit$, $\spadesuit$ and $\clubsuit$ defined below.
\end{itemize}

\boxedthm{
\label{thm: DDPM q xt-1 xt}
The distribution $q_{\vphi}(\vx_{t-1}|\vx_t,\vx_0)$ takes the form of
\begin{equation}
q_{\vphi}(\vx_{t-1}|\vx_t,\vx_0) = \calN(\vx_{t-1} \,|\, \vmu_q(\vx_t,\vx_0), \mSigma_q(t)),
\label{eq: VDM q t-1 | t, 0}
\end{equation}
where
\begin{align}
\vmu_q(\vx_t,\vx_0)    &= \frac{(1-\overline{\alpha}_{t-1})\sqrt{\alpha_t}}{1-\overline{\alpha}_t}\vx_t + \frac{(1-\alpha_t)\sqrt{\overline{\alpha}_{t-1}}}{1-\overline{\alpha}_t}\vx_0 \label{eq: VDM mu q}\\
\mSigma_q(t)           &= \frac{(1-\alpha_t)(1-\sqrt{\overline{\alpha}_{t-1}})}{1-\overline{\alpha}_t}\mI \bydef \sigma_q^2(t)\mI, \label{eq: VDM sigma q}
\end{align}
where $\overline{\alpha}_t = \prod_{i=1}^t \alpha_i$.
}

\begin{figure}[h]
\centering
\includegraphics[width=0.33\linewidth]{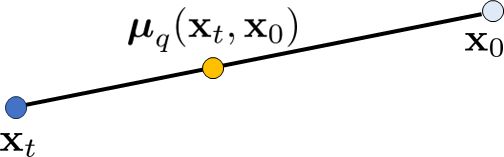}
\caption{According to \eref{eq: VDM mu q}, the mean $\vmu_q(\vx_t,\vx_0)$ is a linear combination of $\vx_t$ and $\vx_0$.}
\label{fig: xt line}
\end{figure}

\eref{eq: VDM mu q} reveals an interesting fact that the mean $\vmu_q(\vx_t,\vx_0)$ is a linear combination of $\vx_t$ and $\vx_0$. Geometrically, $\vmu_q(\vx_t,\vx_0)$ lives on the straight line connecting $\vx_t$ and $\vx_0$, as illustrated in \fref{fig: xt line}.

\boxedproof{
\textbf{Proof of Theorem~\ref{thm: DDPM q xt-1 xt}}. Using the Bayes theorem stated in \eref{eq: diffusion rewrite q via Bayes}, $q(\vx_{t-1}|\vx_t,\vx_0)$ can be determined if we evaluate the following product of Gaussians
\begin{equation}
q(\vx_{t-1}|\vx_t,\vx_0)
= \frac{\calN( \vx_t | \sqrt{\alpha_t}\vx_{t-1}, (1-\alpha_t)\mI) \calN( \vx_{t-1} | \sqrt{\overline{\alpha}_{t-1}}\vx_0, (1-\overline{\alpha}_{t-1}\mI))}
{\calN(\vx_t | \sqrt{\overline{\alpha}_t}\vx_0, (1-\overline{\alpha}_t)\mI)}.
\end{equation}
For simplicity we will treat the vectors are scalars. Then the above product of Gaussians will become
\begin{equation}
q(\vx_{t-1}|\vx_t,\vx_0) \propto
\exp\left\{
\frac{( \vx_t-\sqrt{\alpha_t}\vx_{t-1})^2}{2(1-\alpha_t)}
+ \frac{(\vx_{t-1}-\sqrt{\overline{\alpha}_{t-1}}\vx_0)^2}{2(1-\overline{\alpha}_{t-1})}
- \frac{(\vx_t-\sqrt{\overline{\alpha}_t}\vx_0)^2}{2(1-\overline{\alpha}_t)}
\right\}.
\end{equation}
We consider the following mapping:
\begin{alignat*}{2}
x &= \vx_t,              &\quad & a = \alpha_t\\
y &= \vx_{t-1},          &\quad & b = \overline{\alpha}_{t-1}\\
z &= \vx_{0},            &\quad & c = \overline{\alpha}_t.
\end{alignat*}
Consider a quadratic function
\begin{equation}
f(y) = \frac{(x-\sqrt{a}y)^2}{2(1-a)} + \frac{(y-\sqrt{b}z)^2}{2(1-b)} - \frac{(x-\sqrt{c}z)^2}{2(1-c)}.
\end{equation}
We know that no matter how we rearrange the terms, the resulting function remains a quadratic equation. The minimizer of $f(y)$ is the mean of the resulting Gaussian. So, we can calculate the derivative of $f$ and show that
\begin{align*}
f'(y) = \frac{1-ab}{(1-a)(1-b)}y - \left(\frac{\sqrt{a}}{1-a}x + \frac{\sqrt{b}}{1-b}z\right).
\end{align*}
Setting $f'(y) = 0$ yields
\begin{equation}
y = \frac{(1-b)\sqrt{a}}{1-ab}x + \frac{(1-a)\sqrt{b}}{1-ab}z.
\end{equation}
We note that $ab = \alpha_t \overline{\alpha}_{t-1} = \overline{\alpha}_t$. So,
\begin{equation}
\vmu_q(\vx_t,\vx_0) = \frac{(1-\overline{\alpha}_{t-1})\sqrt{\alpha_t}}{1-\overline{\alpha}_t}\vx_t + \frac{(1-\alpha_t)\sqrt{\overline{\alpha}_{t-1}}}{1-\overline{\alpha}_t}\vx_0.
\end{equation}
Similarly, for the variance, we can check the curvature $f''(y)$. We can easily show that
\begin{equation*}
f''(y) = \frac{1-ab}{(1-a)(1-b)} = \frac{1-\overline{\alpha}_t}{(1-\alpha_t)(1-\overline{\alpha}_{t-1})}.
\end{equation*}
Taking the reciprocal will give us
\begin{equation}\mSigma_q(t) = \frac{(1-\alpha_t)(1-\overline{\alpha}_{t-1})}{1-\overline{\alpha}_t}\mI.
\end{equation}
}

\begin{figure}[h]
\centering
\includegraphics[width=0.4\linewidth]{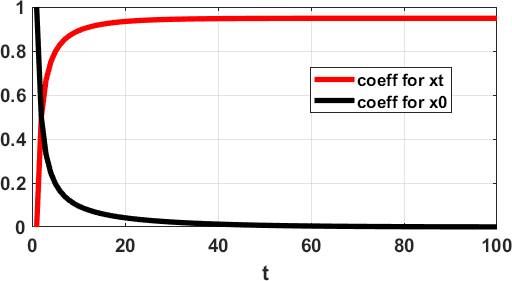}
\caption{The trajectory of the coefficients for $\vx_t$ and for $\vx_0$, as $t$ grows.}
\label{fig: Fig14_coeff}
\end{figure}

In combination weight in the above theorem deserves some study. Recall that
\begin{equation*}
\vmu_q(\vx_t,\vx_0) = \frac{(1-\overline{\alpha}_{t-1})\sqrt{\alpha_t}}{1-\overline{\alpha}_t}\vx_t + \frac{(1-\alpha_t)\sqrt{\overline{\alpha}_{t-1}}}{1-\overline{\alpha}_t}\vx_0.
\end{equation*}
One question we can ask is how does the two coefficients behave as $t$ goes from $T$ to $1$? We show an example in \fref{fig: Fig14_coeff}. For this particular example, we use $\alpha_t = 0.9$ for all $t$. We plot the coefficients as a function of $t$. \fref{fig: Fig14_coeff} suggests that the coefficient for $\vx_t$ shrinks as $t$ decreases from $t = T$ to $t = 1$, whereas the coefficient for $\vx_0$ grows as $t$ decreases.

\begin{figure}[h]
\centering
\includegraphics[width=0.35\linewidth]{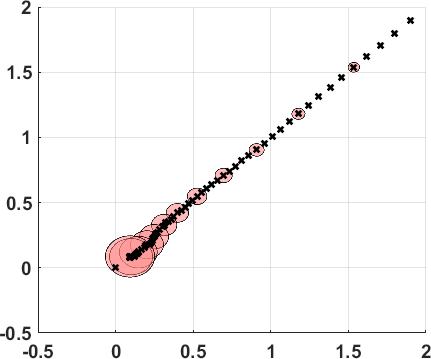}
\caption{The trajectory of $\vx_t$ and the associated radius of the Gaussian $\sigma_q^2(t)$.}
\label{fig: Fig14_muq trajectory}
\end{figure}

As $t$ goes from $T$ to $1$, the variance $\sigma_q^2(t)$ will also change. \fref{fig: Fig14_muq trajectory} shows the trajectory of $\vx_t$ as a function of $t$ by sampling $\vx_t$ according to $q_{\vphi}(\vx_{t-1}|\vx_t,\vx_0)$. On the same plot, we show the radius of the Gaussian, defined by $\sigma_q^2(t)$. Our plot indicates that when $t = T$, the variance $\sigma_q^2(t)$ is fairly large so that $\vx_t$ is closer to white Gaussian noise. As $t$ drops to $t = 1$, the variance $\sigma_q^2(t)$ also drops to zero. This makes sense because eventually we want $\vx_0$ to be the clean image which is noise free.

\vspace{4ex}
\textbf{Constructing $p_{\vtheta}(\vx_{t-1}|\vx_t)$}. The interesting part of \eref{eq: VDM q t-1 | t, 0} is that $q_{\vphi}(\vx_{t-1}|\vx_t,\vx_0)$ is \emph{completely characterized} by $\vx_t$ and $\vx_0$. There is no neural network required to estimate the mean and variance! (You can compare this with VAE where a network is needed.) Since a network is not needed, there is really nothing to ``learn''. The distribution $q_{\vphi}(\vx_{t-1}|\vx_t,\vx_0)$ is automatically determined if we know $\vx_t$ and $\vx_0$.

The realization here is important. Let's look at the consistency term in \eref{eq: ELBO second VDM}:
\begin{align*}
\text{ELBO}_{\vphi,\vtheta}(\vx)
&=
\E_{q_{\vphi}(\vx_1|\vx_0)}[\log
\underset{\text{same as before}}{\underbrace{p_{\vtheta}(\vx_0|\vx_1)}}
]  - \underset{\text{new prior matching}}{\underbrace{\mathbb{D}_{\text{KL}}\Big(q_{\vphi}(\vx_T|\vx_0) \| p(\vx_T) \Big)}} \notag\\
&\qquad\qquad -\sum_{t=2}^{T} \E_{q_{\vphi}(\vx_{t}|\vx_0)} \Big[
\underset{\text{new consistency}}{\underbrace{\mathbb{D}_{\text{KL}}\Big(q_{\vphi}(\vx_{t-1}|\vx_t,\vx_0) \| p_{\vtheta}(\vx_{t-1}|\vx_{t}) \Big) }}
\Big], \qquad \qquad \textcolor{purple}{(\text{from \eref{eq: ELBO second VDM}})}
\end{align*}
where we just showed that
\begin{equation*}
q_{\vphi}(\vx_{t-1}|\vx_t,\vx_0) = \calN(\vx_{t-1} \,|\, \vmu_q(\vx_t,\vx_0), \mSigma_q(t)).
\end{equation*}
There is no ``learning'' for $q_{\vphi}(\vx_{t-1}|\vx_{t},\vx_0)$ because it is defined once the hyperparameter $\alpha_t$ are defined. Therefore, the consistency term is a summation of many KL divergence terms where the $t$-th term is
\begin{equation}
\mathbb{D}_{\text{KL}} (
\underset{\quad \text{nothing to learn} \quad }{\underbrace{ q_{\vphi}(\vx_{t-1}|\vx_{t},\vx_0) }} \|
\underset{\text{need to do something}}{\underbrace{ p_{\vtheta}(\vx_{t-1}|\vx_t) }} ).
\end{equation}
So, to compute the KL divergence, we need to do something about $p_{\vtheta}(\vx_{t-1}|\vx_t)$.

The big idea here is that $q_{\vphi}(\vx_{t-1}|\vx_{t},\vx_0)$ is Gaussian. If we want to quickly calculate the KL divergence, then it would be good if $p_{\vtheta}(\vx_{t-1}|\vx_t)$ is also a Gaussian. So, we \emph{choose} $p_{\vtheta}(\vx_{t-1}|\vx_t)$ to be a Gaussian. Moreover, we should match the form of the mean and variance! Therefore, we define
\begin{equation}
p_{\vtheta}(\vx_{t-1}|\vx_t) = \calN\Big(\vx_{t-1}| \underset{\text{neural network}}{\underbrace{\vmu_{\vtheta}(\vx_t)}}, \sigma_q^2(t)\mI\Big),
\end{equation}
where we assume that the mean vector can be determined using a neural network. As for the variance, we \emph{choose} the variance to be $\sigma_q^2(t)$. This is \emph{identical} to \eref{eq: VDM sigma q}! Thus, if we put \eref{eq: VDM q t-1 | t, 0} side by side with $p_{\vtheta}(\vx_{t-1}|\vx_t)$, we notice a parallel relation between the two:
\begin{align}
q_{\vphi}(\vx_{t-1}|\vx_t,\vx_0) &= \calN\Big(\vx_{t-1} \,|\, \underset{\text{known}}{\underbrace{\vmu_q(\vx_t,\vx_0)}},
\underset{\text{known}}{\underbrace{\sigma_q^2(t)\mI}}\Big),\\
p_{\vtheta}(\vx_{t-1}|\vx_t)     &= \calN\Big(\vx_{t-1}| \underset{\text{neural network}}{\underbrace{\vmu_{\vtheta}(\vx_t)}},
\underset{\text{known}}{\underbrace{\sigma_q^2(t)\mI}}\Big).
\end{align}
Therefore, the KL divergence is simplified to
\begin{align}
& \mathbb{D}_{\text{KL}}\Big( q_{\vphi}(\vx_{t-1}|\vx_t,\vx_0) \;\|\; p_{\vtheta}(\vx_{t-1}|\vx_t) \Big) \notag \\
&=\mathbb{D}_{\text{KL}}\Big( \calN(\vx_{t-1} \,|\, \vmu_q(\vx_t,\vx_0), \sigma_q^2(t)\mI) \;\|\; \calN(\vx_{t-1} \,|\, \vmu_{\vtheta}(\vx_t), \sigma_q^2(t)\mI) \Big) \notag\\
&= \frac{1}{2\sigma_q^2(t)}\|\vmu_q(\vx_t,\vx_0) - \vmu_{\vtheta}(\vx_t)\|^2,
\label{eq: VDM ELBO KL divergence}
\end{align}
where we used the fact that the KL divergence between two identical-variance Gaussians is just the Euclidean distance square between the two mean vectors.

Substituting \eref{eq: VDM ELBO KL divergence} to the definition of ELBO in \eref{eq: ELBO second VDM}, we can rewrite ELBO as follows.
\boxedthm{
\label{thm: ELBO VDM final}
The \textbf{ELBO for a variational diffusion model} in \eref{eq: ELBO second VDM} can be simplified to
\begin{align}
\text{ELBO}_{\vtheta}(\vx)
&=
\E_{q(\vx_1|\vx_0)}[\log p_{\vtheta}(\vx_0|\vx_1)]
-
\underset{\text{nothing to train}}{\underbrace{\mathbb{D}_{\text{KL}}\Big(q(\vx_T|\vx_0) \| p(\vx_T) \Big)}} \notag\\
&\qquad -\sum_{t=2}^{T} \E_{q(\vx_{t}|\vx_0)} \Big[ \frac{1}{2\sigma_q^2(t)} \|\vmu_q(\vx_t,\vx_0) - \vmu_{\vtheta}(\vx_t)\|^2 \Big],
\label{eq: ELBO third VDM muq mutheta}
\end{align}
where $\vx = \vx_0$, and $\vx_T \sim \calN(0,\mI)$.
}

One remark for Theorem~\ref{thm: ELBO VDM final} is that the subscript $\vphi$ is dropped because the distribution $q_{\vphi}$ defined in Theorem~\ref{thm: DDPM q xt-1 xt} is fully characterized by $\vx_t$ and $\vx_0$. There is nothing to learn, and so the optimization does not need to include $\vphi$. Because of this, we can drop the KL-divergence term in \eref{eq: ELBO third VDM muq mutheta}. This leaves us the reconstruction term $\E_{q(\vx_1|\vx_0)}[\log p_{\vtheta}(\vx_0|\vx_1)]$ and transition term $\sum_{t=2}^{T} \E_{q(\vx_{t}|\vx_0)} \Big[ \frac{1}{2\sigma_q^2(t)} \|\vmu_q(\vx_t,\vx_0) - \vmu_{\vtheta}(\vx_t)\|^2 \Big]$.

The reconstruction term can be simplified, since
\begin{align*}
\log p_{\vtheta}(\vx_0|\vx_1)
&= \log \calN(\vx_0 | \vmu_{\vtheta}(\vx_1), \sigma_q^2(1)\mI)\\
&= \log \frac{1}{(\sqrt{2\pi\sigma_q^2(1)})^d} \exp\left\{-\frac{\|\vx_0 - \vmu_{\vtheta}(\vx_1)\|^2}{2\sigma_q^2(1)}\right\}\\
&= -\frac{\|\vx_0 - \vmu_{\vtheta}(\vx_1)\|^2}{2\sigma_q^2(1)} - \frac{d}{2}\log \left(2\pi\sigma_q^2(1)\right).
\end{align*}
So, as soon as we know $\vx_1$, we can send it to a network $\vmu_{\vtheta}(\vx_1)$ to return us a mean estimate. The mean estimate will then be used to compute the likelihood.

\subsection{Training and Inference}
In this section we discuss how to train a variational diffusion model, and turn into a \emph{denoising} diffusion probabilistic model.

We start by looking at the ELBO defined in Theorem~\ref{thm: ELBO VDM final}. \eref{eq: ELBO third VDM muq mutheta} suggests that we need to find a network $\vmu_{\vtheta}$ that can somehow minimize the loss:
\begin{equation}
\frac{1}{2\sigma_q^2(t)} \| \underset{\text{known}}{\underbrace{\vmu_q(\vx_t,\vx_0)}} - \underset{\text{network}}{\underbrace{\vmu_{\vtheta}(\vx_t)}}\|^2.\
\label{eq: denoising mu q minus mu theta}
\end{equation}
Recall from \eref{eq: VDM mu q} that
\begin{equation}
\vmu_q(\vx_t,\vx_0) = \frac{(1-\overline{\alpha}_{t-1})\sqrt{\alpha_t}}{1-\overline{\alpha}_t}\vx_t + \frac{(1-\alpha_t)\sqrt{\overline{\alpha}_{t-1}}}{1-\overline{\alpha}_t}\vx_0.
\label{eq: denoising mu q}
\end{equation}
We see that it is a function of $\vx_t$ and $\vx_0$. Therefore, $\vmu_q(\vx_t,\vx_0)$ is known and determined once we know $\vx_t$ and $\vx_0$.

The subject of interest is $\vmu_{\vtheta}$. Since $\vmu_{\vtheta}$ is our \emph{design}, there is no reason why we cannot define it as something more convenient. So here is an option. We define
\begin{equation}
\underset{\text{a network}}{\underbrace{\vmu_{\vtheta}(\vx_t)}}
\bydef \frac{(1-\overline{\alpha}_{t-1})\sqrt{\alpha_t}}{1-\overline{\alpha}_t}\vx_t + \frac{(1-\alpha_t)\sqrt{\overline{\alpha}_{t-1}}}{1-\overline{\alpha}_t}
\underset{\text{another network}}{\underbrace{\widehat{\vx}_{\vtheta}(\vx_t)}}.
\label{eq: denoising mu theta}
\end{equation}
Substituting \eref{eq: denoising mu q} and \eref{eq: denoising mu theta} into \eref{eq: denoising mu q minus mu theta} will give us
\begin{align}
\frac{1}{2\sigma_q^2(t)} \|\vmu_q(\vx_t,\vx_0) - \vmu_{\vtheta}(\vx_t) \|^2
&= \frac{1}{2\sigma_q^2(t)} \left\| \frac{(1-\alpha_t)\sqrt{\overline{\alpha}_{t-1}}}{1-\overline{\alpha}_t} (\widehat{\vx}_{\vtheta}(\vx_t)-\vx_0) \right\|^2 \notag\\
&= \frac{1}{2\sigma_q^2(t)} \frac{(1-\alpha_t)^2 \overline{\alpha}_{t-1}}{(1-\overline{\alpha}_t)^2}  \left\| \widehat{\vx}_{\vtheta}(\vx_t)-\vx_0 \right\|^2. \label{eq: denoising mu theta 2}
\end{align}
Therefore, ELBO can be written as
\begin{align}
\text{ELBO}_{\vtheta}(\vx)
&=
\E_{q(\vx_1|\vx_0)}[\log p_{\vtheta}(\vx_0|\vx_1)] -\sum_{t=2}^{T} \E_{q(\vx_{t}|\vx_0)} \Big[ \frac{1}{2\sigma_q^2(t)} \frac{(1-\alpha_t)^2 \overline{\alpha}_{t-1}}{(1-\overline{\alpha}_t)^2}  \left\| \widehat{\vx}_{\vtheta}(\vx_t)-\vx_0 \right\|^2 \Big],
\end{align}
where we dropped the term $\mathbb{D}_{\text{KL}}\Big(q(\vx_T|\vx_0) \| p(\vx_T) \Big)$.

Next, we want to simplify ELBO so that we can absorb $\E_{q(\vx_1|\vx_0)}[\log p_{\vtheta}(\vx_0|\vx_1)] $ into the summation. The following is the result.
\boxedthm{
The ELBO for \textbf{denoising diffusion probabilistic model} is
\begin{equation}
\text{ELBO}_{\vtheta}(\vx) = -\sum_{t=1}^{T} \frac{1}{2\sigma_q^2(t)} \frac{(1-\alpha_t)^2 \overline{\alpha}_{t-1}}{(1-\overline{\alpha}_t)^2}  \E_{q(\vx_{t}|\vx_0)} \Big[ \left\| \widehat{\vx}_{\vtheta}(\vx_t)-\vx_0 \right\|^2 \Big].
\label{eq: diffusion final loss}
\end{equation}
}

\boxedproof{
\textbf{Proof}. Substituting \eref{eq: denoising mu theta 2} into \eref{eq: ELBO third VDM muq mutheta}, we can see that
\begin{align}
\text{ELBO}_{\vtheta}(\vx)
&= \E_{q(\vx_1|\vx_0)}[\log p_{\vtheta}(\vx_0|\vx_1)] -\sum_{t=2}^{T} \E_{q(\vx_{t}|\vx_0)} \Big[ \frac{1}{2\sigma_q^2(t)} \|\vmu_q(\vx_t,\vx_0) - \vmu_{\vtheta}(\vx_t)\|^2 \Big] \notag \\
&= \E_{q(\vx_1|\vx_0)}[\log p_{\vtheta}(\vx_0|\vx_1)] -\sum_{t=2}^{T} \E_{q(\vx_{t}|\vx_0)} \Big[ \frac{1}{2\sigma_q^2(t)} \frac{(1-\alpha_t)^2 \overline{\alpha}_{t-1}}{(1-\overline{\alpha}_t)^2}  \left\| \widehat{\vx}_{\vtheta}(\vx_t)-\vx_0 \right\|^2 \Big].
\label{eq: ELBO diffusion, intermediate}
\end{align}

\vspace{-2ex}
The first term is
\begin{alignat}{2}
\log p_{\vtheta}(\vx_0|\vx_1)
&= \log \calN(\vx_{0} | \vmu_{\vtheta}(\vx_1), \sigma_q^2(1)\mI )
\propto -\frac{1}{2\sigma_q^2(1)}\| \vmu_{\vtheta}(\vx_1) - \vx_{0} \|^2       &\quad& \textcolor{purple}{(\text{definition})}        \notag \\
&= -\frac{1}{2\sigma_q^2(1)} \left\|
\frac{(1-\overline{\alpha}_{0})\sqrt{\alpha_1}}{1-\overline{\alpha}_1}\vx_1 + \frac{(1-\alpha_1)\sqrt{\overline{\alpha}_{0}}}{1-\overline{\alpha}_1}\widehat{\vx}_{\vtheta}(\vx_1) - \vx_0
\right\|^2                                                                      &\quad& \textcolor{purple}{(\text{recall $\alpha_0 = 1$})}        \notag \\
&= -\frac{1}{2\sigma_q^2(1)} \left\|\frac{(1-\alpha_1)}{1-\overline{\alpha}_1}\widehat{\vx}_{\vtheta}(\vx_1) - \vx_0\right\|^2 \notag \\
&= -\frac{1}{2\sigma_q^2(1)} \left\|\widehat{\vx}_{\vtheta}(\vx_1) - \vx_0\right\|^2. &\quad& \textcolor{purple}{(\text{recall $\overline{\alpha}_1 = \alpha_1$})}
\label{eq: ELBO diffusion first term}
\end{alignat}
Substituting \eref{eq: ELBO diffusion first term} into \eref{eq: ELBO diffusion, intermediate} will simplify ELBO as
\begin{align*}
\text{ELBO}_{\vtheta}(\vx) = -\sum_{t=1}^{T} \E_{q(\vx_{t}|\vx_0)} \Big[ \frac{1}{2\sigma_q^2(t)} \frac{(1-\alpha_t)^2 \overline{\alpha}_{t-1}}{(1-\overline{\alpha}_t)^2}  \left\| \widehat{\vx}_{\vtheta}(\vx_t)-\vx_0 \right\|^2 \Big].
\end{align*}
}

The loss function defined in \eref{eq: diffusion final loss} is very intuitive. Ignoring the constants and expectations, the main subject of interest, for a particular $\vx_t$, is
\begin{equation*}
\argmin{\vtheta} \;\; \left\| \widehat{\vx}_{\vtheta}(\vx_t)-\vx_0 \right\|^2.
\end{equation*}
This is nothing but a denoising problem because we need to find a network $\widehat{\vx}_{\vtheta}$ such that the denoised image $\widehat{\vx}_{\vtheta}(\vx_t)$ will be close to the ground truth $\vx_0$. What makes it not a typical denoiser is the following reasons:

\begin{itemize}
\itemsep\setlength{-1ex}
\item $\E_{q(\vx_{t}|\vx_0)}$: We are not trying to denoise any random noisy image. Instead, we are carefully choosing the noisy image to be
\begin{alignat*}{2}
                &\qquad &\vx_t \sim q(\vx_{t}|\vx_0) &= \calN(\vx_t \,|\, \sqrt{\overline{\alpha}_t} \vx_0, \;\; (1-\overline{\alpha}_t)\mI)\\
\Leftrightarrow &\qquad &\vx_t &= \sqrt{\overline{\alpha}_t} \vx_0 + \sqrt{(1-\overline{\alpha}_t)}\vepsilon_t, \qquad\text{where}\qquad \vepsilon_t \sim \calN(0,\mI).
\end{alignat*}
\item $\frac{1}{2\sigma_q^2(t)} \frac{(1-\alpha_t)^2 \overline{\alpha}_{t-1}}{(1-\overline{\alpha}_t)^2}$: We do not weight the denoising loss equally for all steps. Instead, there is a scheduler to control the relative emphasis on each denoising loss. Considering this, and using Monte Carlo to approximate the expectation, we can write the optimization problem as
\begin{align}
&\argmax{\vtheta} \;\; \sum_{\vx_0 \in \calX} \;\; \text{ELBO}(\vx_0) \notag \\
&=\argmin{\vtheta} \;\; \sum_{\vx_0 \in \calX} \;\;\sum_{t=1}^{T} \E_{q(\vx_{t}|\vx_0)} \left[ \frac{1}{2\sigma_q^2(t)} \frac{(1-\alpha_t)^2 \overline{\alpha}_{t-1}}{(1-\overline{\alpha}_t)^2} \left\| \widehat{\vx}_{\vtheta}\Big(\sqrt{\overline{\alpha}_t} \vx_0 + \sqrt{(1-\overline{\alpha}_t)}\vepsilon_t\Big)-\vx_0 \right\|^2 \right]\notag\\
&= \argmin{\vtheta} \;\; \sum_{\vx_0 \in \calX} \;\;\sum_{t=1}^T \frac{1}{M}\sum_{m=1}^M \frac{1}{2\sigma_q^2(t)} \frac{(1-\alpha_t)^2 \overline{\alpha}_{t-1}}{(1-\overline{\alpha}_t)^2} \left\| \widehat{\vx}_{\vtheta}\Big(\sqrt{\overline{\alpha}_t} \vx_0 + \sqrt{(1-\overline{\alpha}_t)}\vepsilon_t^{(m)}\Big)-\vx_0 \right\|^2,
\label{eq: DDPM training loss}
\end{align}
where $\vepsilon^{(m)}_t \sim \calN(0,\mI)$, and the summation over $\vx_0\in\calX$ means we consider all samples in the training set $\calX$. As you can see, the training of this model involves training a \emph{denoiser} $\widehat{\vx}_{\vtheta}(\cdot)$. For this reason, the resulting model is known as the \emph{denoising} diffusion probabilistic model (DDPM).
\end{itemize}

\begin{figure}[h]
\centering
\includegraphics[width=0.6\linewidth]{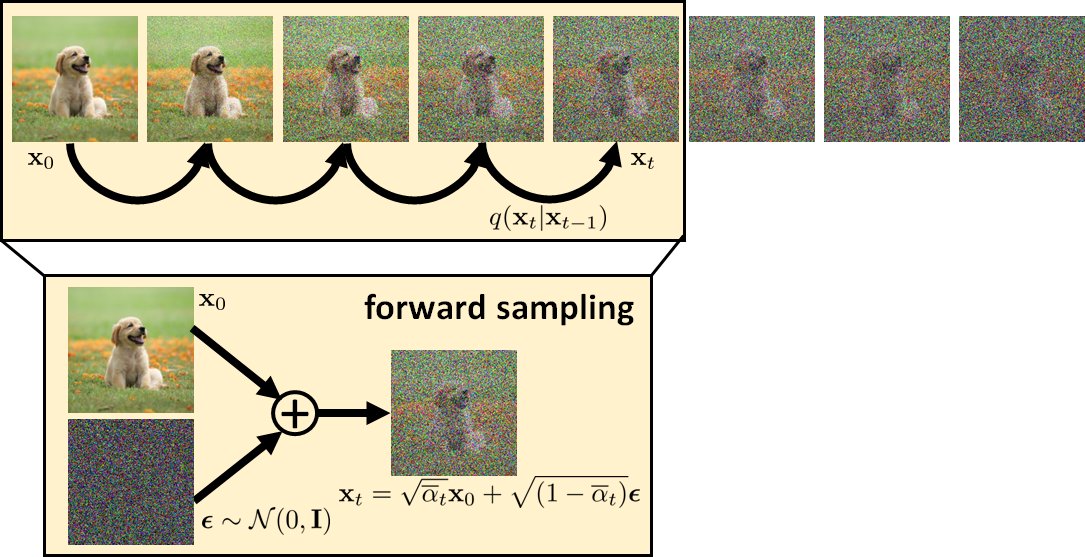}
\caption{Forward diffusion process.}
\label{fig: forward sampling}
\end{figure}

\vspace{1ex}
\textbf{Forward Diffusion in DDPM}. The training of a DDPM involves two parallel branches. The first branch is the forward diffusion. The goal of forward diffusion is to generate the intermediate variables $\vx_1,\ldots,\vx_{T-1}$ by using
\begin{align*}
\vx_t \sim q(\vx_{t}|\vx_0) = \calN(\vx_t \,|\, \sqrt{\overline{\alpha}_t} \vx_0, \;\; (1-\overline{\alpha}_t)\mI), \qquad t = 1,\ldots,T-1.
\end{align*}
The forward diffusion does not require any training. If you give us the clean image $\vx_0$, we can run the forward diffusion and prepare the images $\vx_1,\ldots,\vx_{T}$. A pictorial illustration is shown in \fref{fig: forward sampling}.

\begin{figure}[h]
\centering
\includegraphics[width=0.7\linewidth]{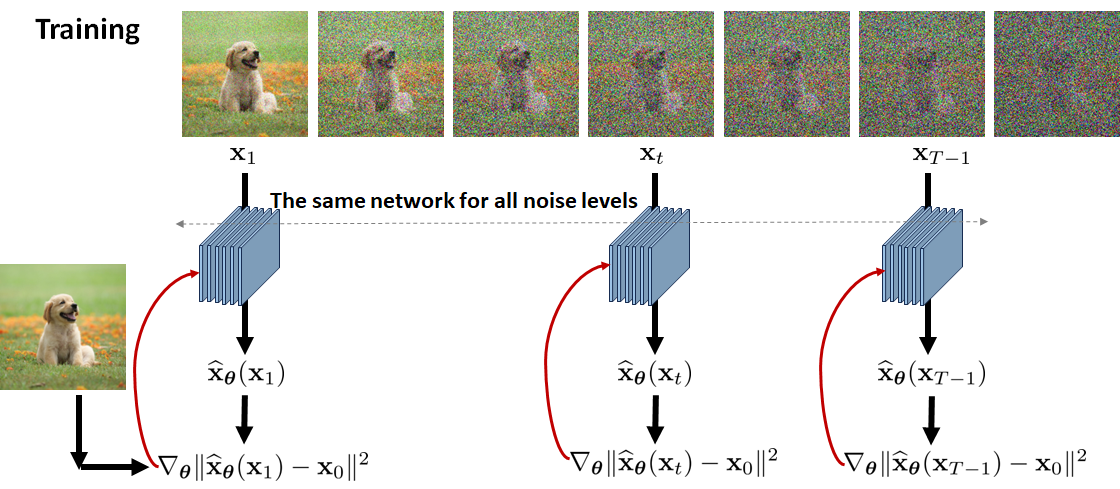}
\caption{Training of a denoising diffusion probabilistic model. For the same neural network $\widehat{\vx}_{\vtheta}$, we send noisy inputs $\vx_t$ to the network. The gradient of the loss is back-propagated to update the network. Note that the noisy images are not arbitrary. They are generated according to the forward sampling process.}
\label{fig: training DDPM}
\end{figure}

\vspace{2ex}
\textbf{Training DDPM}. Once the training samples $\vx_0,\ldots,\vx_T$ are prepared, we can train the DDPM. The training of DDPM is summarized by the optimization in \eref{eq: DDPM training loss} and \fref{fig: training DDPM}. The goal is to train \emph{one} denoiser for \emph{all} noise levels. It is a one denoiser for all noise levels because each $\vx_t$ has a different variance $1-\overline{\alpha}_t$. We are not interested in training many denoisers because it is computationally not feasible.

The training of a denoiser is no different than any conventional supervised learning. Given a pair of clean and noisy image, which in our case is $\vx_0$ and $\vx_t$, we train the denoiser $\widehat{\vx}_{\vtheta}(\cdot)$. The training loss in \eref{eq: DDPM training loss} has three summations. If we run stochastic gradient descent, we can simplify the above optimization into the following procedure.

\boxedmsg{
\textbf{Training Algorithm for DDPM}. For every image $\vx_0$ in your training dataset:
\begin{itemize}
\itemsep\setlength{-1ex}
\item Repeat the following steps until convergence.
\item Pick a random time stamp $t \sim \text{Uniform}[1,T]$.
\item Draw a sample $\vx_t^{(m)} \sim \calN(\vx_t \,|\, \sqrt{\overline{\alpha}_t} \vx_0, \;\; (1-\overline{\alpha}_t)\mI)$, i.e.,
\begin{equation*}
\vx_t^{(m)} = \overline{\alpha}_t \vx_0 +  \sqrt{(1-\overline{\alpha}_t)}\vepsilon_t^{(m)}, \qquad \vepsilon_t^{(m)} \sim \calN(0,\mI).
\end{equation*}
\item Take gradient descent step on
$$
\nabla_{\vtheta}  \left\{ \frac{1}{M} \sum_{m=1}^M  \left\| \widehat{\vx}_{\vtheta}(\vx_t^{(m)})-\vx_0 \right\|^2 \right\}.
$$
\end{itemize}
You can do this in batches, just like how you train any other neural networks. Note that, here, you are training \textbf{one} denoising network $\widehat{\vx}_{\vtheta}$ for \textbf{all} noisy conditions.
}

\begin{figure}[h]
\centering
\includegraphics[width=0.8\linewidth]{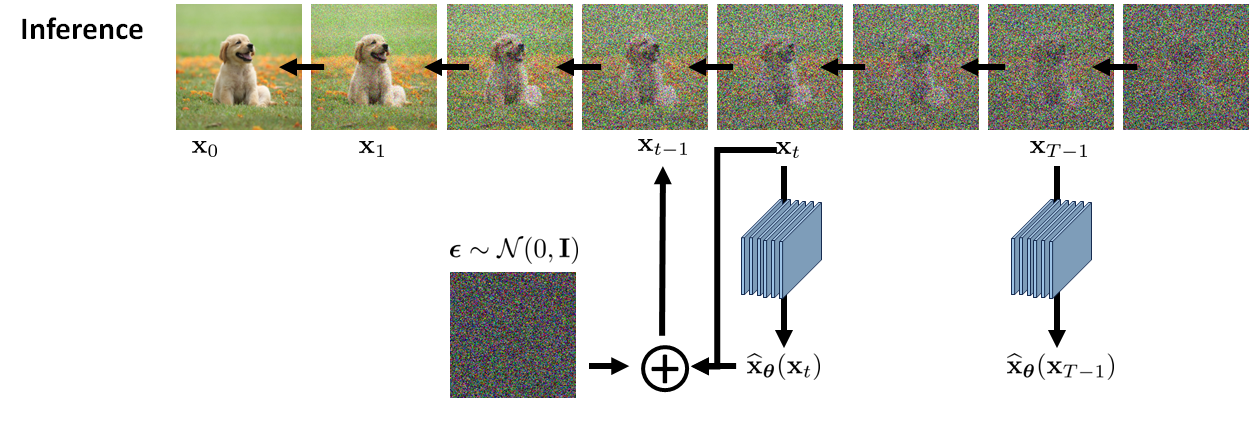}
\caption{Inference of a denoising diffusion probabilistic model.}
\label{fig: DDPM inference}
\end{figure}

\vspace{2ex}
\textbf{Inference of DDPM -- the Reverse Diffusion}. Once the denoiser $\widehat{\vx}_{\vtheta}$ is trained, we can apply it to do the inference. The inference is about sampling images from the distributions $p_{\vtheta}(\vx_{t-1}|\vx_t)$ over the sequence of states $\vx_{T},\vx_{T-1},\ldots,\vx_{1}$. Since it is the reverse diffusion process, we need to do it recursively via:
\begin{align*}
\vx_{t-1}
\sim p_{\vtheta}(\vx_{t-1} \,|\, \vx_t)
&= \calN(\vx_{t-1} \,|\, \vmu_{\vtheta}(\vx_t), \sigma_q^2(t)\mI).
\end{align*}
By reparameterization, we have
\begin{align*}
\vx_{t-1}
&= \vmu_{\vtheta}(\vx_t) + \sigma_q(t) \vepsilon, \qquad\qquad\qquad\text{where}\quad \vepsilon \sim \calN(0,\mI)\\
&= \frac{(1-\overline{\alpha}_{t-1})\sqrt{\alpha_t}}{1-\overline{\alpha}_t}\vx_t + \frac{(1-\alpha_t)\sqrt{\overline{\alpha}_{t-1}}}{1-\overline{\alpha}_t} \widehat{\vx}_{\vtheta}(\vx_t) +  \sigma_q(t) \vepsilon.
\end{align*}
This leads to the following inferencing algorithm. In plain words, we sequentially run the denoiser $T$ times from the white noise vector $\vx_T$ back to the generated image $\widehat{\vx}_0$. A pictorial illustration is shown in \fref{fig: DDPM inference}.

\boxedmsg{
\textbf{Inference of DDPM}.
\begin{itemize}
\itemsep\setlength{-1ex}
\item You give us a white noise vector $\vx_T \sim \calN(0,\mI)$.
\item Repeat the following for $t = T, T-1,\ldots,1$.
\item We calculate $\widehat{\vx}_{\vtheta}(\vx_t)$ using our trained denoiser.
\item Update according to
\begin{align}
\vx_{t-1}
&= \frac{(1-\overline{\alpha}_{t-1})\sqrt{\alpha_t}}{1-\overline{\alpha}_t}\vx_t + \frac{(1-\alpha_t)\sqrt{\overline{\alpha}_{t-1}}}{1-\overline{\alpha}_t} \widehat{\vx}_{\vtheta}(\vx_t) +  \sigma_q(t) \vepsilon, \qquad \vepsilon \sim \calN(0,\mI).
\label{eq: DDPM update eqn}
\end{align}
\end{itemize}
}

\subsection{Predicting Noise}
The final step of our derivation is to connect our results back to the original paper of Ho et al \cite{Ho_2020_DDPM} so that the notations will be more consistent with the literature.

\textbf{Training}. If you are familiar with the denoising literature, you probably know the residue-type of algorithm that predicts the noise instead of the signal. The same spirit applies denoising diffusion, where we can learn to predict the noise. To see why this is the case, we consider \eref{eq: recursion for xt, using x0 and epsilon0}. If we re-arrange the terms we will obtain
\begin{align*}
                &\vx_t = \sqrt{ \overline{\alpha}_t } \vx_0 + \sqrt{1-\overline{\alpha}_t} \vepsilon_0\\
\Rightarrow \qquad    &\vx_0 = \frac{\vx_t - \sqrt{1-\overline{\alpha}_t} \vepsilon_0}{\sqrt{ \overline{\alpha}_t } }.
\end{align*}
Substituting this into $\vmu_q(\vx_t,\vx_0)$, we can show that
\begin{align}
\vmu_q(\vx_t,\vx_0)
&= \frac{\sqrt{\alpha_t} (1-\overline{\alpha}_{t-1} ) \vx_t + \sqrt{\overline{\alpha}_{t-1}}(1-\alpha_t)\vx_0}{1-\overline{\alpha}_t} \notag\\
&= \frac{\sqrt{\alpha_t} (1-\overline{\alpha}_{t-1} ) \vx_t + \sqrt{\overline{\alpha}_{t-1}}(1-\alpha_t) \cdot \frac{\vx_t - \sqrt{1-\overline{\alpha}_t} \vepsilon_0}{\sqrt{ \overline{\alpha}_t } } }{1-\overline{\alpha}_t} \notag \\
&= \text{a few more algebraic steps which we shall skip} \notag \\
&= \frac{1}{\sqrt{\alpha_t}}\vx_t - \frac{1-\alpha_t}{\sqrt{1-\overline{\alpha}_t} \sqrt{\alpha}_t} \vepsilon_0.
\label{eq: muq in terms of epsilon}
\end{align}
In words, we have converted $\vmu_q(\vx_t,\vx_0)$ from a function of $\vx_0$ to a function of $\vepsilon_0$.

Since we do this modification, naturally we should modify the mean estimator $\vmu_{\vtheta}$. In order to match the form of $\vmu_{\vtheta}$ with that of $\vmu_q(\vx_t,\vx_0)$, we choose
\begin{align}
\vmu_{\vtheta}(\vx_t) = \frac{1}{\sqrt{\alpha_t}}\vx_t - \frac{1-\alpha_t}{\sqrt{1-\overline{\alpha}_t} \sqrt{\alpha}_t} \widehat{\vepsilon}_{\vtheta}(\vx_t).
\label{eq: mutheta in terms of epsilon}
\end{align}

Substituting \eref{eq: muq in terms of epsilon} and \eref{eq: mutheta in terms of epsilon} into \eref{eq: denoising mu q minus mu theta} will give us a new ELBO
\begin{align*}
\text{ELBO}_{\vtheta}(\vx_0,\vepsilon_0)
&= -\sum_{t=1}^{T} \E_{q(\vx_{t}|\vx_0)}
\Big[
\frac{1}{2\sigma_q^2(t)}
\frac{(1-\alpha_t)^2 }{(1-\overline{\alpha}_t)\alpha_t}
\left\| \widehat{\vepsilon}_{\vtheta}(\vx_t) - \vepsilon_0 \right\|^2
\Big]\\
&= -\sum_{t=1}^{T} \E_{q(\vx_{t}|\vx_0)}
\Big[
\frac{1}{2\sigma_q^2(t)}
\frac{(1-\alpha_t)^2 }{(1-\overline{\alpha}_t)\alpha_t}
\left\| \widehat{\vepsilon}_{\vtheta}\Big(\sqrt{ \overline{\alpha}_t } \vx_0 + \sqrt{1-\overline{\alpha}_t} \vepsilon_0\Big) - \vepsilon_0 \right\|^2
\Big]
\end{align*}
We remark that this ELBO is a function of $\vx_0$ and $\vepsilon_0$. Therefore, to train the denoiser, we need to solve the optimization
\begin{align*}
&\argmin{\vtheta} \;\; \E_{\vx_0,\vepsilon_0} \text{ELBO}_{\vtheta}(\vx_0,\vepsilon_0)\\
&\approx \argmin{\vtheta} \;\; \sum_{\vx_0 \sim \calX} \frac{1}{M}\sum_{m=1}^M \text{ELBO}_{\vtheta}(\vx_0,\vepsilon_0^{(m)}),
\end{align*}
where the superscript denotes the $m$-th initial noise term. If we run stochastic gradient descent, we can simplify the above description to the following procedure.

\boxedmsg{
\textbf{Training DDPM using $\widehat{\vepsilon}_{\vtheta}(\vx_t)$}. For every image $\vx_0$ in your training dataset:
\begin{itemize}
\itemsep\setlength{-1ex}
\item Repeat the following steps until convergence.
\item Pick a random time stamp $t \sim \text{Uniform}[1,T]$.
\item Draw a sample $\vepsilon_0 \sim \calN(0,\mI)$.
\item Draw a sample $\vx_t \sim \calN(\vx_t \,|\, \sqrt{\overline{\alpha}_t} \vx_0, \;\; (1-\overline{\alpha}_t)\mI)$, i.e.,
\begin{equation*}
\vx_t = \sqrt{\overline{\alpha}_t} \vx_0 +  \sqrt{(1-\overline{\alpha}_t)}\vepsilon_0.
\end{equation*}
\item Take gradient descent step on
$$
\nabla_{\vtheta} \left\| \textcolor{blue}{\widehat{\vepsilon}_{\vtheta}(\vx_t)} - \textcolor{blue}{\vepsilon_0} \right\|^2.
$$
\end{itemize}
}

\textbf{Inference}. For the inference, we know that
\begin{align*}
\vx_{t-1}
\sim p_{\vtheta}(\vx_{t-1} \,|\, \vx_t)
&= \calN(\vx_{t-1} \,|\, \vmu_{\vtheta}(\vx_t), \sigma_q(t)^2\mI).
\end{align*}
Using reparameterization, we can show that
\begin{align}
\vx_{t-1} &= \vmu_{\vtheta}(\vx_t) + \sigma_q(t) \vz, \qquad\text{where}\qquad \vz \sim \calN(0,\mI) \notag \\
&= \left(\frac{1}{\sqrt{\alpha_t}}\vx_t - \frac{1-\alpha_t}{\sqrt{1-\overline{\alpha}_t} \sqrt{\alpha}_t} \widehat{\vepsilon}_{\vtheta}(\vx_t)\right) + \sigma_q(t) \vz \notag \\
&= \frac{1}{\sqrt{\alpha_t}} \left(\vx_t - \frac{1-\alpha_t}{\sqrt{1-\overline{\alpha}_t}} \widehat{\vepsilon}_{\vtheta}(\vx_t)\right) + \sigma_q(t) \vz.
\end{align}
Summarizing it here, we have
\boxedmsg{
\textbf{Inference of DDPM using $\widehat{\vepsilon}_{\vtheta}(\vx_t)$}.
\begin{itemize}
\itemsep\setlength{-1ex}
\item You give us a white noise vector $\vx_T \sim \calN(0,\mI)$.
\item Repeat the following for $t = T, T-1,\ldots,1$.
\item We calculate $\widehat{\vx}_{\vtheta}(\vx_t)$ using our trained denoiser.
\item Update according to
\begin{align}
\vx_{t-1}
&=  \frac{1}{\sqrt{\alpha_t}} \left(\vx_t - \frac{1-\alpha_t}{\sqrt{1-\overline{\alpha}_t}} \widehat{\vepsilon}_{\vtheta}(\vx_t)\right) + \sigma_q(t) \vz, \qquad \vz \sim \calN(0,\mI).
\label{eq: DDPM reverse noise predict}
\end{align}
\end{itemize}
}

\subsection{Denoising Diffusion Implicit Model (DDIM)}
\textbf{From DDPM to DDIM}. One of the most prevalent drawbacks of DDPM is that they need a large number of iterations to generate a reasonably good looking image. As mentioned by Song et al in \cite{Song_2023_DDIM}, a DDPM would take more than 1000 hours to generate 50k images (of size $256\times 256$) on a standard GPU. The reason is that when running the reverse diffusion steps, we need to perform denoising. If the reverse diffusion process intrinsically requires many steps to converge, then it will take us many denoising steps. Therefore, to speed up the computing, it is necessary to reduce the number of iterations. DDIM was an invention to overcome this difficulty.

Recall from \eref{eq: transition q from t-1 to t} that the original DDPM transition probability takes the form
\begin{equation*}
q(\vx_t|\vx_{t-1}) \bydef \calN\Big(\vx_t \,|\, \sqrt{\alpha_t} \vx_{t-1},(1-\alpha_t)\mI\Big).
\end{equation*}
In addition, \eref{eq: q t | 0} shows that the probability of $\vx_t$ given $\vx_0$ is
\begin{equation*}
q(\vx_{t}|\vx_0) = \calN\Big(\vx_t \,|\, \sqrt{\overline{\alpha}_{t}}\vx_0, \;\; (1-\overline{\alpha}_{t})\mI \Big),
\end{equation*}
where $\overline{\alpha}_t = \prod_{i=0}^t \alpha_i$ for any decreasing sequence $0 < \alpha_t \le 1$. One important observation here is that the transition probability $q(\vx_t|\vx_{t-1})$ follows a \emph{Markov chain}, meaning that the probability of $\vx_t$ is purely dependent on $\vx_{t-1}$ but not the previous states $\vx_{t-2}$ and so on. The advantage of a Markovian structure is that the system is memoryless. Once we know $\vx_{t-1}$, we will know $\vx_t$. But the downside is that a Markov chain can take many steps to converge. DDIM overcomes this issue by departing from the Markovian structure to non-Markovian.

\textbf{Probability Distributions in DDIM}. To start our discussion, let's follow \cite{Song_2023_DDIM} by picking a special choice of parameters where we replace $\alpha_t$ by a ratio $\alpha_t/\alpha_{t-1}$. This means
\begin{equation*}
q(\vx_t|\vx_{t-1}) \bydef \calN\Big(\vx_t \,\Big|\, \sqrt{\frac{\alpha_t}{\alpha_{t-1}}} \vx_{t-1}, \;\; (1-\frac{\alpha_t}{\alpha_{t-1}})\mI\Big).
\end{equation*}
There is no particularly strong physical meaning for this choice, except that it makes the notation simpler. With this choice, the product term is simplified to $\overline{\alpha}_t = \prod_{i=1}^t \frac{\alpha_i}{\alpha_{i-1}} = \alpha_t$ assuming that $\alpha_0 = 1$. Therefore,
\begin{equation}
q(\vx_{t}|\vx_0) = \calN\Big(\vx_t \,|\, \sqrt{\alpha_t}\vx_0, \;\; (1-\alpha_t)\mI \Big).
\label{eq: DDIM q xt | x0}
\end{equation}

Expressing \eref{eq: DDIM q xt | x0} by means of reparametrization, we note that $\vx_t$ in \eref{eq: DDIM q xt | x0} can be represented in terms of $\vx_0$ as follows
\begin{align*}
\vx_t = \sqrt{\alpha_t}\vx_0 + \sqrt{1-\alpha_t}\vepsilon, \qquad \text{where} \quad \vepsilon \sim \calN(0,\mI).
\end{align*}
By the same argument, we can write
\begin{align}
\vx_{t-1} = \sqrt{\alpha_{t-1}}\vx_0 + \sqrt{1-\alpha_{t-1}}\vepsilon, \qquad \text{where} \quad \vepsilon \sim \calN(0,\mI).
\label{eq: DDIM q x t-1 | x0 (1)}
\end{align}
So here comes an interesting trick. Let's replace $\vepsilon$ by something so that $\vx_{t-1}$ is no longer $\vx_0$ perturbed by white noise. Perhaps we can consider the following derivation
\begin{alignat*}{2}
                & \qquad &\vx_t                         &= \sqrt{\alpha_t}\vx_0 + \sqrt{1-\alpha_t}\vepsilon\\
\Longrightarrow & \qquad &\sqrt{1-\alpha_t}\vepsilon  &= \vx_t - \sqrt{\alpha_t}\vx_0 \\
\Longrightarrow & \qquad &\vepsilon                   &= \frac{\vx_t - \sqrt{\alpha_t}\vx_0}{\sqrt{1-\alpha_t}}.
\end{alignat*}
So, substituting $\vepsilon$ into \eref{eq: DDIM q x t-1 | x0 (1)}, we obtain
\begin{align}
\vx_{t-1}
&= \sqrt{\alpha_{t-1}}\vx_0 + \sqrt{1-\alpha_{t-1}} \textcolor{blue}{\vepsilon} \notag \\
&= \sqrt{\alpha_{t-1}}\vx_0 + \sqrt{1-\alpha_{t-1}} \textcolor{blue}{\left( \frac{\vx_t - \sqrt{\alpha_t}\vx_0}{\sqrt{1-\alpha_t}} \right)}.
\label{eq: DDIM q x t-1 | x0 (2)}
\end{align}
The difference between \eref{eq: DDIM q x t-1 | x0 (2)} and \eref{eq: DDIM q x t-1 | x0 (1)} is that in \eref{eq: DDIM q x t-1 | x0 (1)}, the noise term is $\vepsilon$ which is $\calN(0,\mI)$. It is this Gaussian that makes the derivations of DDPM easy, but it is also this Gaussian that makes the reverse diffusion slow. In contrast, \eref{eq: DDIM q x t-1 | x0 (2)} replaces the Gaussian by an estimate. This estimate uses the previous signal $\vx_t$ combined with the initial signal $\vx_0$. Of course, one can argue that in DDPM (e.g., \eref{eq: VDM q t-1 | t, 0}) also uses a combination of $\vx_t$ and $\vx_0$. The difference is that the combination in \eref{eq: DDIM q x t-1 | x0 (2)} allows us to do something that \eref{eq: VDM q t-1 | t, 0} does not, which is the derivation of the marginal distribution $q(\vx_{t-1}|\vx_0)$ and make it to a desired form.

Let's elaborate more on the marginal distribution. Referring to \eref{eq: DDIM q x t-1 | x0 (2)}, we notice that we can \emph{choose}
\begin{align*}
q(\vx_{t-1}|\vx_t,\vx_0) = \calN\left(\sqrt{\alpha_{t-1}}\vx_0 + \sqrt{1-\alpha_{t-1}}\left( \frac{\vx_t - \sqrt{\alpha_t}\vx_0}{\sqrt{1-\alpha_t}} \right) \;\;, \;\;\; \text{something} \right).
\end{align*}
where ``something'' stands for the variance of the Gaussian which can be made as $\sigma_t^2\mI$ for some hyperparameter $\sigma_t$. One important (likely the most important) argument in DDIM is that we want the marginal distribution $q(\vx_{t-1}|\vx_0)$ to have the same form as $q(\vx_t|\vx_0)$:
\begin{equation*}
q(\vx_{t-1}|\vx_0) = \calN(\sqrt{\alpha_{t-1}}\vx_0, \;\; (1-\alpha_{t-1})\mI).
\end{equation*}
The reason of aiming for this distribution is that ultimately we care about the marginal distribution $q(\vx_{t}|\vx_0)$ which we want it to become pure white noise when $t = T$ and it is the original image when $t = 0$. Therefore, while we can have millions of different choice of the transitional distribution $q(\vx_{t-1}|\vx_t,\vx_0)$, only some very specialized transition probabilities can ensure that $q(\vx_{t-1}|\vx_0)$ takes a form we like.

\textbf{Derivation of the Transition Distribution}. With this goal in mind, we now state our mathematical problem. Suppose that
\begin{align}
q(\vx_{t}|\vx_0)         &= \calN\Big(\sqrt{\alpha_t}\vx_0, \;\; (1-\alpha_t)\mI \Big), \notag \\
q(\vx_{t-1}|\vx_t,\vx_0) &= \calN\left(\sqrt{\alpha_{t-1}}\vx_0 + \sqrt{1-\alpha_{t-1}}\left( \frac{\vx_t - \sqrt{\alpha_t}\vx_0}{\sqrt{1-\alpha_t}} \right) \;\;, \;\;\; \sigma_t^2\mI \right),
\label{eq: DDIM q x t-1 | x0 (3)}
\end{align}
can we ensure that $q(\vx_{t-1}|\vx_0) = \calN(\sqrt{\alpha_{t-1}}\vx_0, \;\; (1-\alpha_{t-1})\mI)$? If not, what additional changes do we need?

The answer to this mathematical question requires some tools from textbooks. We recall the following result from Bishop's textbook \cite{Bishop_2006}.
\boxedthm{
\textbf{Bishop \cite[Eqn 2.115]{Bishop_2006}}
Suppose that we have two random variables $\vx$ and $\vy$ following the distributions
\begin{align*}
p(\vx)      &= \calN(\vmu,\mLambda^{-1}),\\
p(\vy|\vx)  &= \calN(\mA\vx+\vb, \mL^{-1}).
\end{align*}
Then we can show that the marginal distribution is
\begin{align*}
p(\vy) = \int p(\vy|\vx)p(\vx)d\vx = \calN\Big(\mA\vmu+\vb, \quad \mL^{-1} + \mA\mLambda^{-1}\mA^{-1}\Big).
\end{align*}
}
Let's see how we can apply this result to our problem. Looking at \eref{eq: DDIM q x t-1 | x0 (3)}, we can identify the following qualities:
\begin{align*}
\mA = \sqrt{\frac{1-\alpha_{t-1}}{1-\alpha_t}}, \;\quad\; \vmu = \sqrt{\alpha_t}\vx_0, \;\quad\; \vb = \sqrt{\alpha_{t-1}}\vx_0 - \sqrt{\frac{1-\alpha_{t-1}}{1-\alpha_t}}\sqrt{\alpha_t}\vx_0.
\end{align*}

Suppose that $q(\vx_{t-1}|\vx_0) = \calN(\vmu_{t-1}, \sigma_{t-1}^2\mI)$ for some unknown choices of mean $\vmu_{t-1}$ and variance $\sigma_{t-1}^2$. If we can show that $\vmu_{t-1} = \sqrt{\alpha_{t-1}}\vx_0$ and $\sigma_{t-1}^2 = (1-\alpha_{t-1})$, then we are done. To this end, we show that
\begin{align*}
\vmu_{t-1}
&= \mA\vmu + \vb\\
&= \sqrt{\frac{1-\alpha_{t-1}}{1-\alpha_t}} \cdot \sqrt{\alpha_t}\vx_0 + \sqrt{\alpha_{t-1}}\vx_0 - \sqrt{\frac{1-\alpha_{t-1}}{1-\alpha_t}}\sqrt{\alpha_t}\vx_0 \\
&= \sqrt{\alpha_{t-1}}\vx_0.
\end{align*}
Oh, great news! We have shown that $\vmu_{t-1} = \sqrt{\alpha_{t-1}}\vx_0$. That means for the transitional distribution $q(\vx_{t-1}|\vx_t,\vx_0)$ we have chosen, the marginal distribution has the desired mean.

So it remains to check the variance. We show that
\begin{align*}
\sigma_{t-1}^2
&= \mL^{-1} + \mA\mLambda^{-1}\mA^T\\
&= \sigma_t^2 + \sqrt{\frac{1-\alpha_{t-1}}{1-\alpha_t}} \cdot (1-\alpha_t) \cdot \sqrt{\frac{1-\alpha_{t-1}}{1-\alpha_t}}\\
&= \sigma_t^2 + (1-\alpha_{t-1}).
\end{align*}
Oh, no! We cannot show that $\sigma_{t-1}^2 = 1-\alpha_{t-1}$. There is an additional term $\sigma_t^2$ here. But this is not such a big deal. How about we do a quick fix by \emph{adding} $\sigma_{t}^2$ into $\mA$:
\begin{align*}
\sigma_{t-1}^2
&= \sigma_t^2 + \sqrt{\frac{1-\alpha_{t-1} - \textcolor{blue}{\sigma_t^2} }{1-\alpha_t}} \cdot (1-\alpha_t) \cdot \sqrt{\frac{1-\alpha_{t-1}- \textcolor{blue}{\sigma_t^2} }{1-\alpha_t}}\\
&= \sigma_t^2 + 1-\alpha_{t-1} - \textcolor{blue}{\sigma_t^2} \\
&= 1-\alpha_{t-1}.
\end{align*}
Aha! $\sigma_{t-1}^2$ now takes the desired form such that $\sigma_{t-1}^2 = 1-\alpha_{t-1}$. Let's do a quick check to make sure this additional $\sigma_t^2$ does not affect the mean:
\begin{align*}
\vmu_{t-1}
&= \mA\vmu + \vb\\
&= \sqrt{\frac{1-\alpha_{t-1}- \textcolor{blue}{\sigma_t^2}}{1-\alpha_t}} \cdot \sqrt{\alpha_t}\vx_0 + \sqrt{\alpha_{t-1}}\vx_0 - \sqrt{\frac{1-\alpha_{t-1}- \textcolor{blue}{\sigma_t^2}}{1-\alpha_t}}\sqrt{\alpha_t}\vx_0 \\
&= \sqrt{\alpha_{t-1}}\vx_0.
\end{align*}
So, the mean remains the desired form despite we changed the variance term.

To summarize, we can \emph{choose} $q(\vx_{t-1}|\vx_t,\vx_0)$ to be the following.
\boxedthm{
\textbf{DDIM Transition Distribution}. In DDIM, the transition distribution is defined as
\begin{equation}
q(\vx_{t-1}|\vx_t,\vx_0) = \calN\left(\sqrt{\alpha_{t-1}}\vx_0 + \sqrt{1-\alpha_{t-1}-\sigma_t^2}\left( \frac{\vx_t - \sqrt{\alpha_t}\vx_0}{\sqrt{1-\alpha_t}} \right) \;\;, \;\;\; \sigma_t^2\mI \right).
\end{equation}
}
If $q(\vx_t|\vx_0) = \calN\Big(\sqrt{\alpha_t}\vx_0, \;\; (1-\alpha_t)\mI \Big)$, then it follows from our derivation that $q(\vx_{t-1}|\vx_0) = \calN(\sqrt{\alpha_{t-1}}\vx_0, \;\; (1-\alpha_{t-1})\mI)$.

\textbf{Inference for DDIM}. The inference for DDIM is derived based on the transition distribution. Starting with the forward process, if we want to perform the reverse, we will need to find out $\vx_0$ from the following equation:
\begin{align*}
\underset{\text{given}}{\underbrace{\vx_t}} = \underset{\text{want to find}}{\underbrace{\sqrt{\alpha_t} \vx_0}} + \underset{\text{estimated by network}}{\underbrace{\sqrt{1-\alpha_t}\vepsilon}}.
\end{align*}
By rearranging the terms, we see that
\begin{alignat*}{2}
                    & \qquad & \vx_0                       & = \frac{1}{\sqrt{\alpha_t}}\left(\vx_t - \sqrt{1-\alpha_t}\vepsilon\right)\\
\Longrightarrow     & \qquad & \textcolor{blue}{f_{\vtheta}^{(t)}(\vx_t)}    &\bydef \frac{1}{\sqrt{\alpha_t}}\left( \vx_t - \sqrt{1-\alpha_t} \textcolor{blue}{\vepsilon_{\vtheta}^{(t)}(\vx_t)}\right).
\end{alignat*}
There are two new terms in this equation. The first one is $\vepsilon_{\vtheta}^{(t)}(\vx_t)$ which replaces $\vepsilon$. It is the estimate of the noise based on the current input $\vx_t$. The second term is $f_{\vtheta}^{(t)}(\vx_t)$ which is defined as the equation $\frac{1}{\sqrt{\alpha_t}}\left( \vx_t - \sqrt{1-\alpha_t} \vepsilon_{\vtheta}^{(t)}(\vx_t)\right)$. We can think of it as the estimate of the true signal $\vx_0$.

Going back to the transition distribution, we recall that it is denoted by $q(\vx_{t-1}|\vx_t,\vx_0)$. If we do not have access to $\vx_0$, we can replace it $f_{\vtheta}^{(t)}(\vx_t)$. This means that
\begin{alignat}{2}
                    & \qquad & p_{\vtheta}(\vx_{t-1}|\vx_t)     &= q(\vx_{t-1}|\vx_t,\vx_0) \notag \\
\Longrightarrow     & \qquad & p_{\vtheta}(\vx_{t-1}|\vx_t)     &\bydef q(\vx_{t-1}|\vx_t,f_{\vtheta}^{(t)}(\vx_t)) \notag \\
                    & \qquad &                                  &=
                    \calN\left(\sqrt{\alpha_{t-1}}f_{\vtheta}^{(t)}(\vx_t) + \sqrt{1-\alpha_{t-1}-\sigma_t^2} \cdot \frac{\vx_t - \sqrt{\alpha_t}f_{\vtheta}^{(t)}(\vx_t) }{\sqrt{1-\alpha_t}}  ,\quad  \sigma_t^2\mI\right) \notag\\
                    & \qquad &                                  &=
                    \calN\Bigg(\sqrt{\alpha_{t-1}}\cdot
                    \textcolor{blue}{\frac{1}{\sqrt{\alpha_t}}\left( \vx_t - \sqrt{1-\alpha_t} \vepsilon_{\vtheta}^{(t)}(\vx_t)\right)} + \notag \\
                    & \qquad &                                  &\quad \qquad +
                    \sqrt{1-\alpha_{t-1}-\sigma_t^2} \cdot \frac{\vx_t - \sqrt{\alpha_t} \textcolor{blue}{\left(\frac{1}{\sqrt{\alpha_t}}\left( \vx_t - \sqrt{1-\alpha_t} \vepsilon_{\vtheta}^{(t)}(\vx_t)\right)\right)} }{\sqrt{1-\alpha_t}}  ,\quad  \sigma_t^2\mI\Bigg) \notag \\
                    & \qquad &                                  &=
                    \calN\Bigg(\sqrt{\alpha_{t-1}} \left( \frac{\vx_t - \sqrt{1-\alpha_t} \vepsilon_{\vtheta}^{(t)}(\vx_t)}{\sqrt{\alpha_t}} \right)
                    + \sqrt{1-\alpha_{t-1}-\sigma_t^2}\cdot \vepsilon_{\vtheta}^{(t)}(\vx_t), \qquad \sigma_t^2\mI \Bigg).
                    \label{eq: DDIM p xt-1 | xt}
\end{alignat}
For the special case where $t = 1$, we define $p_{\vtheta}(\vx_{t-1}|\vx_t) = \calN(f_{\vtheta}^{(1)}(\vx_1), \sigma_1^2\mI)$ so that the reverse process is supported everywhere. Looking at \eref{eq: DDIM p xt-1 | xt}, we use reparametrization to write it as follows and interpret the equation according to \cite{Song_2023_DDIM}.
\begin{equation}
\text{(DDIM)} \qquad \vx_{t-1} = \sqrt{\alpha_{t-1}}
            \underset{\text{predicted $\vx_0$}}{\underbrace{\left( \frac{\vx_t - \sqrt{1-\alpha_t} \vepsilon_{\vtheta}^{(t)}(\vx_t)}{\sqrt{\alpha_t}} \right)}}
                    +
            \underset{\text{direction pointing to $\vx_t$}}{\underbrace{\sqrt{1-\alpha_{t-1}-\sigma_t^2} \cdot \vepsilon_{\vtheta}^{(t)}(\vx_t)}} + \sigma_t \underset{\sim \calN(0,\mI)}{\underbrace{\vepsilon_t}}.
\label{eq: DDIM x t-1 reparametrized}
\end{equation}
It would be helpful to compare this equation with the DDPM equation in \eref{eq: DDPM reverse noise predict}:
\begin{equation}
\text{(DDPM)} \qquad  \vx_{t-1}
=  \frac{1}{\sqrt{\alpha_t}} \left(\vx_t - \frac{1-\alpha_t}{\sqrt{1-\overline{\alpha}_t}} \vepsilon^{(t)}_{\vtheta}(\vx_t)\right) + \sigma_t \vepsilon_t, \qquad \vepsilon_t \sim \calN(0,\mI).
\label{eq: DDIM x t-1 reparametrized DDPM}
\end{equation}
The main difference between DDPM and DDIM is subtle. While they both use $\vx_{t}$ and $\vepsilon^{(t)}(\vx_t)$ in their updates, the specific update formula makes a different convergence speed. In fact, later in the differential equation literature where people connect DDIM and DDPM with stochastic differential equations, it was observed that DDIM employed some special accelerated first-order numerical schemes when solving the differential equation.

\subsection{Concluding Remark}
The literature of DDPM is quickly exploding. The original paper by Sohl-Dickstein et al. \cite{Sohl-Dickstein_2015_ICML} and Ho et al. \cite{Ho_2020_DDPM} are the must-reads to understand the topic. For a more ``user-friendly'' version, we found that the tutorial by Luo very useful \cite{Luo_2022_arXiv}. Some follow up works are highly cited, including the denoising diffusion implicit models by Song et al. \cite{Song_2023_DDIM}. In terms of application, people have been using DDPM for various image synthesis applications, e.g., \cite{Rombach_2022_CVPR, Saharia_2022_Text2Image}.

\newpage
\section{Score-Matching Langevin Dynamics (SMLD)}
\setcounter{figure}{0}
\setcounter{equation}{0}

Score-based generative models \cite{Song_2021_SGM} are alternative approaches to generate data from a desired distribution. There are several core ingredients: the Langevin equation, the (Stein) score function, and the score-matching loss. The focus of this section is more on the latter two. We will make a few hand-waving arguments about the Langevin equation and explain how to make it computationally feasible for our generative task. More in-depth discussions about the Langevin equation will be postponed to the next section.

\subsection{Sampling from a Distribution} Imagine that we are given a distribution $p(\vx)$ and we want to draw samples from $p(\vx)$. If $\vx$ is a one-dimensional variable, we can achieve the goal easily by inverting the cumulative distribution function (CDF) \cite[Chapter 4]{Chan_2021_book} and sending a $\text{uniform}[0,1]$ random variable through this inverted CDF. For high dimensional distributions, the same CDF technique does not apply. However, the intuition of giving a higher weight to samples with a higher probability remains valid. For example, if we want to draw a sample from the orange dataset, it is almost certain that we want a spherical orange than a cubical orange.

\begin{figure}[h]
\centering
\includegraphics[width=0.5\linewidth]{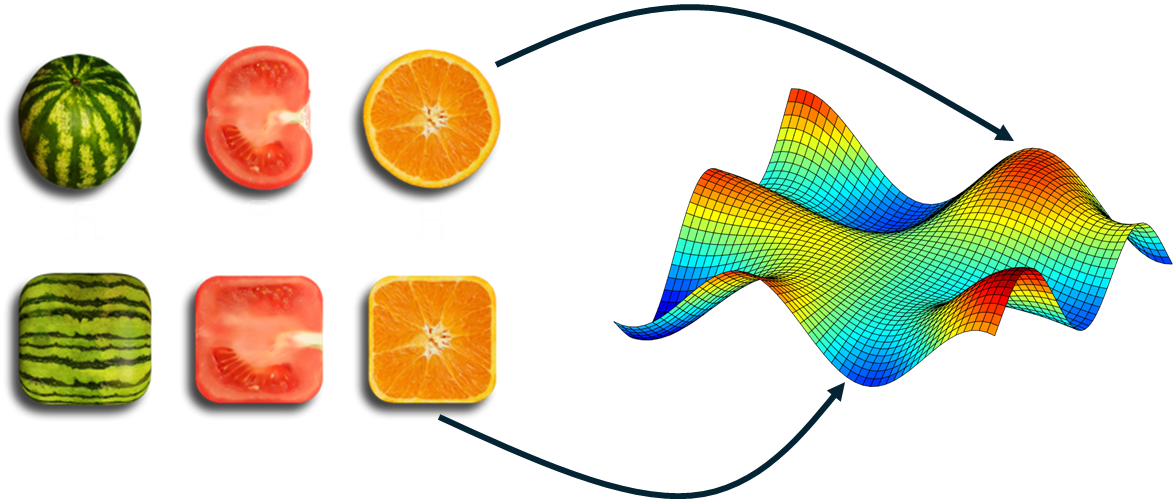}
\caption{If our goal is to generate a fruit image, then we should expect the underlying distribution $p(\vx)$ will have a higher value for those ``normal-looking'' fruit images than those ``weird-looking'' images. Therefore, to sample from a distribution, it is more natural to pick a sample from a higher position of the distribution. The fruit image is taken from \url{https://cognitiveseo.com/blog/4224/unnatural-links-definition-examples/}}
\end{figure}

Therefore, in a non-rigorous way, we can argue that if we are given $p(\vx)$, we should aim to draw samples from a location where $p(\vx)$ has a high value. This idea of searching for a higher probability can be translated into an optimization
\begin{equation*}
\vx^* = \argmax{\vx} \;\; \log p(\vx),
\end{equation*}
where the goal is to maximize the log-likelihood of the distribution $p(\vx)$. Certainly, this maximization does not provide any clue on how to draw a low probability sample which we will explain through the lens of Langevin equation. For now, we want to make a remark about the difference between the maximization here and maximum likelihood estimation. In maximum likelihood, the data point $\vx$ is fixed but the model parameters are changing. Here, the model parameters are fixed but the data point is changing. We are given a \emph{fixed} model. Our goal is to draw the most likely sample from \emph{this} model. The table below shows the difference between our sampling problem and maximum likelihood estimation.

\begin{table}[h]
\centering
\renewcommand{\arraystretch}{1.5}
\begin{tabular}{p{2in}p{2in}p{2in}}
\hline
\hline
Problem             & Sampling & Maximum Likelihood \\
\hline
Optimization target & A sample $\vx$  & Model parameter $\vtheta$ \\
Formulation         & $\vx^* = \argmax{\vx} \;\; \log p(\vx;\vtheta)$ & $\vtheta^* = \argmax{\vtheta} \;\; \log p(\vx;\vtheta)$\\
\hline
\end{tabular}
\end{table}

Let's continue our argument about the maximization. If $p(\vx)$ is a simple parametric model, the maximization will have analytic solutions. However, in general, optimizations in a high-dimensional space are ill-posed with many local minima. Therefore, there is no single algorithm that is globally converging in all situations. A reasonable trade-off between computational complexity, memory requirement, difficulty of implementation, and solution quality is the gradient descent algorithm which is a first-order method. For our optimization where the objective function is $\log p(\vx)$, the gradient descent algorithm defines a sequence of iterates via the update rule
\begin{align*}
\vx_{t+1} = \vx_t + \tau \nabla_{\vx} \log p(\vx_t),
\end{align*}
where $\nabla_{\vx} \log p(\vx_t)$ denotes the gradient of $\log p(\vx)$ evaluated at $\vx_t$, and $\tau$ is the step size. Here we use ``$+$'' instead of the typical ``$-$'' because we are solving a maximization problem.

If you agree with the above approach, we can now provide an information introduction about the Langevin equation. Without worrying too much about its roots in physics, we can treat Langevin equation as an iterative procedure that allows us to draw samples.
\boxeddef{
\label{def: Langevin discrete}
The (discrete-time) \textbf{Langevin equation} for sampling from a known distribution $p(\vx)$ is an iterative procedure for $t = 1,\ldots,T$:
\begin{equation}
\vx_{t+1} = \vx_t + \tau \nabla_{\vx} \log p(\vx_t) + \sqrt{2\tau} \vz, \qquad \vz \sim \calN(0,\mI),
\label{eq: Langevin dynamics main equation}
\end{equation}
where $\tau$ is the step size which users can control, and $\vx_0$ is white noise.
}

\boxedeg{
Consider a Gaussian distribution $p(x) = \calN(x \,|\, \mu, \sigma^2)$, we can show that the Langevin equation is
\begin{align*}
x_{t+1}
&= x_t + \tau \cdot \nabla_x  \log \left\{ \frac{1}{\sqrt{2\pi \sigma^2}} e^{-\frac{(x_t-\mu)^2}{2\sigma^2}} \right\} + \sqrt{2\tau} z & \\
&= x_t - \tau \cdot \frac{x_t-\mu}{\sigma^2} + \sqrt{2\tau} z, &\quad z \sim \calN(0,1),
\end{align*}
where the initial state can be set as $x_0 \sim \calN(0,1)$.
}

If we ignore the noise term $\sqrt{2\tau}\vz$, the Langevin equation in \eref{eq: Langevin dynamics main equation} is exactly \textbf{gradient descent}, but for a particular function --- the log likelihood of the random variable $\vx$. Therefore, while gradient descent is a generic first-order optimization algorithm for any objective function, the Langevin equation focuses on the distribution if we use it in the context of generative models. The gradient descent algorithm plus a small noise perturbation gives us a simple summary.
\boxedmsg{
In generative models,
\begin{center}
Langevin equation = \textbf{gradient descent + noise}
\end{center}
\begin{flushright}
applied to the log-likelihood function.
\end{flushright}
}

But why do we want gradient descent + noise instead of gradient descent? One interpretation is that we are not interested in solving the optimization problem. Instead, we are more interested in \emph{sampling} from a distribution. By introducing the random noise to the gradient descent step, we randomly pick a sample that is following the objective function's trajectory while not staying at where it is. If we are closer to the peak, we will move left and right slightly. If we are far from the peak, the gradient direction will pull us towards the peak. If the curvature around the peak is sharp, we will concentrate most of the steady state points $\vx_T$ there. If the curvature around the peak is flat, we will spread around. Therefore, by repeatedly initializing the gradient descent (plus noise) algorithm at a uniformly distributed location, we will eventually collect samples that will follow the distribution we designate.

A slightly more formal way to justify the Langevin equation is the Fokker-Planck equation. The Fokker-Planck equation is a fundamental result of stochastic processes. For any Markovian processes (e.g., Wiener process and Brownian motion), the dynamics of the solution $\vx_t$ is described by a stochastic differential equation (i.e., Langevin equation). However, since $\vx_t$ is a random variable at any time $t$, there is an underlying probability distribution $p(\vx,t)$ associated with each $\vx_t$. Fokker-Planck equation provides a mathematical statement about the distribution. The distribution must satisfy a partial differential equation. Roughly speaking, in the context of our problem, the Fokker-Planck equation can be described below.
\boxedthm{
In the context of our problem, the solution $\vx_t$ of the Langevin equation will have a probability distribution $p(\vx,t)$ at time $t$ satisfying the \textbf{Fokker-Planck Equation}
\begin{equation}
\partial_t p(\vx,t) = -\partial_{\vx} \Big\{ [\partial_{\vx} (\log p(\vx))] p(\vx,t) \Big\} + \partial^2_{\vx}p(\vx,t),
\end{equation}
where $\log p(\vx)$ is the log-likelihood of the ground truth distribution.
}
Deriving the Fokker-Planck equation will take a tremendous amount of effort. However, if we are given a candidate solution, verifying whether it satisfies the Fokker-Planck equation is not hard.

\boxedproof{
\textbf{Verification of Theorem}. Suppose that we have run Langevin equation for long enough that we have reached a converging solution $\vx_t$ as $t \rightarrow \infty$. We argue that this limiting distribution is $p(\vx)$. Indeed, we can show that
\begin{align*}
\partial_{\vx} \Big\{ \log p(\vx) \Big\} = \frac{\partial_{\vx}p(\vx)}{p(\vx)}.
\end{align*}
Then, substituting $p(\vx,t)$ by $p(\vx)$ when $t\rightarrow\infty$, we can show that
\begin{align*}
-\partial_{\vx} \Big\{ [\partial_{\vx} (\log p(\vx))] p(\vx) \Big\} + \partial^2_{\vx}p(\vx)
&= \partial_{\vx} \Big\{ [-\partial_{\vx} (\log p(\vx))] p(\vx)   + \partial_{\vx}p(\vx) \Big\}\\
&= \partial_{\vx} \Big\{ -\frac{\partial_{\vx}p(\vx)}{p(\vx)} p(\vx)   + \partial_{\vx}p(\vx) \Big\}\\
&= \partial_{\vx} \Big\{ -\partial_{\vx}p(\vx) + \partial_{\vx}p(\vx)\Big\} = 0.
\end{align*}
On the other hand, when $t \rightarrow \infty$, it holds that $\partial_t p(\vx) = 0$. Therefore, the Fokker-Planck equation is verified.
}

\boxedeg{
Consider a Gaussian mixture $p(x) = \pi_1\calN(x \,|\, \mu_1, \sigma_1^2) + \pi_2\calN(x \,|\, \mu_2, \sigma_2^2)$. We can calculate the gradient $\nabla_{x} \log p(x)$ analytically or numerically. For demonstration, we choose $\pi_1 = 0.6$. $\mu_1 = 2$, $\sigma_1 = 0.5$, $\pi_2 = 0.4$, $\mu_2 = -2$, $\sigma_2 = 0.2$. We initialize $x_0 = 0$. We choose $\tau = 0.05$. We run the above gradient descent iteration for $T = 500$ times, and we plot the trajectory of the values $p(x_t)$ for $t = 1,\ldots,T$. As we can see in the figure below, the sequence $\{x_1,x_2,\ldots,x_T\}$ simply follows the shape of the Gaussian and climb to one of the peaks.

What is more interesting is when we add the noise term. Instead of landing at the peak, the sequence $x_t$ moves around the peak and finishes somewhere near the peak. (Remark: To terminate the algorithm, we can gradually make $\tau$ smaller or we can early stop.)

\begin{center}
\begin{tabular}{cc}
\includegraphics[width=0.45\linewidth]{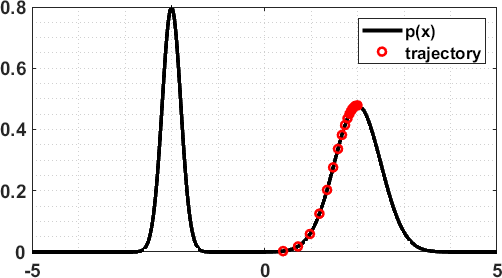}&
\includegraphics[width=0.45\linewidth]{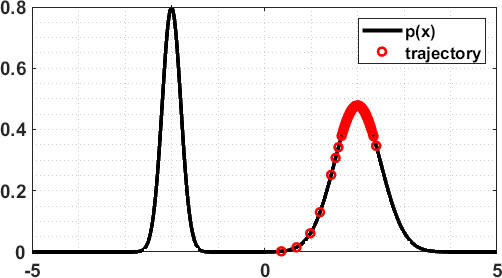}\\
$\vx_{t+1} = \vx_t + \tau \nabla_{\vx} \log p(\vx_t)$
&
$\vx_{t+1} = \vx_t + \tau \nabla_{\vx} \log p(\vx_t) + \sqrt{2\tau}\vz$
\end{tabular}
\captionof{figure}{Deterministic algorithm aiming to pick a sample that maximizes the likelihood, versus a stochastic algorithm which adds noise at every iteration.}
\end{center}
}

\begin{figure}[h]
\centering
\includegraphics[width=0.5\linewidth]{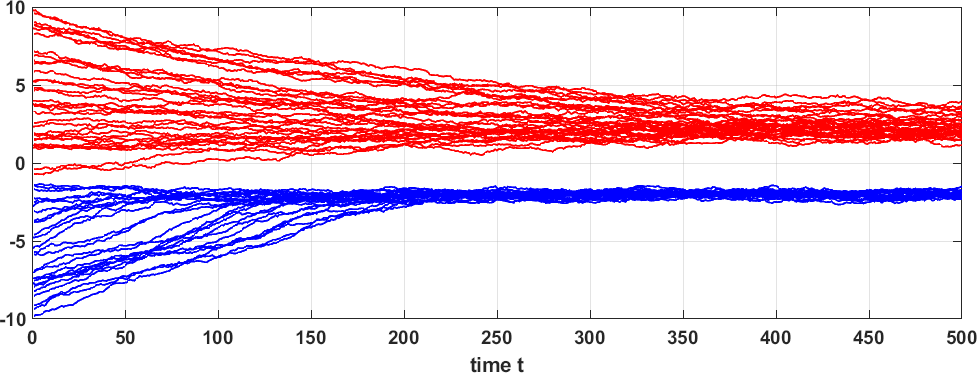}
\caption{Trajectory of sample evolutions using the Langevin dynamics. We colored the two modes of the Gaussian mixture in different colors for better visualization. The setting here is identical to the example above, except that the step size is $\tau = 0.001$.}
\label{fig: Langevin trajectory 1D}
\end{figure}

\fref{fig: Langevin trajectory 1D} shows an interesting description of the sample trajectory. Starting with an arbitrary location, the data point $\vx_t$ will do a random walk according to the Langevin dynamics equation. The direction of the random walk is not completely arbitrary. There is a certain amount of pre-defined drift while at every step there is some level of randomness. The drift is determined by $\nabla_{\vx} \log p(\vx)$ whereas the randomness comes from $\vz$.

\boxedeg{
Following the previous example we again consider a Gaussian mixture $$p(x) = \pi_1\calN(x \,|\, \mu_1, \sigma_1^2) + \pi_2\calN(x \,|\, \mu_2, \sigma_2^2).$$ We choose $\pi_1 = 0.6$. $\mu_1 = 2$, $\sigma_1 = 0.5$, $\pi_2 = 0.4$, $\mu_2 = -2$, $\sigma_2 = 0.2$. Suppose we initialize $M = 10000$ uniformly distributed samples $x_0 \sim \text{Uniform}[-3,3]$. We run Langevin updates for $t = 100$ steps. The histograms of generated samples are shown in the figures below.

\begin{center}
\begin{tabular}{cccc}
\hspace{-2ex}\includegraphics[trim={1cm 0 1cm 0},clip, width=0.24\linewidth]{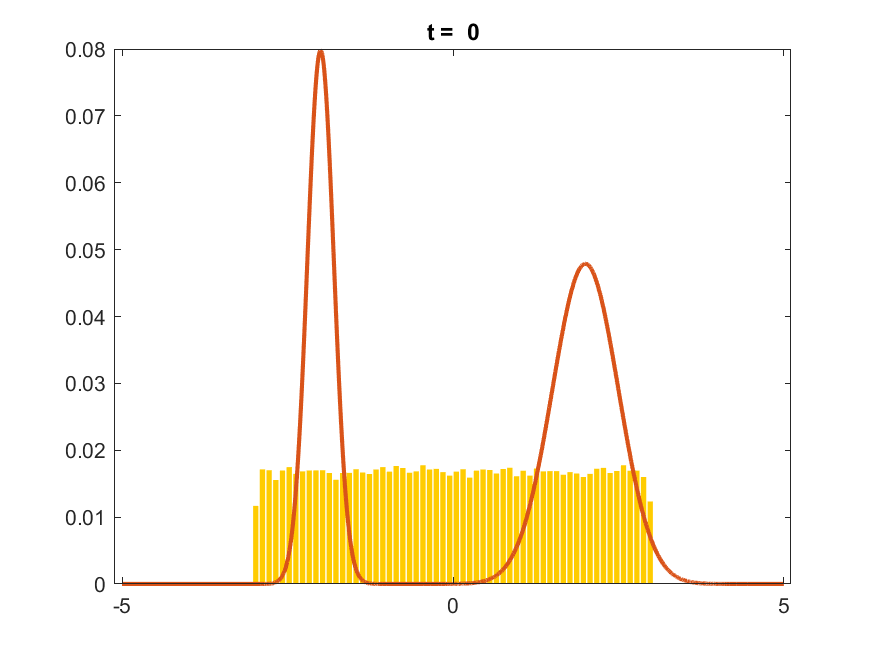}&
\hspace{-2ex}\includegraphics[trim={1cm 0 1cm 0},clip, width=0.24\linewidth]{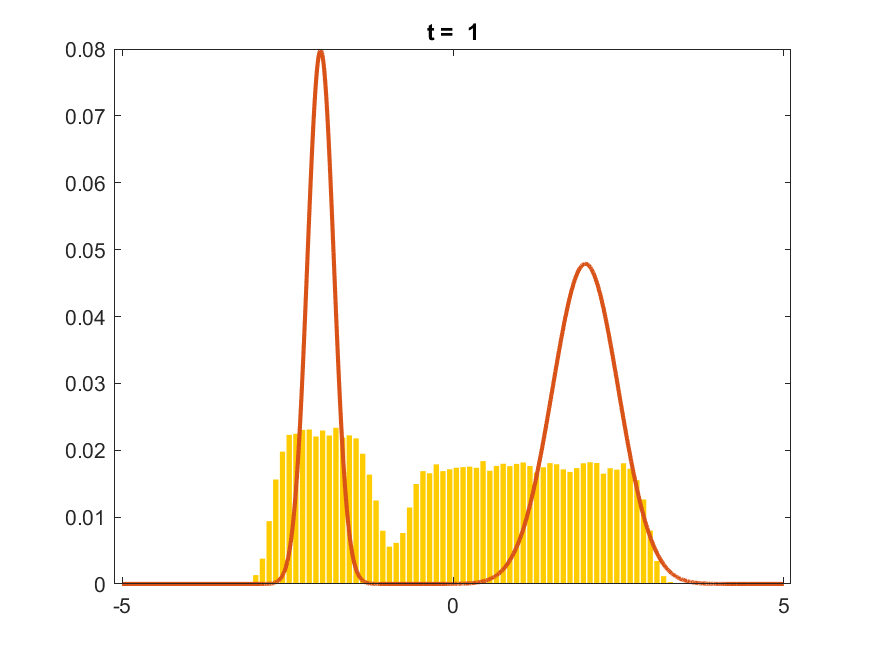}&
\hspace{-2ex}\includegraphics[trim={1cm 0 1cm 0},clip, width=0.24\linewidth]{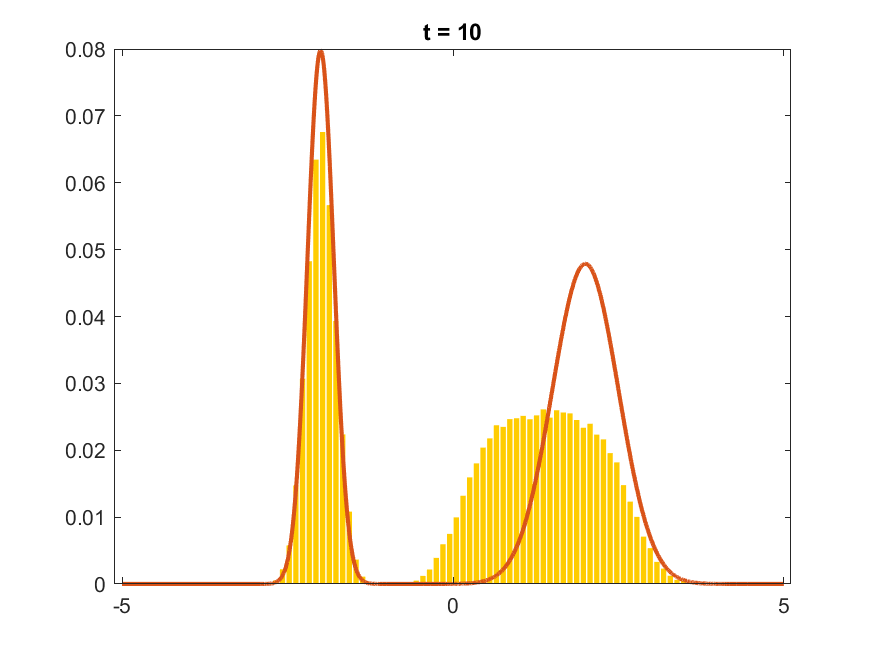}&
\hspace{-2ex}\includegraphics[trim={1cm 0 1cm 0},clip, width=0.24\linewidth]{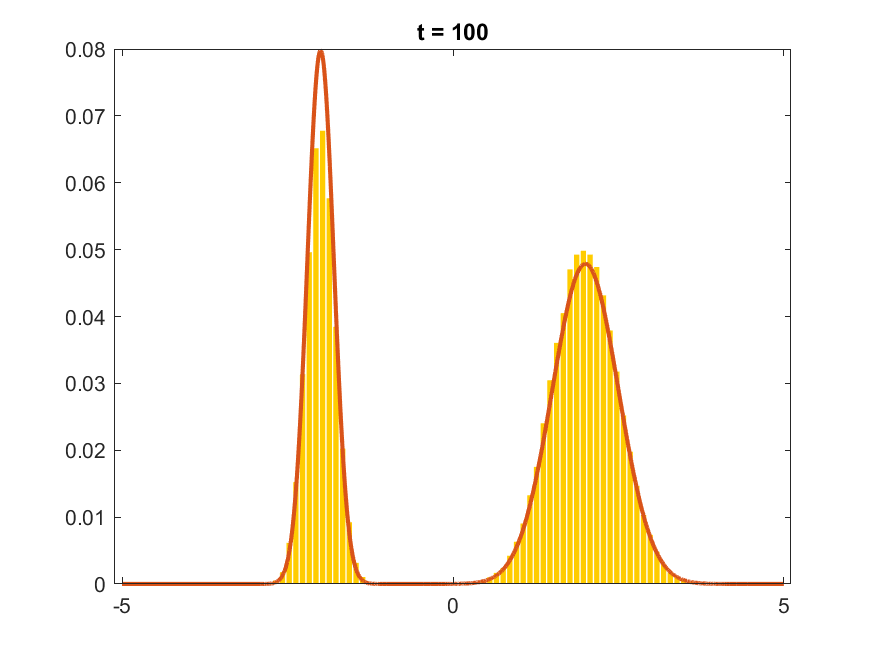}
\end{tabular}
\captionof{figure}{Samples generated by Langevin dynamics. Initially the samples are uniformly distributed. As time progresses, the distribution of the samples become the desired distribution.}
\end{center}
}

\boxedmsg{
\textbf{Remark 1: Stochastic Gradient Langevin Dynamics}. The dynamical behavior of $\vx_t$ governed by the Langevin equation is often known as the Langevin dynamics. Langevin dynamics uses gradient descent plus noise. This is not the same as stochastic gradient descent (SGD). SGD uses minibatches to approximate the full gradient. There is no noise. The randomness in SGD comes from the minibatch which are uniformly sampled from the training dataset. SGD can be paired with Langevin dynamics to make stochastic gradient Langevin dynamics as outlined in \cite{Welling_2011_Langevin_dynamics} which provided a clear comparison in the context of classical maximum-a-posteriori (MAP) estimation.
\begin{alignat*}{2}
\text{Stochast Gradient Descent} & \quad & \Delta \vtheta_t &= \frac{\epsilon_t}{2}\left(\nabla_{\vtheta}\log p(\vtheta_t) + \tfrac{N}{n}\sum_{i=1}^n \nabla \log p(\vy_{t_i}|\vtheta_t) \right)\\
\text{Langevin Dynamics}  & \quad & \Delta \vtheta_t &= \frac{\epsilon}{2}\left(\nabla_{\vtheta}\log p(\vtheta_t) + \sum_{i=1}^N \nabla \log p(\vy_{i}|\vtheta_t) \right) + \veta_t\\
\text{SG Langevin Dynamics} & \quad &\Delta \vtheta_t &= \frac{\epsilon_t}{2}\left(\nabla_{\vtheta}\log p(\vtheta_t) + \tfrac{N}{n}\sum_{i=1}^n \nabla \log p(\vy_{t_i}|\vtheta_t) \right) + \veta_t,
\end{alignat*}
where $\epsilon_t$ is a sequence of step sizes satisfying the properties that $\sum_{t=1}^{\infty} \epsilon_t = \infty$ and $\sum_{t=1}^{\infty}\epsilon_t^2 < \infty$, and $\veta_t \sim \calN(0,\mI)$ is white noise. The set $\{\vy_1,\ldots,\vy_N\}$ denotes the training set. The constants $N$ and $n$ denote the number of training samples in the full training set and in the minibatch, respectively.
}

\subsection{(Stein's) Score Function}
Putting aside the Langevin dynamics and the underlying Fokker-Planck equation, the key subject of the iterative procedure is the gradient of the log-likelihood function $\nabla_{\vx} \log p(\vx)$. It has a formal name known as the \textbf{Stein's score function}, denoted by
\begin{equation}
\vs_{\vtheta}(\vx) \bydef \nabla_{\vx} \log p_{\vtheta}(\vx).
\end{equation}
We should be careful not to confuse Stein's score function with the \textbf{ordinary score function} which is defined as
\begin{equation}
\vs_{\vx}(\vtheta) \bydef \nabla_{\vtheta} \log p_{\vtheta}(\vx).
\end{equation}
The ordinary score function is the gradient (with respect to $\vtheta$) of the log-likelihood. In contrast, Stein's score function is the gradient with respect to the data point $\vx$. Maximum likelihood estimation uses the ordinary score function, whereas Langevin dynamics uses Stein's score function. However, since most people in the diffusion literature calls Stein's score function as the score function, we follow this culture.

\boxedeg{
If $p(x)$ is a Gaussian with $p(x) = \frac{1}{\sqrt{2\pi\sigma^2}}e^{-\frac{(x-\mu)^2}{2\sigma^2}}$, then
\begin{equation*}
s(x) = \nabla_x \log p(x) = -\frac{(x-\mu)}{\sigma^2}.
\end{equation*}
}

\boxedeg{
If $p(x)$ is a Gaussian mixture with $p(x) = \sum_{i=1}^N \pi_i \frac{1}{\sqrt{2\pi\sigma_i^2}}e^{-\frac{(x-\mu_i)^2}{2\sigma_i^2}}$, then
\begin{equation*}
s(x) = \nabla_x \log p(x) = -\frac{ \sum_{j=1}^N  \pi_j \frac{1}{\sqrt{2\pi\sigma_j^2}}e^{-\frac{(x-\mu_j)^2}{2\sigma_j^2}} \frac{(x-\mu_j)}{\sigma_j^2} }{\sum_{i=1}^N \pi_i \frac{1}{\sqrt{2\pi\sigma_i^2}}e^{-\frac{(x-\mu_i)^2}{2\sigma_i^2}}}.
\end{equation*}
}

\begin{figure}[h]
\centering
\begin{tabular}{cc}
\includegraphics[width=0.45\linewidth]{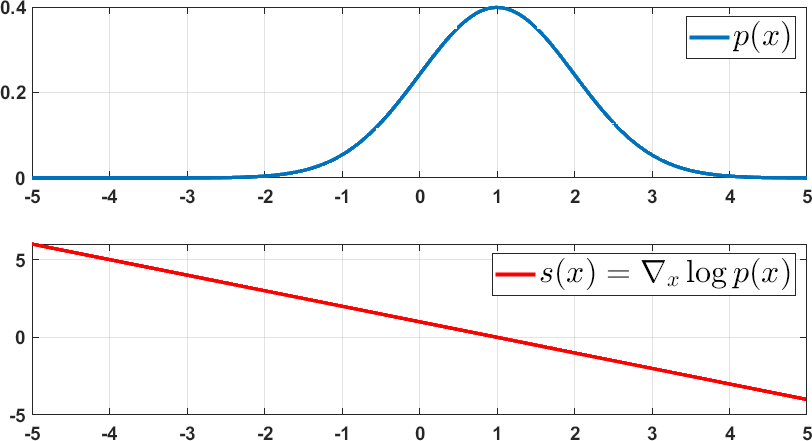}&
\includegraphics[width=0.45\linewidth]{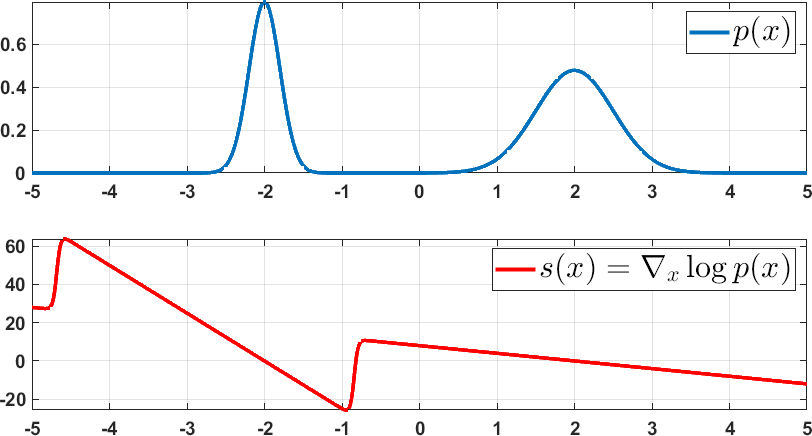}\\
(a) $\calN(1,1)$ & (b) $0.6\calN(2,0.5^2) + 0.4\calN(-2,0.2^2)$
\end{tabular}
\caption{Examples of score functions}
\label{fig: score function examples}
\end{figure}

The probability density function and the corresponding score function of the above two examples are shown in \fref{fig: score function examples}.

\textbf{Geometric Interpretations of the Score Function}. The way to understand the score function is to remember that it is the gradient with respect to the data $\vx$. For any high-dimensional distribution $p(\vx)$, the gradient will give us a vector field. There are a few useful interpretations of the score functions:
\begin{itemize}
\item The magnitude of the vectors are the strongest at places where the change of $\log p(\vx)$ is the biggest. Therefore, in regions where $\log p(\vx)$ is close to the peak will be mostly very weak gradient.
\item The vector field indicates \emph{how} a data point should travel in the contour map. In \fref{fig: contour map for score, two examples} we show the contour map of a Gaussian mixture (with two Gaussians). We draw arrows to indicate the vector field. Now if we consider a data point living the space, the Langevin dynamics equation will basically move the data points along the direction pointed by the vector field towards the basin.
\item In physics, the score function is equivalent to the ``\emph{drift}''. This name suggests how the diffusion particles should flow to the lowest energy state.
\end{itemize}

\begin{figure}[h]
\centering
\begin{tabular}{cc}
\includegraphics[width=0.4\linewidth]{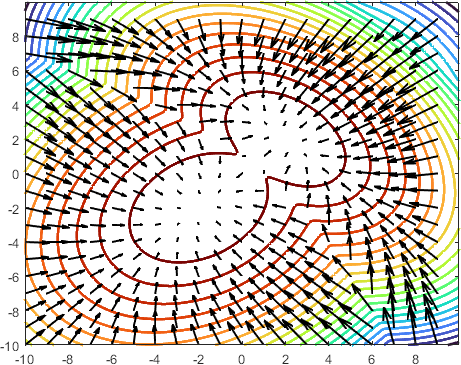}&
\includegraphics[width=0.4\linewidth]{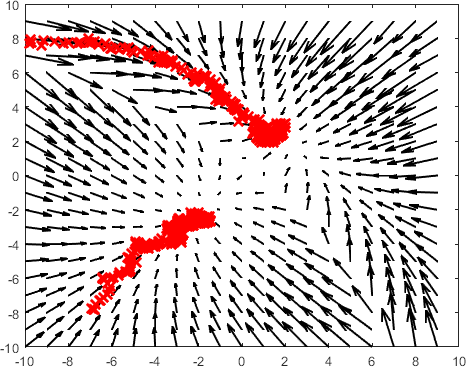}\\
(a) vector field of $\nabla_{\vx} \log p(\vx)$ & (b) $\vx_t$ trajectory
\end{tabular}
\caption{The contour map of the score function, and the corresponding trajectory of two samples.}
\label{fig: contour map for score, two examples}
\end{figure}

\subsection{Score Matching Techniques}
The most difficult question in Langevin dynamics is how to obtain $\nabla_{\vx} \log p(\vx)$ because we have no access to $p(\vx)$. In this section, we briefly discuss a few known techniques.

\begin{figure}[h]
\centering
\begin{tabular}{cc}
\includegraphics[width=0.45\linewidth]{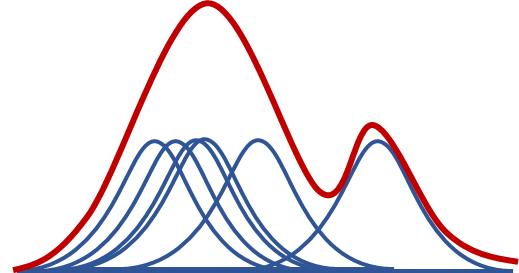}&
\includegraphics[width=0.45\linewidth]{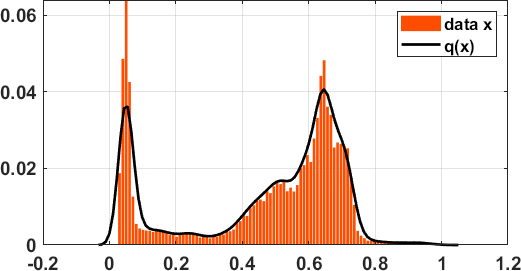}
\end{tabular}
\caption{Illustration of kernel density estimation.}
\label{fig: KDE}
\end{figure}

\textbf{Explicit Score-Matching} \cite{Vincent_2011_DSM}. Suppose that we are given a dataset $\calX = \{\vx^{(1)},\ldots,\vx^{(M)}\}$. The solution people came up with is to consider the classical kernel density estimation by defining a distribution
\begin{equation}
q_h(\vx) = \frac{1}{M} \sum_{m=1}^M \frac{1}{h} K \left(\frac{\vx-\vx^{(m)}}{h}\right),
\end{equation}
where $h$ is just some hyperparameter for the kernel function $K(\cdot)$, and $\vx^{(m)}$ is the $m$-th sample in the training set. \fref{fig: KDE} illustrates the idea of kernel density estimation. In the cartoon figure shown on the left, we show multiple kernels $K(\cdot)$ centered at different data points $\vx^{(m)}$. The sum of all these individual kernels gives us the overall kernel density estimate $q(\vx)$. On the right hand side we show a real histogram and the corresponding kernel density estimate. We remark that $q(\vx)$ is at best an approximation to the true data distribution $p(\vx)$ which is never known.

Since $q(\vx)$ is an approximation to $p(\vx)$ which is never accessible, we can learn $\vs_{\vtheta}(\vx)$ based on $q(\vx)$. This leads to the following definition of a loss function which can be used to train a network.

\boxedthm{
The \textbf{explicit score matching} loss is
\begin{align}
J_{\text{ESM}}(\vtheta)
&\bydef \frac{1}{2} \E_{p(\vx)} \|\vs_{\vtheta}(\vx) -  \nabla_{\vx} \log p(\vx)\|^2 \notag \\
&\approx \frac{1}{2} \E_{q_h(\vx)} \|\vs_{\vtheta}(\vx) -  \nabla_{\vx} \log q_h(\vx)\|^2.
\end{align}
}
By substituting the kernel density estimation, we can show that the loss is
\begin{align}
J_{\text{ESM}}(\vtheta)
&= \E_{q_h(\vx)} \|\vs_{\vtheta}(\vx) -  \nabla_{\vx} \log q_h(\vx)\|^2 \notag \\
&= \int \|\vs_{\vtheta}(\vx) -  \nabla_{\vx} \log q_h(\vx)\|^2 q_h(\vx) d\vx \notag \\
&\approx \frac{1}{M} \sum_{m=1}^M  \int \|\vs_{\vtheta}(\vx) -  \nabla_{\vx} \log q_h(\vx)\|^2 \frac{1}{h} K \left(\frac{\vx-\vx^{(m)}}{h}\right) d\vx.
\end{align}
So, we have derived a loss function that can be used to train the network. Once we train the network $\vs_{\vtheta}$, we can replace it in the Langevin dynamics equation to obtain the recursion:
\begin{equation}
\vx_{t+1} = \vx_t + \tau \vs_{\vtheta}(\vx_t) + \sqrt{2\tau}  \vz.
\end{equation}

The issue of explicit score matching is that the kernel density estimation is a fairly poor non-parameter estimation of the true distribution. Especially when we have a limited number of samples and the samples live in a high dimensional space, the kernel density estimation performance can be poor.

\vspace{4ex}
\textbf{Implicit Score Matching} \cite{Hyvarinen_2005_ScoreMatching}. In implicit score matching, the explicit score matching loss is replaced by an implicit one.
\begin{equation}
J_{\text{ISM}}(\vtheta) \bydef \E_{p(\vx)} \left[ \text{Tr}(\nabla_{\vx} \vs_{\vtheta}(\vx)) + \frac{1}{2}\| \vs_{\vtheta}(\vx) \|^2 \right],
\end{equation}
where $\nabla_{\vx} \vs_{\vtheta}(\vx)$ denotes the Jacobian of $\vs_{\vtheta}(\vx)$. The implicit score matching loss can be approximated by Monte Carlo
\begin{align*}
J_{\text{ISM}}(\vtheta) \approx \frac{1}{M}\sum_{m=1}^M \sum_i \left( \partial_i \vs_{\vtheta}(\vx^{(m)}) + \frac{1}{2}|[\vs_{\vtheta}(\vx^{(m)})]_i|^2 \right),
\end{align*}
where $\partial_i \vs_{\vtheta}(\vx^{(m)}) = \frac{\partial}{\partial x_i}[\vs_{\vtheta}(\vx)]_i = \frac{\partial^2}{\partial x_i^2}\log p(\vx)$. If the model for the score function is realized by a deep neural network, the trace operator can be difficult to compute, hence making the implicit score matching not scalable \cite{Song_2019_NeurIPS}.

\vspace{4ex}
\textbf{Denoising Score Matching}. Given the potential drawbacks of explicit and implicit score matching, we now introduce a more popular score matching known as the denoising score matching (DSM) by Vincent \cite{Vincent_2011_DSM}. In DSM, the loss function is defined as follows.

\begin{equation}
J_{\text{DSM}}(\vtheta)
\bydef \E_{q(\vx,\vx')}\left[ \frac{1}{2}\left\|\vs_{\vtheta}(\vx) - \nabla_{\vx} \log q(\vx|\vx') \right\|^2\right]
\label{eq: loss function of DSM}
\end{equation}
The key difference here is that we replace the distribution $q(\vx)$ by a conditional distribution $q(\vx|\vx')$. The former requires an approximation, e.g., via kernel density estimation, whereas the latter does not.

In the special case where $q(\vx|\vx') = \calN(\vx \;|\; \vx', \sigma^2)$, we can let $\vx = \vx' + \sigma\vz$. This will give us
\begin{align*}
\nabla_{\vx} \log q(\vx|\vx')
&= \nabla_{\vx} \log \frac{1}{(\sqrt{2\pi\sigma^2})^d} \exp\left\{- \frac{\|\vx-\vx'\|^2}{2\sigma^2}\right\}\\
&= \nabla_{\vx} \left\{ - \frac{\|\vx-\vx'\|^2}{2\sigma^2} - \log (\sqrt{2\pi\sigma^2})^d \right\}\\
&= -\frac{\vx-\vx'}{\sigma^2} = -\frac{\vz}{\sigma}.
\end{align*}
As a result, the loss function of the denoising score matching becomes
\begin{align*}
J_{\text{DSM}}(\vtheta)
&\bydef \E_{q(\vx,\vx')}\left[ \frac{1}{2}\left\|\vs_{\vtheta}(\vx) - \nabla_{\vx} \log q(\vx|\vx') \right\|^2\right]\\
&= \E_{q(\vx')}\left[ \frac{1}{2}\left\|\vs_{\vtheta}(\vx' + \sigma\vz) + \frac{\vz}{\sigma} \right\|^2\right].
\end{align*}
If we replace the dummy variable $\vx'$ by $\vx$, and we note that sampling from $q(\vx)$ can be replaced by sampling from $p(\vx)$ when we are given a training dataset, we can conclude the following.
\boxedthm{
The \textbf{Denoising Score Matching} has a loss function defined as
\begin{equation}
J_{\text{DSM}}(\vtheta) = \E_{p(\vx)}\left[ \frac{1}{2}\left\|\vs_{\vtheta}(\vx + \sigma\vz) + \frac{\vz}{\sigma} \right\|^2\right]
\label{eq: loss function of DSM Gaussian case}
\end{equation}
}

The beauty about \eref{eq: loss function of DSM Gaussian case} is that it is highly interpretable. The quantity $\vx+\sigma\vz$ is effectively adding noise $\sigma\vz$ to a clean image $\vx$. The score function $\vs_{\vtheta}$ is supposed to take this noisy image and predict the noise $\frac{\vz}{\sigma}$. Predicting noise is equivalent to denoising, because any denoised image plus the predicted noise will give us the noisy observation. Therefore, \eref{eq: loss function of DSM Gaussian case} is a \emph{denoising} step.

The following theorem, proven by Vincent \cite{Vincent_2011_DSM}, establishes the equivalence between DSM and ESM. It is this equivalence that allows us to use DSM to estimate the score function.
\boxedthm{
\label{thm: DSM and ESM}
[Vincent \cite{Vincent_2011_DSM}] For up to a constant $C$ which is independent of the variable $\vtheta$, it holds that
\begin{equation}
J_{\text{DSM}}(\vtheta) = J_{\text{ESM}}(\vtheta) + C.
\label{eq: equivalence of DSM and ESM}
\end{equation}
}

\boxedproof{
\textbf{Proof of Theorem~\ref{thm: DSM and ESM}}
The proof here is based on \cite{Vincent_2011_DSM}. We start with the explicit score matching loss function, which is given by
\begin{alignat*}{2}
J_{\text{ESM}}(\vtheta)
&= \E_{q(\vx)}\left[ \frac{1}{2} \left\| \vs_{\vtheta}(\vx) - \nabla_{\vx} \log q(\vx) \right\|^2\right]\\
&= \E_{q(\vx)}\Big[ \frac{1}{2}\left\| \vs_{\vtheta}(\vx) \right\|^2 - \vs_{\vtheta}(\vx)^T \nabla_{\vx} \log q(\vx) +
\underset{\bydef C_1, \text{independent of $\vtheta$}}{\underbrace{\frac{1}{2}\left\|\nabla_{\vx} \log q(\vx)  \right\|^2}}\Big].
\end{alignat*}

Let's zoom into the second term. We can show that
\begin{alignat*}{2}
\E_{q(\vx)}\left[ \vs_{\vtheta}(\vx)^T \nabla_{\vx} \log q(\vx) \right]
&= \int \left(\vs_{\vtheta}(\vx)^T \nabla_{\vx} \log q(\vx) \right) q(\vx) d\vx,                                &\quad &
\textcolor{purple}{\text{(expectation)}}\\
&= \int \left(\vs_{\vtheta}(\vx)^T \frac{\nabla_{\vx} q(\vx)}{ \cancel{q(\vx)}} \right) \cancel{q(\vx)} d\vx,   &\quad & \textcolor{purple}{\text{(gradient)}}\\
&= \int \vs_{\vtheta}(\vx)^T \nabla_{\vx} q(\vx) d\vx.
\end{alignat*}

Next, we consider conditioning by recalling $q(\vx) = \int q(\vx') q(\vx|\vx') d\vx'$. This will give us
\begin{alignat*}{2}
\int \vs_{\vtheta}(\vx)^T \nabla_{\vx} \textcolor{blue}{q(\vx)} d\vx
&= \int \vs_{\vtheta}(\vx)^T \nabla_{\vx}
\underset{=q(\vx)}{\underbrace{\left( \int q(\vx') q(\vx|\vx') d\vx' \right)}}
d\vx                    &\quad & \textcolor{purple}{\text{(conditional)}}\\
&= \int \vs_{\vtheta}(\vx)^T \left( \int q(\vx') \textcolor{blue}{\nabla_{\vx}} q(\vx|\vx') d\vx' \right) d\vx
&\quad & \textcolor{purple}{\text{(move gradient)}}\\
&= \int \vs_{\vtheta}(\vx)^T \left( \int q(\vx') \nabla_{\vx} q(\vx|\vx') \times \textcolor{blue}{ \frac{ q(\vx|\vx')}{q(\vx|\vx')}} d\vx' \right) d\vx
&\quad & \textcolor{purple}{\text{(multiple and divide)}}\\
&= \int \vs_{\vtheta}(\vx)^T \int q(\vx')
\underset{=\nabla_{\vx} \log q(\vx|\vx')}{\underbrace{ \left( \frac{ \nabla_{\vx} q(\vx|\vx') }{q(\vx|\vx')}\right) }}
 q(\vx|\vx')  d\vx'  d\vx
 &\quad & \textcolor{purple}{\text{(rearrange terms)}}\\
&= \int \vs_{\vtheta}(\vx)^T \left( \int q(\vx')  \Big(  \nabla_{\vx} \log q(\vx|\vx') \Big) q(\vx|\vx')  d\vx' \right) d\vx
&\quad & \\
&= \int \int \underset{=q(\vx,\vx')}{\underbrace{q(\vx|\vx') q(\vx')}}  \Big( \vs_{\vtheta}(\vx)^T  \nabla_{\vx} \log q(\vx|\vx') \Big)   d\vx' d\vx
&\quad & \textcolor{purple}{\text{(move integration)}}\\
&= \E_{q(\vx,\vx')}\left[\vs_{\vtheta}(\vx)^T  \nabla_{\vx} \log q(\vx|\vx')\right].
\end{alignat*}

So, if we substitute this result back to the definition of ESM, we can show that
\begin{align*}
J_{\text{ESM}}(\vtheta) = \E_{q(\vx)}\Big[ \frac{1}{2}\left\| \vs_{\vtheta}(\vx) \right\|^2 \Big]
- \E_{q(\vx,\vx')}\left[\vs_{\vtheta}(\vx)^T  \nabla_{\vx} \log q(\vx|\vx')\right] + C_1.
\end{align*}
Comparing this with the definition of DSM, we can observe that
\begin{align*}
J_{\text{DSM}}(\vtheta)
&\bydef \E_{q(\vx,\vx')}\left[ \frac{1}{2}\left\|\vs_{\vtheta}(\vx) - \nabla_{\vx} q(\vx|\vx') \right\|^2\right]\\
&= \E_{q(\vx,\vx')}\Big[ \frac{1}{2}\left\| \vs_{\vtheta}(\vx) \right\|^2 - \vs_{\vtheta}(\vx)^T \nabla_{\vx} \log q(\vx|\vx') +
\underset{\bydef C_2, \text{independent of $\vtheta$}}{\underbrace{\frac{1}{2}\left\|\nabla_{\vx} \log q(\vx|\vx')  \right\|^2}}\Big]\\
&= \E_{q(\vx)}\Big[ \frac{1}{2}\left\| \vs_{\vtheta}(\vx) \right\|^2 \Big]
- \E_{q(\vx,\vx')}\left[\vs_{\vtheta}(\vx)^T  \nabla_{\vx} \log q(\vx|\vx')\right] + C_2.
\end{align*}
Therefore, we conclude that
\begin{equation*}
J_{\text{DSM}}(\vtheta) = J_{\text{ESM}}(\vtheta) - C_1 + C_2.
\end{equation*}
}

\begin{figure}[h]
\centering
\includegraphics[width=0.7\linewidth]{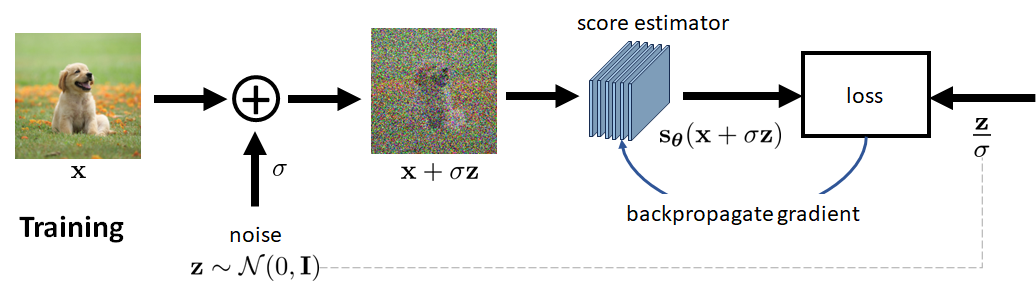}
\caption{Training of $\vs_{\vtheta}$ for denoising score matching. The network $\vs_{\vtheta}$ is trained to estimate the noise.}
\label{fig: training of s for score matching}
\end{figure}

\vspace{4ex}
The \textbf{training} procedure in a score matching model is typically done by minimizing the denoising score matching loss function. If we are given a training dataset $\{\vx^{(m)}\}_{m=1}^M$, the optimization goal is to
\begin{align*}
\vtheta^*
&= \argmin{\vtheta} \;\;  \E_{p(\vx)}\left[ \frac{1}{2}\left\|\vs_{\vtheta}(\vx + \sigma\vz) + \frac{\vz}{\sigma} \right\|^2\right] \\
&\approx \argmin{\vtheta} \;\; \frac{1}{M}\sum_{m=1}^M \frac{1}{2}\left\| \vs_{\vtheta}\left(\vx^{(m)}+\sigma\vz^{(m)} \right) + \frac{\vz^{(m)}}{\sigma}\right\|^2, \qquad\text{where}\quad \vz^{(m)} \sim \calN(0,\mI).
\end{align*}
\fref{fig: training of s for score matching} illustrates the training procedure of the score function $\vs_{\vtheta}(\vx)$.

The above training procedure assumes a fixed noise level $\sigma$. Generalizing it to multiple noise levels is not difficult. The noise conditioned score network (NCSN) by Song and Ermon \cite{Song_2019_NeurIPS} argued that one can instead optimize the following loss
\begin{equation}
J_{\text{NCSN}}(\theta) = \frac{1}{L}\sum_{i=1}^L \lambda(\sigma_{i})\ell(\vtheta;\sigma_i),
\end{equation}
where the individual loss function is defined according to the noise levels $\sigma_1,\ldots,\sigma_L$:
\begin{equation*}
\ell(\vtheta;\sigma)
= \E_{p(\vx)} \left[ \frac{1}{2}\left\|\vs_{\vtheta}(\vx + \sigma\vz) + \frac{\vz}{\sigma} \right\|^2\right].
\end{equation*}
The coefficient function $\lambda(\sigma_i)$ is often chosen as $\lambda(\sigma) = \sigma^2$ based on empirical findings \cite{Song_2019_NeurIPS}. The noise level sequence often satisfies $\tfrac{\sigma_1}{\sigma_2} = \ldots = \tfrac{\sigma_{L-1}}{\sigma_L} > 1$.

For \textbf{inference}, we assume that we have already trained the score estimator $\vs_{\vtheta}$. To generate an image, we use the Langevin equation to iteratively draw samples by denoising the image. In case of NCSN, the corresponding Langevin equation can be implemented via an annealed importance sampling:
\begin{equation*}
\vx_{t+1} = \vx_t + \frac{\alpha_i}{2}\vs_{\vtheta}(\vx_t,\sigma_i) + \sqrt{\alpha_i}\vz_t, \qquad \vz_t \sim \calN(0,\mI),
\end{equation*}
where $\alpha_i = \sigma_i^2/\sigma_L^2$ is the step size and $\vs_{\vtheta}(\vx_t,\sigma_i)$ denotes the score matching function for noise level $\sigma_i$. The iteration over $t$ is repeated sequentially for each $\sigma_i$ from $i = 1$ to $L$. For additional details of the implementation, we refer readers to Algorithm 1 of the original paper by Song and Ermon \cite{Song_2019_NeurIPS}.

\subsection{Concluding Remark}
Additional readings about score-matching should start with Vincent's technical report \cite{Vincent_2011_DSM}. A very popular paper in the recent literature is Song and Ermon \cite{Song_2019_NeurIPS}, their follow up work \cite{Song_2020_ScoreBased}, and \cite{Song_2021_SGM}. In their papers, they brought up an important discussion that when training a score function, we need a noise schedule so that the score function is trained better.

Score matching has a wide range of applicability beyond generating images. They can be used in solving many important image restoration problems such as deblurring, denoising, super-resolution, etc. Kadkhodaie and Simoncellil \cite{Kadkhodaie_2020_Implicit} is among the earlier papers to explicitly employ the score function as part of the image reconstruction process. A similar concept is presented by Kawar et al. \cite{Kawar_2022_DDRM}, where they sample from a posterior distribution which contains the forward image formation model and the prior. For problems with a better structured forward model, it has been shown that employing the ideas of proximal maps would improve the performance. This line of work can be built upon the operator splitting strategy such as the plug-and-play ADMM \cite{Chan_2017_ADMM} by extending it to Plug-and-Play diffusion model, e.g., Zhu et al. \cite{Zhu_2023_NITRE} or the generative plug and play model by Bouman and Buzzard \cite{Bouman_2023_GPnP}. Recently, it was also observed that one can directly perform the regression to mean when we do not have access to the degradation process \cite{Delbracio_2023_InDI}. Various applications papers \cite{Chung_2022_MRI, Hu_2024_Patch, Sanghvi_2024_ECCV}.

\newpage
\section{Stochastic Differential Equation (SDE)}
\setcounter{figure}{0}
\setcounter{equation}{0}

In the previous two sections we studied the diffusion models via the DDPM and the SMLD perspectives. In this section, we will study diffusion models through the lens of differential equation. As we mentioned during our discussions about the Langevin equation, the steady state solution of the Langevin equation is a random variable whose probability distribution satisfies the Fokker-Planck equation. This unusual linkage between the iterative algorithm and a differential equation makes the topic remarkably interesting. The purpose of this section is to provide the basic principles of stochastic differential equations and how they are used to understand diffusion models.

\subsection{From Iterative Algorithms to Ordinary Differential Equations}
The first and the foremost question, at least to readers with little background in differential equations, is why an iterative algorithm can be related to a differential equation. Let's understand this through two examples.

\boxedeg{
\textbf{Simple First-Order ODE}. Imagine that we are given a discrete-time algorithm with the iterations defined by the recursion:
\begin{equation}
\vx_i = \left(1- \frac{\beta \Delta t}{2}\right)\vx_{i-1}, \qquad \text{for} \;\; i = 1,2,\ldots,N,
\end{equation}
for some hyperparameter $\beta$ and a step-size parameter $\Delta t$. We can turn this iterative scheme into a continuous-time differential equation.

Suppose that there is a continuous time function $\vx(t)$. We define a discretization scheme by letting $\vx_i = \vx(\tfrac{i}{N})$ for $i=1,\ldots,N$, and $\Delta t = \tfrac{1}{N}$, and $t \in \{0,\tfrac{1}{N},\ldots,\tfrac{N-1}{N}\}$. Then the above recursion can be written as
\begin{align*}
\vx(t+\Delta t) = \left(1- \frac{\beta \Delta t}{2}\right) \vx(t).
\end{align*}
Rearranging the terms will give us
\begin{align*}
\frac{\vx(t+\Delta t) - \vx(t)}{\Delta t} = -\frac{\beta}{2} \vx(t),
\end{align*}
where at the limit when $\Delta t \rightarrow 0$, we can write the discrete equation as an ordinary differential equation (ODE)
\begin{equation}
\frac{d\vx(t)}{dt} = -\frac{\beta}{2}\vx(t).
\label{eq: ODE example 1}
\end{equation}
Not only that, we can solve for an analytic solution for the ODE where the solution is given by
\begin{equation}
\vx(t) = e^{-\frac{\beta}{2}t}.
\label{eq: ODE example 1 sol}
\end{equation}
Verification of the solution can be done by substituting \eref{eq: ODE example 1 sol} into \eref{eq: ODE example 1}.

\begin{center}
\centering
\includegraphics[width=0.5\linewidth]{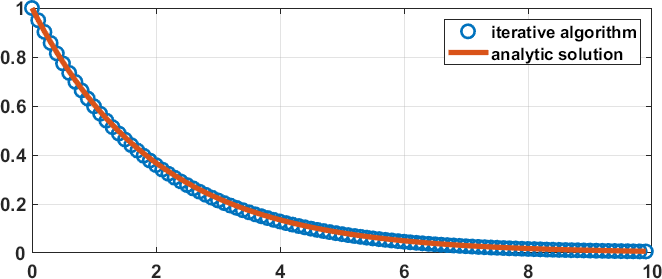}
\captionof{figure}{Analytic solution and the estimates produced by the numerical scheme.}
\end{center}

The power of the ODE is that it offers us an \emph{analytic} solution. Instead of resorting to the iterative scheme (which will take hundreds to thousands of iterations), the analytic solution tells us exactly the behavior of the solution at \emph{any} time $t$. To illustrate this fact, we show in the figure above the trajectory of the solution $\vx_1,\vx_2,\ldots,\vx_{i},\ldots,\vx_{N}$ defined by the algorithm. Here, we choose $\Delta t = 0.1$. In the same plot, we directly plot the continuous-time solution $\vx(t) = \exp\{-\beta  t/2\}$ for arbitrary $t$. As shown, the analytic solution is exactly the same as the trajectory predicted by the iterative scheme.
}

What we observe in this motivating example are two interesting facts:
\begin{itemize}
\setlength\itemsep{0ex}
\item The discrete-time iterative scheme can be written as a continuous-time ODE. In fact, many discrete-time algorithms have their associating ODEs.
\item For simple ODEs, we can write down the analytic solution in closed form. More complicated ODE would be hard to write an analytic solution. But we can still use ODE tools to analyze the behavior of the solution. We can also derive the limiting solution $t\rightarrow 0$.
\end{itemize}

\boxedeg{
\noindent\textbf{Gradient Descent}. Recall that a gradient descent algorithm for a (well-behaved) convex function $f$ is the following recursion. For $i = 1, 2, \ldots, N$, do
\begin{equation}
\vx_i = \vx_{i-1} - \beta_{i-1} \nabla f(\vx_{i-1}),
\label{eq: GD}
\end{equation}
for step-size parameter $\beta_i$. Using the same discretization as we did in the previous example, we can show that (by letting $\beta_{i-1} = \beta(t)\Delta t$):
\begin{align}
\vx_i = \vx_{i-1} - \beta_{i-1} \nabla f(\vx_{i-1})
&\qquad \Longrightarrow \qquad \vx(t+\Delta t) = \vx(t) - \beta(t)\Delta t \nabla f(\vx(t)) \notag \\
&\qquad \Longrightarrow \qquad \frac{\vx(t+\Delta t) - \vx(t)}{\Delta t} = - \beta(t) \nabla f(\vx(t)) \notag \\
&\qquad \Longrightarrow \qquad \frac{d\vx(t)}{dt} = - \beta(t) \nabla f(\vx(t)).
\label{eq: GD ODE main}
\end{align}
The ordinary differential equation shown on the right has a solution trajectory $\vx(t)$. This $\vx(t)$ is known as the \emph{gradient flow} of the function $f$.

For simplicity, we can make $\beta(t) = \beta$ for all $t$. Then there are two simple facts about this ODE. First, we can show that
\begin{alignat*}{2}
\frac{d}{dt} f(\vx(t))
&= \nabla f(\vx(t))^T \frac{d \vx(t)}{dt}               & \qquad & (\text{chain rule}) \\
&= \nabla f(\vx(t))^T \left[ -\beta \nabla f(\vx(t)) \right]  & \qquad & (\text{\eref{eq: GD ODE main}}) \\
&= - \beta \nabla f(\vx(t))^T \nabla f(\vx(t))                & \qquad & \\
&= - \beta \|\nabla f(\vx(t))\|^2 \le 0                        & \qquad & (\text{norm-squares}).
\end{alignat*}
Therefore, as we move from $\vx_{i-1}$ to $\vx_i$, the objective value $f(\vx(t))$ has to go down. This is consistent with our expectation because a gradient descent algorithm should bring the cost down as the iteration goes on. Second, at the limit when $t \rightarrow \infty$, we know that $\frac{d\vx(t)}{dt} \rightarrow 0$. Hence, $\frac{d\vx(t)}{dt} = -\beta \nabla f(\vx(t))$ will imply that
\begin{equation}
\nabla f(\vx(t)) \rightarrow 0, \qquad \text{as } t\rightarrow \infty.
\end{equation}
Therefore, the solution trajectory $\vx(t)$ will approach the minimizer for the function $f$.
}

\vspace{2ex}
\noindent\textbf{Forward and Backward Updates}.
\vspace{2ex}

Let's use the gradient descent example to illustrate one more aspect of the ODE. Going back to \eref{eq: GD}, we recognize that the recursion can be written equivalently as (assuming $\beta(t) = \beta)$:
\begin{equation}
\underset{\Delta \vx}{\underbrace{\vx_i - \vx_{i-1}}} = - \underset{\beta \Delta t}{\underbrace{\beta_{i-1}}} \nabla f(\vx_{i-1}) \;\; \Rightarrow \;\;
d\vx = - \beta \nabla f(\vx) dt,
\label{eq: GD Ito}
\end{equation}
where the continuous equation holds when we set $\Delta t \rightarrow 0$ and $\Delta \vx \rightarrow 0$. The interesting point about this equality is that it gives us a summary of the update $\Delta\vx$ by writing it in terms of $dt$. It says that if we move the along the time axis by $dt$, then the solution $\vx$ will be updated by $d\vx$.

\eref{eq: GD Ito} defines the relationship between \emph{changes}. If we consider a sequence of iterates $i = 1,2,\ldots,N$, and if we are told that the progression of the iterates follows \eref{eq: GD Ito}, then we can write
\begin{align*}
\text{(forward)} \hspace{2cm} \vx_{i} = \vx_{i-1} + \Delta \vx_{i-1}
&\approx \vx_{i-1} + d\vx \\
&= \vx_{i-1} - \nabla f(\vx_{i-1}) \beta dt\\
&\approx \vx_{i-1} - \beta_{i-1} \nabla f(\vx_{i-1}).
\end{align*}
We call this as the \textbf{forward} equation because we update $\vx$ by $\vx + \Delta \vx$ assuming that $t \leftarrow t + \Delta t$.

Now, consider a sequence of iterates $i = N, N-1,\ldots,2,1$. If we are told that the progression of the iterates follows \eref{eq: GD Ito}, then the time-reversal iterates will be
\begin{align*}
\text{(reverse)} \hspace{2cm}  \vx_{i-1}
= \vx_i - \Delta \vx_i
&\approx \vx_{i} - d\vx\\
&= \vx_i + \beta \nabla f(\vx_i)dt\\
&\approx \vx_i + \beta_i \nabla f(\vx_i).
\end{align*}
Note the change in sign when reversing the progression direction. We call this the \textbf{reverse} equation.

\subsection{What is an SDE?}
Stochastic differential equation (SDE) can be regarded as an extension to the ordinary differential equation (ODE). In this subsection, we briefly introduce the notations and explain their meanings.

\vspace{2ex}
\textbf{ODE}. We consider a first-order ODE of the following form:
\begin{equation*}
\frac{d\vx(t)}{dt} = \vf(t,\vx),
\end{equation*}
where $\vf(t,\vx)$ is some function of $t$ and $\vx$. Assuming that the initial condition is $\vx(0) = \vx_0$, the solution takes the form of
\begin{equation*}
\vx(t) = \vx_0 + \int_0^t \vf(s,\vx(s))ds.
\end{equation*}
The integral form of the solution often comes with a short-hand notation known as the \textbf{differential form}, which can be written as
\begin{equation}
d\vx = \vf(t,\vx)dt.
\end{equation}

\vspace{2ex}
\textbf{SDE}. In an SDE, in addition to a deterministic function $\vf(t,\vx)$, we consider a stochastic perturbation. For example, the stochastic perturbation can take the following form:
\begin{equation*}
\frac{d\vx(t)}{dt} = \vf(t,\vx) + \vg(t,\vx)\vxi(t), \qquad \text{where} \qquad \vxi(t) \sim \calN(0,\mI),
\end{equation*}
where $\vxi(t)$ is a noise function, e.g., white noise. We can define $d\vw= \vxi(t)dt$, where $d\vw$ is often known as the differential form of the Brownian motion. Then, the differential form of this SDE can be written as
\begin{equation}
d\vx = \vf(t,\vx) dt + \vg(t,\vx)d\vw.
\end{equation}
Because $\vxi(t)$ is random, the solution to this differential equation is also random. To be explicit about the randomness of the solution, we should interpret the differential form via the integral equation
\begin{equation*}
\vx(t,\omega) = \vx_0 + \int_0^t \vf(s,\vx(s,\omega)) ds + \int_0^t \vg(s,\vx(s,\omega)) d\vw(s,\omega),
\end{equation*}
where $\omega$ denotes the index of the state of $\vx$. Therefore, as we pick a particular state of the random process $\vw(s,\omega)$, we solve a differential equation corresponding to this particular $\omega$.

\boxedeg{
Consider the stochastic differential equation
\begin{equation*}
d\vx = a d\vw,
\end{equation*}
for some constants $a$. Based on our discussions above, the solution trajectory will take the form
\begin{equation*}
\vx(t) = \vx_0 + \int_0^t a d\vw(s) = \vx_0 + a \int_0^t \vxi(s) ds,
\end{equation*}
where the last equality uses the fact that $d\vw = \vxi(t)dt$. We can visualize the solution trajectory by numerically implementing $d\vx = a d\vw$ via the discrete-time iteration:
\begin{align*}
\vx_{i} - \vx_{i-1} = a \underset{\bydef \vz_{i-1} \sim \calN(0,\mI)}{\underbrace{(\vw_i - \vw_{i-1})}}
&\quad \Rightarrow \quad \vx_i = \vx_{i-1} + a \vz_{i-1}.
\end{align*}
For example, if $a = 0.05$, one possible trajectory of $\vx(t)$ will behave as shown below. The initial point $\vx_0 = 0$ is marked as in red to indicate that the process is moving forward in time.
\begin{center}
\includegraphics[width=0.45\linewidth]{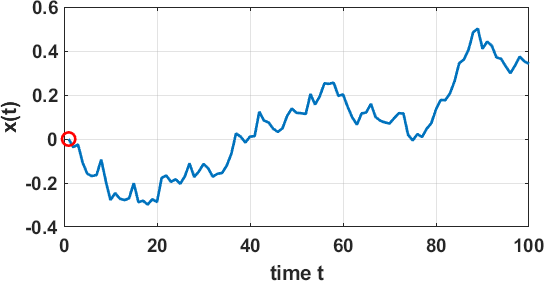}
\captionof{figure}{Trajectory of the process $d\vx = a \, d\vw$.}
\end{center}
}

\boxedeg{
Consider the stochastic differential equation
\begin{equation*}
d\vx = -\frac{\alpha}{2}\vx dt + \beta d\vw.
\end{equation*}
The solution to this problem is given by [Risken Eqn 3.7]
\begin{equation*}
\vx(t) = \vx_0e^{-\frac{\alpha}{2}t} + \beta \int_0^t e^{-\frac{\alpha}{2}(t-s)} \vxi(s)ds.
\end{equation*}
To visualize the trajectory, we consider $\alpha = 1$ and $\beta = 0.1$. The discretization will give us
\begin{align*}
\vx_{i} - \vx_{i-1} = -\frac{\alpha}{2} \vx_{i-1} + \beta (\vw_i - \vw_{i-1}) \quad \Rightarrow \quad
\vx_i = \left(1-\frac{\alpha}{2}\right) \vx_{i-1} + \beta \vz_{i-1}.
\end{align*}
\begin{center}
\includegraphics[width=0.45\linewidth]{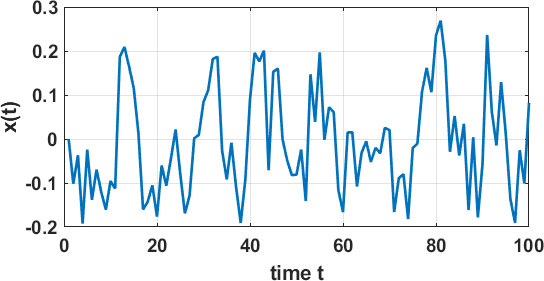}
\captionof{figure}{Trajectory of the process $d\vx = -\frac{\alpha}{2}\vx dt + \beta d\vw$.}
\end{center}
}

\subsection{Stochastic Differential Equation for DDPM and SMLD}
Without drilling into the physics of SDE (which we will do later), we want to connect SDE with the denoising diffusion probabilistic model (DDPM) and the score matching Langevin dynamics (SMLD) we studied in the previous sections.

\vspace{2ex}
\textbf{Forward and Reverse Diffusion}. Diffusion models involve a pair of equations: the forward diffusion process and the reverse diffusion process. Expressed in terms of derivatives, a forward diffusion process can be written as
\begin{equation*}
\frac{d\vx(t)}{dt} = \vf(\vx,t) + g(t) \vxi(t), \qquad \text{where} \quad \vxi(t) \sim \calN(0,\mI).
\end{equation*}
We emphasize that this is a particular diffusion process based on Brownian motion. The random perturbation $\vxi$ is assumed to be a random process with $\vxi(t)$ being an i.i.d. Gaussian random variable at all $t$. Thus, the autocorrelation function is a delta function $\E[\vxi(t)\vxi(t')] = \delta(t-t')$. For this diffusion process, we can write it in terms of the differential:
\boxeddef{
\textbf{Forward Diffusion}.
\begin{equation}
d\vx =
\underset{\text{drift}}{\underbrace{\vf (\vx,t)}} \; dt +
\underset{\text{diffusion}}{\underbrace{g(t)}} \; d\vw.
\label{eq: Ito calculus}
\end{equation}
}
Here, the differential is $d\vw = \vxi(t)dt$. This suggests that we can view $\vxi(t)$ as some rate of change (over $t$) from which the integration of $\vxi(t)dt$ will give us $d\vw$.

The two terms $\vf(\vx,t)$ and $g(t)$ carry physical meanings. The drift coefficient is a vector-valued function $\vf(\vx,t)$ defining how molecules in a closed system would move in the absence of random effects. For the gradient descent algorithm, the drift is defined by the negative gradient of the objective function. That is, we want the solution trajectory to follow the gradient of the objective. The diffusion coefficient $g(t)$ is a scalar function describing how the molecules would randomly walk from one position to another. The function $g(t)$ determines how strong the random movement is.

The reverse direction of the diffusion equation is to move backward in time. The reverse-time SDE, according to Anderson \cite{Anderson_1982_ReverseTimeDiffusion}, is given as follows.
\boxeddef{
\textbf{Reverse Diffusion}.
\begin{equation}
d \vx = [
\underset{\text{drift}}{\underbrace{\vf(\vx,t)}} -
g(t)^2
\underset{\text{score function}}{\underbrace{\nabla_{\vx} \log p_t(\vx)}}] \; dt \;\;\;\; +
\underset{\text{reverse-time diffusion}}{\underbrace{ g(t) d\overline{\vw}}},
\label{eq: SDE reverse general}
\end{equation}
where $p_t(\vx)$ is the probability distribution of $\vx$ at time $t$, and $\overline{\vw}$ is the Wiener process when time flows backward.
}

Let's briefly talk about the reverse-time diffusion. The reverse-time diffusion is nothing but a random process that proceeds in the reverse time order. So while the forward diffusion defines $\vz_i = \vw_{i+1}-\vw_i$, the reverse diffusion defines $\vz_i = \vw_{i-1} - \vw_{i}$. The following is an example.
\boxedeg{
Consider the reverse diffusion equation
\begin{equation}
d\vx = a d\overline{\vw}.
\end{equation}
We can write the discrete-time recursion as follows. For $i = N, N-1, \ldots, 1$, do
\begin{align*}
\vx_{i-1} = \vx_i + a \underset{=\vz_i}{\underbrace{(\vw_{i-1}-\vw_{i})}} = \vx_i + a \vz_i, \quad \vz_i \sim \calN(0,\mI).
\end{align*}
In the figure below we show the trajectory of this reverse-time process. Note that the initial point marked in red is at $\vx_N$. The process is tracked backward to $\vx_0$.
\begin{center}
\includegraphics[width=0.45\linewidth]{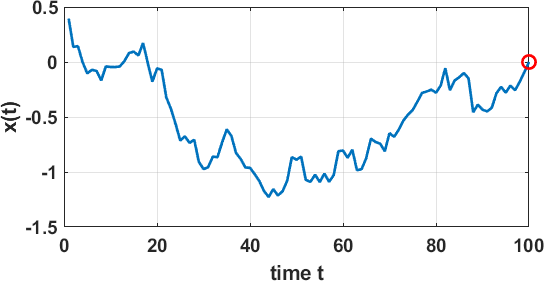}
\captionof{figure}{Trajectory of the reverse diffusion $d\vx = a d\overline{\vw}$.}
\end{center}
}

\vspace{4ex}

\textbf{Stochastic Differential Equation for DDPM}. In order to draw the connection between DDPM and SDE, we consider the discrete-time DDPM forward iteration. For $i = 1,2,\ldots,N$:
\begin{equation}
\vx_i = \sqrt{1-\beta_i} \vx_{i-1} + \sqrt{\beta_i}\vz_{i-1}, \qquad \vz_{i-1} \sim \calN(0,\mI).
\end{equation}
We can show that this equation can be derived from the forward SDE equation below.
\boxedthm{
The forward sampling equation of \textbf{DDPM} can be written as an SDE via
\begin{equation}
d\vx =
\underset{=\vf(\vx,t)}{\underbrace{-\frac{\beta(t)}{2}\; \vx }}
\; dt +
\underset{=g(t)}{\underbrace{\sqrt{\beta(t)}}}d \vw.
\end{equation}
}

\boxedproof{
\textbf{Proof}. We define a step size $\Delta t = \tfrac{1}{N}$, and consider an auxiliary noise level $\{\overline{\beta}_i\}_{i=1}^N$ where $\beta_i = \tfrac{\overline{\beta}_i}{N}$. Then
\begin{equation*}
\beta_i =
\underset{\overline{\beta}_i}{\underbrace{\beta\left( \tfrac{i}{N} \right)}} \cdot \frac{1}{N} = \beta(t+\Delta t) \Delta t,
\end{equation*}
where we assume that in the $N \rightarrow \infty$, $\overline{\beta}_i \rightarrow \beta(t)$ which is a continuous time function for $0 \le t \le 1$. Similarly, we define
\begin{align*}
\vx_i = \vx\left(\tfrac{i}{N}\right) = \vx(t+\Delta t), \quad \vz_i = \vz\left(\tfrac{i}{N}\right) = \vz(t+\Delta t).
\end{align*}
Hence, we have
\begin{alignat*}{2}
&\quad          \qquad & \vx_i            &= \sqrt{1-\beta_i} \vx_{i-1} + \sqrt{\beta_i} \vz_{i-1} \\
&\Rightarrow    \qquad & \vx_i            &= \sqrt{1-\tfrac{\overline{\beta}_i}{N}} \vx_{i-1} + \sqrt{\tfrac{\overline{\beta}_i}{N}} \vz_{i-1} \\
&\Rightarrow    \qquad & \vx(t+\Delta t)  &= \sqrt{1-\beta(t+\Delta t) \cdot \Delta t} \; \vx(t) + \sqrt{\beta(t+\Delta t) \cdot \Delta t} \; \vz(t)\\
&\Rightarrow    \qquad & \vx(t+\Delta t)  &\approx \left( 1-\frac{1}{2}\beta(t+\Delta t) \cdot \Delta t \right) \; \vx(t) + \sqrt{\beta(t+\Delta t) \cdot \Delta t} \; \vz(t)\\
&\Rightarrow    \qquad & \vx(t+\Delta t)  &\approx \vx(t) - \frac{1}{2}\beta(t)\Delta t \; \vx(t) + \sqrt{\beta(t) \cdot \Delta t} \; \vz(t).
\end{alignat*}
Thus, as $\Delta t \rightarrow 0$, we have
\begin{equation}
d\vx = -\frac{1}{2}\beta(t)\vx dt + \sqrt{\beta(t)} \; d\vw.
\end{equation}
Therefore, we showed that the DDPM forward update iteration can be equivalently written as an SDE.
}

Being able to write the DDPM forward update iteration as an SDE means that the DDPM estimates can be determined by solving the SDE. In other words, for an appropriately defined SDE solver, we can throw the SDE into the solver. The solution returned by an appropriately chosen solver will be the DDPM estimate. Of course, we are not required to use the SDE solver because the DDPM iteration itself is solving the SDE. It may not be the best SDE solver because the DDPM iteration is only a first order method. Nevertheless, if we are not interested in using the SDE solver, we can still use the DDPM iteration to obtain a solution. Here is an example.

\boxedeg{
Consider the DDPM forward equation with $\beta_i = 0.05$ for all $i = 0,\ldots,N-1$. We initialize the sample $\vx_0$ by drawing it from a Gaussian mixture such that
\begin{equation*}
\vx_0 \sim \sum_{k=1}^K \pi_k \calN(\vx_0|\vmu_k,\sigma_k^2\mI),
\end{equation*}
where $\pi_1 = \pi_2 = 0.5$, $\sigma_1 = \sigma_2 = 1$, $\vmu_1 = 3$ and $\vmu_2 = -3$. Then, using the equation
\begin{equation*}
\vx_i = \sqrt{1-\beta_i} \vx_{i-1} + \sqrt{\beta_i}\vz_{i-1}, \qquad \vz_{i-1} \sim \calN(0,\mI),
\end{equation*}
we can plot the trajectory and the distribution as follows.
\begin{center}
\includegraphics[width=0.4\linewidth]{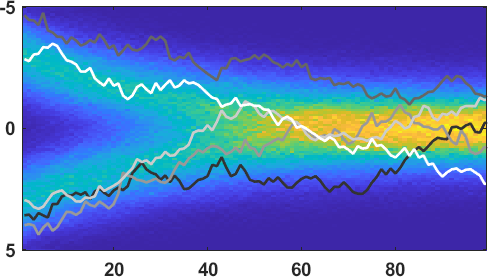}
\captionof{figure}{Realizations of the trajectories of $\vx_t$, starting with a Gaussian mixture and ending with a single Gaussian.}
\end{center}
}

The reverse diffusion equation follows from \eref{eq: SDE reverse general} by substituting the appropriate quantities: $\vf(\vx,t) = -\frac{\beta(t)}{2}$ and $g(t) = \sqrt{\beta(t)}$. This will give us
\begin{align*}
d\vx
&= [\vf(\vx,t) - g(t)^2 \nabla_{\vx}\log p_t(\vx)] dt + g(t)d\overline{\vw}\\
&= \left[-\frac{\beta(t)}{2}\; \vx - \beta(t) \nabla_{\vx}\log p_t(\vx)\right] dt + \sqrt{\beta(t)}d\overline{\vw},
\end{align*}
which will give us the following equation:
\boxedthm{
The reverse sampling equation of \textbf{DDPM} can be written as an SDE via
\begin{equation}
d\vx  = -\beta(t)\left[ \frac{\vx}{2} + \nabla_{\vx}\log p_t(\vx) \right] dt + \sqrt{\beta(t)}d\overline{\vw}.
\label{eq: SDE reverse DDPM}
\end{equation}
}

\boxedproof{
\textbf{Proof}. The iterative update scheme can be written by considering $d\vx = \vx(t) - \vx(t-\Delta t)$, and $d\overline{\vw} = \vw(t-\Delta t) - \vw(t) = -\vz(t)$. Then, letting $dt = \Delta t$, we can show that
\begin{alignat*}{2}
&            &  \vx(t) - \vx(t-\Delta t) &= -\beta(t)\Delta t \left[\frac{\vx(t)}{2} + \nabla_{\vx} \log p_t(\vx(t)) \right] - \sqrt{\beta(t)\Delta t} \vz(t)\\
&\Rightarrow \quad &  \vx(t-\Delta t)   &= \vx(t)+\beta(t)\Delta t \left[\frac{\vx(t)}{2} + \nabla_{\vx} \log p_t(\vx(t)) \right] + \sqrt{\beta(t)\Delta t} \vz(t).
\end{alignat*}
By grouping the terms, and assuming that $\beta(t) \Delta t \ll 1$, we recognize that
\begin{alignat*}{2}
\vx(t-\Delta t)
&= \vx(t)\left[1 + \frac{\beta(t)\Delta t}{2} \right] + \beta(t)\Delta t \nabla_{\vx} \log p_t(\vx(t)) + \sqrt{\beta(t)\Delta t} \vz(t) & \quad & \\
&\approx \vx(t)\left[1 + \frac{\beta(t)\Delta t}{2} \right] + \beta(t)\Delta t \nabla_{\vx} \log p_t(\vx(t)) + \textcolor{blue}{ \tfrac{(\beta(t)\Delta t)^2}{2} \nabla_{\vx} \log p_t(\vx(t))} +  \sqrt{\beta(t)\Delta t} \vz(t) & \quad & \\
&= \left[1 + \frac{\beta(t)\Delta t}{2} \right]\Big( \vx(t) + \beta(t)\Delta t \nabla_{\vx} \log p_t(\vx(t)) \Big) + \sqrt{\beta(t)\Delta t} \vz(t). & \quad &
\end{alignat*}
Then, following the discretization scheme by letting $t \in \{0,\ldots,\frac{N-1}{N}\}$, $\Delta t = 1/N$, $\vx(t-\Delta t) = \vx_{i-1}$, $\vx(t) = \vx_i$, and $\beta(t)\Delta t = \beta_i$, we can show that
\begin{align}
\vx_{i-1}
&= (1+\tfrac{\beta_i}{2})\Big[ \vx_i + \tfrac{\beta_i}{2} \nabla_{\vx} \log p_i(\vx_i) \Big] + \sqrt{\beta_i} \vz_i \notag \\
&\approx \tfrac{1}{\sqrt{1-\beta_i}} \Big[ \vx_i + \tfrac{\beta_i}{2} \nabla_{\vx} \log p_i(\vx_i) \Big] + \sqrt{\beta_i} \vz_i,
\end{align}
where $p_i(\vx)$ is the probability density function of $\vx$ at time $i$. For practical implementation, we can replace $\nabla_{\vx} \log p_i(\vx_i)$ by the estimated score function $\vs_{\vtheta}(\vx_i)$.
}

So, we have recovered the DDPM iteration that is consistent with the one defined by Song and Ermon in \cite{Song_2021_SGM}. This is an interesting result, because it allows us to connect DDPM's iteration using the score function. Song and Ermon \cite{Song_2021_SGM} called the SDE an \textbf{variance preserving} (VP) SDE.

\boxedeg{
Following from the previous example, we perform the reverse diffusion equation using
\begin{equation*}
\vx_{i-1} = \tfrac{1}{\sqrt{1-\beta_i}} \Big[ \vx_i + \tfrac{\beta_i}{2} \nabla_{\vx} \log p_i(\vx_i) \Big] + \sqrt{\beta_i} \vz_i,
\end{equation*}
where $\vz_i \sim \calN(0,\mI)$. The trajectory of the iterates is shown below.
\begin{center}
\includegraphics[width=0.4\linewidth]{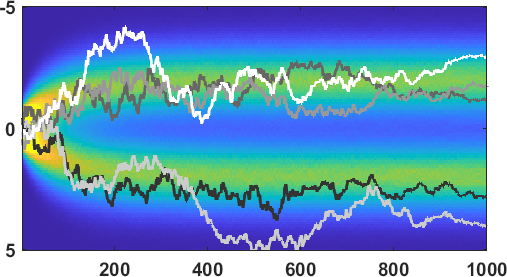}
\captionof{figure}{Realizations of the trajectories of $\vx_t$, starting with a single Gaussian and ending with a Gaussian mixture.}
\end{center}
}

\vspace{4ex}
\textbf{Stochastic Differential Equation for SMLD}. The score-matching Langevin Dynamics model can also be described by an SDE. To start with, we notice that in the SMLD setting, there isn't really a ``forward diffusion step''. However, we can roughly argue that if we divide the noise scale in the SMLD training into $N$ levels, then the recursion should follow a Markov chain
\begin{equation}
\vx_i = \vx_{i-1} + \sqrt{\sigma_i^2 - \sigma_{i-1}^2} \vz_{i-1}, \qquad i = 1,2,\ldots,N.
\label{eq: SMLD recursion forward}
\end{equation}
This is not too hard to see. If we assume that the variance of $\vx_{i-1}$ is $\sigma_{i-1}^2$, then we can show that
\begin{align*}
\Var[\vx_i]
&= \Var[\vx_{i-1}] + (\sigma_i^2 - \sigma_{i-1}^2) \\
&= \sigma_{i-1}^2 + (\sigma_i^2 - \sigma_{i-1}^2) = \sigma_i^2.
\end{align*}
Therefore, given a sequence of noise levels, \eref{eq: SMLD recursion forward} will indeed generate estimates $\vx_i$ such that the noise statistics will satisfy the desired property.

If we agree \eref{eq: SMLD recursion forward}, it is easy to derive the SDE associated with \eref{eq: SMLD recursion forward}. We summarize our results as follows.

\boxedthm{
The forward sampling equation of \textbf{SMLD} can be written as an SDE via
\begin{equation}
d\vx = \sqrt{ \frac{d[\sigma(t)^2]}{dt}} \; d\vw.
\end{equation}
}
\boxedproof{
\textbf{Proof}. Assuming that in the limit $\{\sigma_i\}_{i=1}^N$ becomes the continuous time $\sigma(t)$ for $0 \le t \le 1$, and $\{\vx_i\}_{i=1}^N$ becomes $\vx(t)$ where $\vx_i = \vx(\tfrac{i}{N})$ if we let $t \in \{0,\tfrac{1}{N},\ldots,\tfrac{N-1}{N}\}$. Then we have
\begin{align*}
\vx(t+\Delta t)
&= \vx(t) + \sqrt{ \sigma(t+\Delta t)^2 - \sigma(t)^2 } \vz(t) \\
&\approx \vx(t) + \sqrt{ \frac{d[\sigma(t)^2]}{dt} \Delta t} \; \vz(t).
\end{align*}
At the limit when $\Delta t \rightarrow 0$, the equation converges to
\begin{equation*}
d\vx = \sqrt{ \frac{d[\sigma(t)^2]}{dt}} \; d\vw.
\end{equation*}
}

Mapping this to \eref{eq: SDE reverse general}, we derive the reverse-time diffusion equation.
\boxedthm{
The reverse sampling equation of \textbf{SMLD} can be written as an SDE via
\begin{equation}
d\vx  = - \left( \frac{d[\sigma(t)^2]}{dt} \nabla_{\vx} \log p_t(\vx)\right)dt + \sqrt{ \frac{d[\sigma(t)^2]}{dt} } \; d\overline{\vw}.
\end{equation}
}
\boxedproof{
\textbf{Proof}. We recognize that
\begin{align*}
\vf(\vx,t) = 0, \qquad \text{and} \qquad g(t) = \sqrt{ \frac{d[\sigma(t)^2]}{dt} }.
\end{align*}
As a result, if we write the reverse equation \eref{eq: SDE reverse general}, we should have
\begin{align*}
d\vx
&= [\vf(\vx,t) - g(t)^2 \nabla_{\vx} \log p_t(\vx)] \; dt \;\; + \;\; g(t) d\overline{\vw}\\
&= - \left( \frac{d[\sigma(t)^2]}{dt} \nabla_{\vx} \log p_t(\vx(t))\right)dt + \sqrt{ \frac{d[\sigma(t)^2]}{dt} } \; d\overline{\vw}.
\end{align*}
}

For the discrete-time iterations, we first define $\alpha(t) = \frac{d[\sigma(t)^2]}{dt}$. Then, using the same set of discretization setups as the DDPM case, we can show that
\begin{alignat}{2}
& \quad             & \vx(t+\Delta t) - \vx(t) &= - \Big( \alpha(t)  \nabla_{\vx} \log p_t(\vx)\Big) \Delta t - \sqrt{\alpha(t)\Delta t} \; \vz(t) \notag\\
& \Rightarrow \quad & \vx(t)    &= \vx(t+\Delta t) + \alpha(t)\Delta t \nabla_{\vx} \log p_t(\vx) + \sqrt{\alpha(t)\Delta t} \; \vz(t) \notag\\
& \Rightarrow \quad & \vx_{i-1} &= \vx_i + \alpha_i \nabla_{\vx} \log p_i(\vx_i) + \sqrt{\alpha_i} \; \vz_i \\
& \Rightarrow \quad & \vx_{i-1} &= \vx_i + (\sigma_i^2-\sigma_{i-1}^2) \nabla_{\vx} \log p_i(\vx_i) + \sqrt{(\sigma_i^2-\sigma_{i-1}^2)} \; \vz_i, \notag
\end{alignat}
which is identical to the SMLD reverse update equation. Song and Ermon \cite{Song_2021_SGM} called the SDE an \textbf{variance exploding} (VE) SDE.

\vspace{2ex}
\textbf{Equivalence between VP and VE's Inference}. As we have just seen, DDPM and SMLD correspond to two variants of the stochastic differential equation: the variance exploding (VE) and the variance preserving (VP). An observation made by Kawar et al. \cite{Kawar_2022_DDRM} was that the \emph{inference} process determined by VP and VE are equivalent. Therefore, if we want to use a diffusion model as part of another task such as image restoration, it does not matter if we use VE or VP. Note, however, that the specific choice of hyperparameters will still affect the \emph{training} and hence the model itself.

\subsection{Numerical Solvers for ODE and SDE}
SDE and ODE are not easy to solve analytically. In many situations, these differential equations are solved numerically. In this subsection, we briefly highlight the overall idea of these methods.

To make our discussion slightly easier, we shall focus on a simple ODE:
\begin{equation}
\frac{d\vx(t)}{dt} = \vf(t,\vx(t)).
\end{equation}
Geometrically, this ODE means that the \emph{derivative} of $\vx(t)$ equals to the function $\vf(t, \vx)$ evaluated at $\vx(t)$. For this to happen, we need the slope of $\vx(t)$ to match with the functional value. In the 1D case, the above geometric interpretation can be translated to the following equation:
\begin{align*}
\frac{x(t) - x(t_0)}{t-t_0} = f(t_0, x_0),
\end{align*}
where $t_0$ is a time fairly close to $t$. By rearranging the terms, we see that this equation is equivalent to
\begin{equation*}
x(t) = x(t_0) + f(t_0,x_0) (t-t_0).
\end{equation*}
So we have recovered something very similar to gradient descent. This is the Euler method.

\vspace{2ex}
\textbf{Euler Method}. The Euler method is a first-order numerical method for solving the ODE. Given $\frac{dx(t)}{dt}= f(t, x)$, and $x(t_0) = x_0$, Euler method solves the problem via an iterative scheme for $i = 0, 1, \ldots, N-1$ such that
\begin{align*}
x_{i+1} = x_i + \alpha \cdot f(t_i, x_i), \qquad\qquad  i= 0, 1, \ldots, N-1,
\end{align*}
where $\alpha$ is the step size. Let's consider a simple example.

\boxedeg{
\cite[Example 2.2]{Atkinson_2009_Book} Consider the following ODE
\begin{equation*}
\frac{dx(t)}{dt} = \frac{x(t) + t^2 - 2}{t+1}.
\end{equation*}
If we apply the Euler method with a step size $\alpha$, then the iteration will take the form
\begin{align*}
x_{i+1} = x_i + \alpha \cdot f(t_i, x_i) = x_i + \alpha \cdot \frac{(x_i+t_i^2-2)}{t_i+1}.
\end{align*}
}

\textbf{Runge-Kutta (RK) Method}. Another popularly used ODE solver is the Runge-Kutta (RK) method. The classical RK-4 algorithm solves the ODE via the iteration
\begin{align*}
x_{i+1} = x_i + \frac{\alpha}{6} \cdot \Big(k_1 + 2k_2 + 2k_3 + k_4\Big), \qquad i = 1, 2, \ldots, N,
\end{align*}
where the quantities $k_1$, $k_2$, $k_3$ and $k_4$ are defined as
\begin{align*}
k_1 &= f(x_i, t_i),\\
k_2 &= f\left(t_i + \tfrac{\alpha}{2}, \;\; x_i + \alpha \tfrac{k_1}{2}\right),\\
k_3 &= f\left(t_i + \tfrac{\alpha}{2}, \;\; x_i + \alpha \tfrac{k_2}{2}\right),\\
k_4 &= f\left(t_i + \alpha,            \;\; x_i + \alpha k_3 \right).
\end{align*}
For more details, you can consult numerical methods textbooks such as \cite{Atkinson_2009_Book}.

\begin{figure}[h]
\centering
\includegraphics[width=0.4\linewidth]{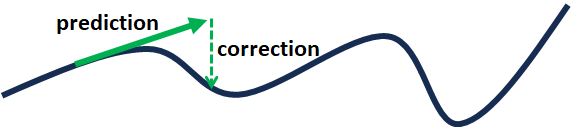}
\caption{Prediction and correction algorithm.}
\label{fig: Prediction_Correction}
\end{figure}

\vspace{2ex}
\textbf{Predictor-Corrector Algorithm} \cite{Song_2021_SGM}. Since different numerical solvers have different behavior in terms of the error of approximation, throwing the ODE (or SDE) into an off-the-shelf numerical solver will result in various degrees of error \cite{Karras_2022_Elucidating}. However, if we are specifically trying to solve the reverse diffusion equation, it is possible to use techniques other than numerical ODE/SDE solvers to make the appropriate corrections, as illustrated in \fref{fig: Prediction_Correction}.

Let's use DDPM as an example. In DDPM, the reverse diffusion equation is given by
\begin{equation*}
\vx_{i-1} = \tfrac{1}{\sqrt{1-\beta_i}} \Big[ \vx_i + \tfrac{\beta_i}{2} \nabla_{\vx} \log p_i(\vx_i) \Big] + \sqrt{\beta_i} \vz_i.
\end{equation*}
We can consider it as an Euler method for the reverse diffusion. However, if we have already trained the score function $\vs_{\vtheta}(\vx_{i},i)$, we can run the score-matching equation, i.e.,
\begin{equation*}
\vx_{i-1} = \vx_i + \epsilon_i \vs_{\vtheta}(\vx_{i},i) + \sqrt{2\epsilon_i} \vz_i,
\end{equation*}
for $M$ times to make the correction. The algorithm below summarizes the idea. (Note that we have replaced the score function by the estimate.)

\begin{algorithm}[h]
\label{alg: Prediction Correction}
\caption{Prediction Correction Algorithm for DDPM. \cite{Song_2021_SGM}}
\begin{algorithmic}
\STATE $\vx_N = \calN(0,\mI)$.
\FOR{$i=N-1,\ldots,0$}
    \STATE
    \begin{equation}
    (\text{Prediction}) \qquad \vx_{i-1} = \tfrac{1}{\sqrt{1-\beta_i}} \Big[ \vx_i + \tfrac{\beta_i}{2} \vs_{\vtheta}(\vx_{i},i) \Big] + \sqrt{\beta_i} \vz_i.
    \end{equation}
    \FOR{$m=1,\ldots,M$}
        \STATE \begin{equation}
                \hspace{-1.8cm}(\text{Correction}) \qquad  \vx_{i-1} = \vx_i + \epsilon_i \vs_{\vtheta}(\vx_{i},i) + \sqrt{2\epsilon_i} \vz_i,
                \end{equation}
    \ENDFOR
\ENDFOR
\end{algorithmic}
\end{algorithm}

For the SMLD algorithm, the two equations are:
\begin{alignat*}{2}
\vx_{i-1} &= \vx_i + (\sigma_{i}^2-\sigma_{i-1}^2)\vs_{\vtheta}(\vx_i,\sigma_i) + \sqrt{\sigma_i^2-\sigma_{i-1}^2}\vz &\quad & \text{Prediction},\\
\vx_{i-1} &= \vx_i + \epsilon_i \nabla_{\vx}\vs_{\vtheta}(\vx_i,\sigma_i) + \sqrt{\epsilon_i} \; \vz                  &\quad & \text{Correction}.
\end{alignat*}
We can pair them up as in the case of DDPM's prediction-correction algorithm by repeating the correction iteration a few times.

\subsection{Concluding Remark}
The connection between the iterative algorithms with stochastic differential equations is not only a discovery that has its own scientific value, the finding has practical utilities as commented by some papers. By unifying multiple diffusion models to the same SDE framework, one can compare the algorithms. In some cases, one can improve the numerical scheme by borrowing ideas from the SDE literature as well as the probabilistic sampling literature. For example, the predictor-corrector scheme in \cite{Song_2021_SGM} was a hybrid SDE solver coupled with a Markov Chain Monte Carlo. Mapping the diffusion iterations to SDE, according to some papers such as \cite{Albergo_2023_Interpolant}, offers more design flexibility. Outside the context diffusion algorithms, in general stochastic gradient descent algorithms have corresponding SDE such as the Fokker-Planck equations. People have demonstrated how to theoretically analyze the limiting distribution of the estimates, in exact closed-form. This alleviates the difficulty of analyzing the random algorithm by means of analyzing a well-defined limiting distribution.

There is a growing interest in developing accelerated SDE solvers. Lu et al. \cite{Lu_2022_DPMsolver} showed that the typical SDE we encounter in diffusion models can be decomposed into a semi-linear term and a nonlinear term. Since the semi-linear term has a closed-form solution, one just needs to use a specialized numerical solver to handle the nonlinear term. It was further showed that DDIM can be considered as an accelerated SDE solver. Other approaches to accelerate the SDE solver is by means of knowledge distillation \cite{Salimans_2022_Distillation, Meng_2023_Distillation}.

\newpage
\section{Langevin and Fokker-Planck Equations}
\setcounter{figure}{0}
\setcounter{equation}{0}

Many recent papers on diffusion models are focusing on either improving the image generation quality (by generalizing it to more semantically meaningful images) or applying it to problems in new domains (e.g. medical data). To help beginner readers understand better the foundations of these applications, we want to go backward in time by discussing the \emph{physics} of the stochastic differential equations associated with the diffusion models. By studying the fundamental principles of these SDEs, we want to gain a deeper understanding of where these equations are coming from.

More specifically, we are interested in studying two major sets of equations related to diffusion: the Langevin equation and the Fokker-Planck equation. There are several goals of this section. First, we want to discuss the origin of the Langevin equation, and explain why is it related to diffusion models. We want to explain the general properties of a Markovian process, and its associated differential equations. Finally, we want to show how the Fokker-Planck equation is derived, and discuss why it plays an important role in diffusion models.

\subsection{Brownian Motion}
\textbf{Historical Perspective}. In 1827, botanist Robert Brown observed a phenomenon that small pollen grains have irregular motion when placed in water \cite{Brown_1828}. The motion of the pollen grains is later named the \emph{Brownian motion}. The explanation of the Brownian motion was made by Albert Einstein in 1905 \cite{Einstein_1905} and independently by Marian Smoluchowski \cite{Smoluchowski_1906} around the same time. Einstein's main argument was that the motion of the pollen grains is caused by the impact of the water molecules. However, since there are trillions of molecules in the system and we never know the initial state of individual molecules, it is nearly impossible to use classical analysis to study the microscope states. Einstein showed that the mean squared displacement can be related to a diffusion coefficient, and so he introduced a probabilistic approach by considering the statistical behavior of the molecules\footnote{A remark here is that Andrey Kolmogorov's probability book wasn't published until 1933 \cite{Kolmogorov_1933}. So back in 1905, the notion of probability theory was in its infancy.}. Einstein's theoretical prediction was then empirically confirmed by Jean-Baptiste Perrin who later won the Nobel Prize for Physics in 1926. A few years later since Einstien's paper, in 1908, French physicist Paul Langevin constructed a random Markovian force to describe the collision and interactions of the particles. This then led to the development of a partial differential equation by Dutch physicist Adriaan Fokker in 1914 and German physicist Max Planck in 1917, which is now known as the Fokker-Planck equation. The Kramers-Moyal expansion was introduced by Hans Kramers in 1940 and José Enrique Moyal in 1949, which showed a Taylor expansion technique to describe the time evolution of a probability distribution.

\vspace{2ex}
\textbf{Derivation of Brownian Motion}. So, what is Brownian motion and how is it related to diffusion models? Assume that there is a particle suspended in fluid. Stoke's law states that the friction applied to the particle is given by
\begin{equation}
F(t) = -\alpha v(t),
\end{equation}
where $F$ is the friction, $v$ is the velocity, and $\alpha = 6\pi \mu R$. Here, $R$ is the radius of the particle, and $\mu$ is the viscosity of the fluid. By Newton's second law, we further know that $F(t) = m \dot{v}(t)$, where $m$ is the mass of the particle. Equating the two equations
\begin{align*}
\begin{cases}
F(t) &= -\alpha v(t)\\
F(t) &= m  \frac{dv(t)}{dt},
\end{cases}
\end{align*}
we will obtain the following differential equation
\begin{align*}
 m \frac{dv(t)}{dt} + \alpha v(t) = 0.
\end{align*}
By defining $\gamma \bydef \tfrac{\alpha}{m}$, we can simplify the above equation as
\begin{equation}
\frac{dv(t)}{dt} + \gamma v(t) = 0.
\end{equation}

The above deterministic equation is accurate when the particle is significantly more massive than the fluid molecules. The reason is that, by conservation of momentum, the massive particle will gain little velocity when it collides with the fluid molecule. However, since pollen grains are so light, the bombardment of water molecules will cause them to accelerate in a small but non-negligible way. This will create a random fluctuation. The total force of the water molecule acting on the particle is then modified to
\begin{equation*}
F(t) = -\alpha v(t) + F_f(t),
\end{equation*}
where $F_f(t)$ is a stochastic term. This will then give us a modified differential equation
\begin{equation*}
m \frac{d v(t)}{dt} = -\alpha v(t) + F_f(t).
\end{equation*}
By defining $\Gamma(t) = F_f(t)/m$, we arrive at a new differential equation
\begin{equation}
\frac{d v(t)}{dt} + \gamma v(t) = \Gamma(t),
\end{equation}
which can be written in terms of a short-hand notation
\begin{equation}
\dot{v} + \gamma v = \Gamma(t).
\label{eq: Brownian main}
\end{equation}

In \eref{eq: Brownian main}, the random process $\Gamma(t)$ represents a stochastic force known as the \textbf{Langevin force}. It satisfies two properties:
\begin{enumerate}
\setlength\itemsep{0ex}
\item[(i)] $\E[\Gamma(t)] = 0$ for all $t$ so that its mean function is a constant zero;
\item[(ii)] $\E[\Gamma(t)\Gamma(t')] = q \delta(t-t')$ for all $t$ and $t'$, i.e., the autocorrelation function is a delta function with an amplitude $q$.
\end{enumerate}

\boxedproof{
\textbf{Remark}. Properties (i) and (ii) are special cases of a wide sense stationary process. A wide sense stationary process is a random process that has a constant mean function (the constant is not necessarily zero), and that the autocorrelation function $R(t,t') \bydef \E[\Gamma(t)\Gamma(t')]$ is a function of the difference $t-t'$. This function is not necessarily a delta function. For example $R(t,t') = e^{-|t-t'|}$ can be a valid autocorrelation function of a wide sense stationary process.

A random process satisfying properties (i) and (ii) are sometime called a delta-correlated process in the statistical mechanics literature. There are different ways to construct a delta-correlated process. For example, we can assume $\Gamma(t) \sim \calN(0,1)$ for every $t$, or any other independent and identically distributions defined in the same way. Gaussian distributions are more often used because many physical phenomena can be described by a Gaussian, e.g., thermal noise. A Gaussian random process satisfying properties (i) and (ii) is called a Gaussian white noise.

For any wide sense stationary process, \textbf{Wiener-Khinchin Theorem} says that the \emph{power spectral density} can be defined through the Fourier transform of the autocorrelation function. More specifically, if $R(\tau) = \E[\Gamma(t+\tau)\Gamma(t)]$ is the autocorrelation function (we can write $R(t,t')$ as $R(\tau)$ if $\Gamma(t)$ is a wide sense stationary process), Wiener-Khinchin Theorem states that the power spectral density is
\begin{equation}
S(\omega) = \int_{-\infty}^{\infty} R(\tau) e^{-j\omega\tau} d\tau.
\end{equation}
So if $R(\tau)$ is a delta function, $S(\omega)$ will have a constant value for all $\omega$.
}

\boxedproof{
\textbf{Remark}. The name ``\textbf{Gaussian white noise}'' comes from the fact that the power spectral density $S(\omega)$ is uniform for every frequency $\omega$ (so it contains all the colors in the visible spectrum). A white noise is defined as $\Gamma(t) \sim \calN(0,\sigma^2)$ for all $t$. It is easy to show that such a $\Gamma(t)$ would satisfy the above two criteria.

Firstly, $\E[\Gamma(t)]$ is $\E[\Gamma(t)] = 0$ by construction (since $\Gamma(t) \sim \calN(0,\sigma^2)$). Secondly, if $\Gamma(t) \sim \calN(0,\sigma^2)$, it is necessary that $R(\tau) = \E[\Gamma(t+\tau)\Gamma(t)]$ is a delta function. Wiener-Khinchin Theorem then states that the power spectral density is flat because it is the Fourier transform of a delta function. The figures below shows the random realizations of a white noise, and their autocorrelation functions.

\begin{center}
\begin{tabular}{cc}
\includegraphics[width=0.47\linewidth]{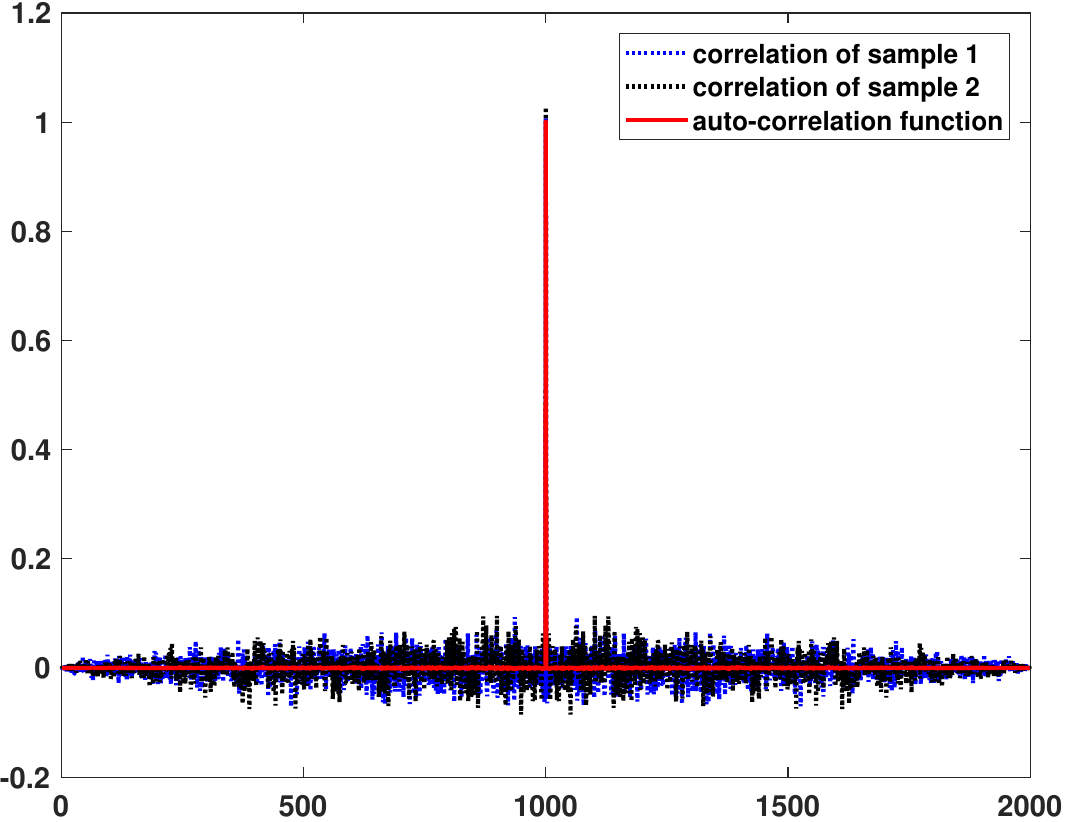}&
\includegraphics[width=0.47\linewidth]{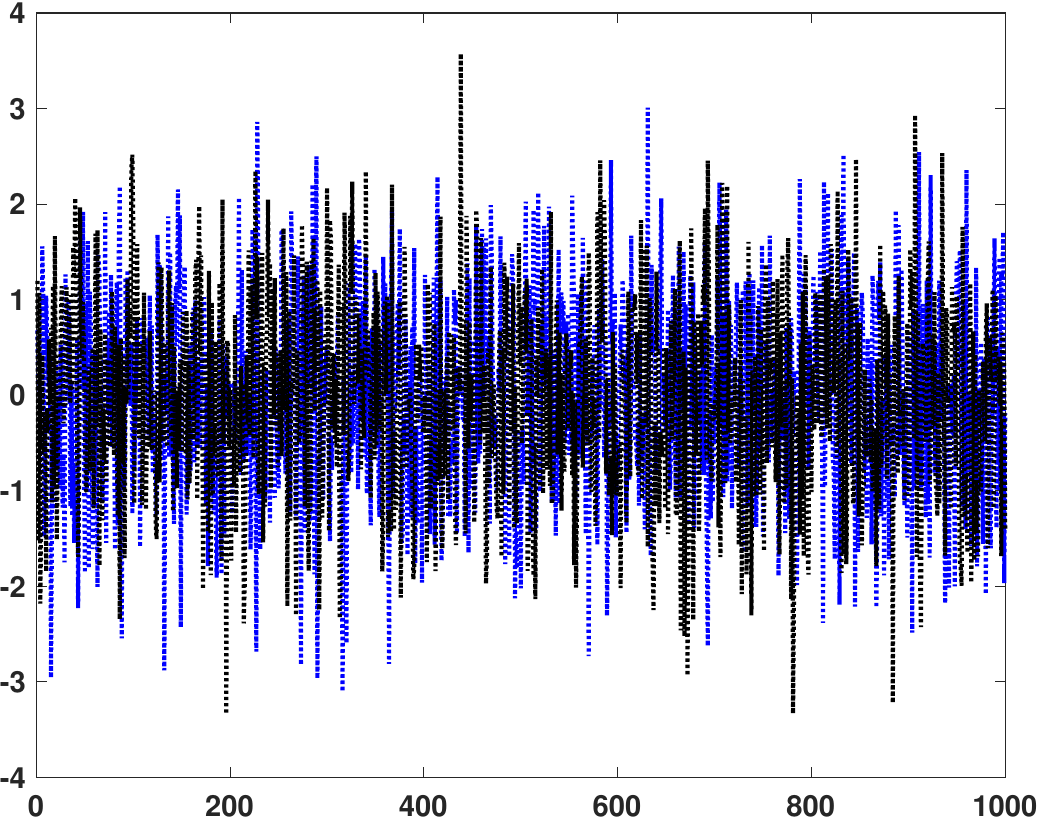}\\
(a) $R(\tau)$ & (b)  $\Gamma(t)$
\end{tabular}
\captionof{figure}{(a) Autocorrelation function $R(\tau)$ of a white Gaussian noise. (b) The random realization of the random process $\Gamma(t)$.}
\end{center}
}

\textbf{From Physics to Generative AI}. Because of the randomness exhibited in $\Gamma(t)$, the differential equation given by \eref{eq: Brownian main} is a stochastic differential equation (SDE). The solution to this SDE is therefore a random process where the value $v(t)$ is a random variable at any time $t$. Brownian motion refers to the trajectory of this random process $v(t)$ as a function of time. The resulting SDE in \eref{eq: Brownian main} is a special case of the \emph{Langevin equation}. We call it a linear Langevin equation with a $\delta$-correlated Langevin force:
\boxeddef{
\label{def: Langevin Brownian}
A \textbf{linear Langevin equation} with a $\delta$-correlated Langevin force is a stochastic differential equation of the form
\begin{equation}
\dot{\xi} + \gamma \xi = \Gamma(t),
\label{eq: Brownian main 2}
\end{equation}
where $\Gamma(t)$ is a random process satisfying two properties that (i) $\E[\Gamma(t)] = 0$ for all $t$ and (ii) $\E[\Gamma(t)\Gamma(t')] = q \delta(t-t')$ for all $t$ and $t'$.
}

At this point we can connect the Langevin equation in \eref{eq: Brownian main 2} with a diffusion model, e.g., DDPM.
\boxedeg{
\textbf{Forward DDPM}. Recall that a DDPM forward diffusion equation is given by
\begin{equation*}
d\vx =
-\underset{=f(t)}{\underbrace{\frac{\beta(t)}{2}}} \; \vx
\; dt +
\underset{=g(t)}{\underbrace{\sqrt{\beta(t)}}}d \vw.
\end{equation*}
Expressing it in the Langevin equation form, we can write it as
\begin{equation*}
\dot{\vxi}(t) + f(t)\vxi(t) = g(t)\Gamma(t), \qquad \text{where} \quad \Gamma(t) \sim \calN(0,\mI).
\end{equation*}
}

\boxedeg{
\label{example: reverse DDPM}
\textbf{Reverse DDPM}. The reverse DDPM diffusion is given by \eref{eq: SDE reverse DDPM}
\begin{equation*}
d\vx  = \underset{=\vf(\vxi,t)}{\underbrace{-\beta(t)\left[ \frac{\vx}{2} + \nabla_{\vx}\log p_t(\vx) \right]}} dt +
\underset{=g(t)}{\underbrace{\sqrt{\beta(t)}}}d\overline{\vw}.
\end{equation*}
Expressing it in the Langevin equation form, we can write it as
\begin{equation*}
\dot{\vxi}(t) = \vf(\vxi,t) + g(t)\Gamma(t), \qquad \text{where} \quad \Gamma(t) \sim \calN(0,\mI).
\end{equation*}
}
We can continue these examples for other diffusion models such as SMLD. We leave these as exercises for the readers. Our bottom message is that the diffusion equations we see in the previous chapters can all be formulated through the Langevin equation. Therefore, if we want to know the probability distributions of what these diffusion equations produce, we should look for tools in the literature of Langevin equations.

\textbf{Solution to (Linear) Langevin Equation}. The linear Langevin equation presented in \eref{eq: Brownian main 2} is a simple one. It is possible to analytically derive the solution $\xi(t)$ at any time $t$.

We start by considering the simpler problem where $\Gamma(t) = 0$. In this case, the differential equation is
\begin{equation*}
\dot{\xi}(t) + \gamma \xi(t) = 0,
\end{equation*}
and it is called a first-order homogeneous differential equation. The solution of this differential equation is as follows.

\boxedthm{
Consider the following differential equation
\begin{equation*}
\dot{\xi}(t) + \gamma \xi(t) = 0,
\end{equation*}
with an initial condition $\xi(0) = \xi_0$. The solution is given by
\begin{equation}
\xi(t) = \xi_0 e^{-\gamma t}.
\end{equation}
}

\boxedproof{
\textbf{Proof}. By rearranging the terms, we can show that
\begin{equation*}
\frac{\dot{\xi}(t)}{\xi(t)} = -\gamma,
\end{equation*}
where we assume that $\xi(t) \not= 0$ for all $t$ so that we can take $1/\xi(t)$. Integrating both sides will give us
\begin{equation*}
\int_{0}^{t} \frac{\dot{\xi}(t')}{\xi(t')} \; dt' = - \int_0^{t} \gamma \; dt'.
\end{equation*}
The left hand side of the equation will give us $\log \xi (t) - \log \xi (0)$ where as the right hand side will give us $-\gamma t$. Equating them will give us
\begin{equation*}
\log \xi(t) - \log \xi(0) = -\gamma t \quad \Longrightarrow \quad \xi (t) = \xi_0e^{-\gamma t}.
\end{equation*}
}

Now let's consider the case where $\Gamma(t)$ is present. The differential equation becomes
\begin{equation*}
\dot{\xi} + \gamma \xi = \Gamma(t),
\end{equation*}
which is called a first-order non-homogeneous differential equation. To solve this differential equation, we employ a technique known as the \emph{variation of parameter} or \emph{variation of constant} \cite[Theorem 1.2.3]{Nagy_2024_note}. The idea can be summarized in two steps. We know from our previous derivation that the solution to a homogeneous equation is $\xi (t) = \xi_0e^{-\gamma t}$. So let's make an educated guess about the solution of the non-homogenous case that the solution takes the form of $s(t) = A(t)e^{-\gamma t}$ for some $A(t)$. For notation simplicity we define $h(t) = e^{-\gamma t}$. If $s(t)$ is indeed the solution to the differential equation, then we can evaluate $\dot{s}(t) + \gamma s(t) = \Gamma(t)$. The left hand side of this equation is
\begin{align*}
\dot{s}(t) + \gamma s(t)
&= [A(t)h(t)]' + \gamma [A(t) h(t)] \\
&= A'(t)h(t) + A(t)h'(t) + \gamma A(t) h(t)\\
&= A(t)[ h'(t) + \gamma h(t)] + A'(t) h(t)\\
&= A'(t) h(t),
\end{align*}
where the last equality follows from the fact that $h(t) = e^{-\gamma t}$ is a solution to the homogeneous equation, hence $h'(t) + \gamma h(t) = 0$. Therefore, for $\dot{s}(t) + \gamma s(t)  = \Gamma(t)$, it is necessary for $A'(t) h(t) = \Gamma(t)$ by finding an appropriate $A'(t)$. But this is not difficult. The equation $A'(t) h(t) = \Gamma(t)$ can be written as
\begin{align*}
A'(t) e^{-\gamma t} = \Gamma(t) \qquad \Rightarrow \qquad A'(t) = e^{\gamma t}\Gamma(t).
\end{align*}
Integrating both side will give us
\begin{equation*}
A(t) = \int_{0}^{t} e^{\gamma t'} \Gamma(t') \; dt',
\end{equation*}
and since $s(t) = A(t)e^{-\gamma t}$, we can show that
\begin{align*}
s(t)
&= e^{-\gamma t} \int_{0}^{t} e^{\gamma t'} \Gamma(t') \; dt' = \int_{0}^{t} e^{-\gamma (t-t')} \Gamma(t') \; dt'.
\end{align*}
Therefore, the complete solution (which is the sum of the homogeneous part and the non-homogeneous part) is
\begin{equation*}
\xi(t) = \xi_0 e^{-\gamma t} + \int_{0}^{t} e^{-\gamma (t-t')} \Gamma(t') \; dt'.
\end{equation*}
We summarize the result as follows.
\boxedthm{
Consider the following differential equation
\begin{equation*}
\dot{\xi}(t) + \gamma \xi(t) = \Gamma(t),
\end{equation*}
with an initial condition $\xi(0) = \xi_0$. The solution is given by
\begin{equation}
\xi(t) = \xi_0 e^{-\gamma t} + \int_{0}^{t} e^{-\gamma (t-t')} \Gamma(t') \; dt'.
\end{equation}
}

\textbf{Distribution at Equilibrium}. The previous result shows that the solution $\xi(t)$ is a function of a random process $\Gamma(t)$. Since we do not know the particular realization of $\Gamma(t)$ every time we run the (Brownian motion) experiment, it is often more useful to characterize $\xi(t)$ by looking at the probability distribution of $\xi(t)$. In what follows, we follow Risken \cite{Risken_1989} to analyze the probability distribution at the equilibrium where $t\rightarrow \infty$ and $\xi(t) \rightarrow x$.

\boxedthm{
\label{thm: Langevin Brownian solution}
Consider the Langevin equation in Definition~\ref{def: Langevin Brownian} where
\begin{equation}
\dot{\xi} + \gamma \xi = \Gamma(t),
\end{equation}
and $\Gamma(t)$ is a white Gaussian noise so that it satisfies the properties aforementioned. Let $\xi(t)=x$ be the solution at equilibrium to this SDE, and let $p(x)$ be the probability distribution of $x$. It holds that
\begin{equation}
p(x) = \frac{1}{ \sqrt{2\pi \sigma^2}} e^{-\frac{x^2}{2 \sigma^2 }},
\end{equation}
where $\sigma = \sqrt{\frac{q}{2\gamma}}$. In other words, the solution $\xi(t)=x$ at equilibrium is a zero-mean Gaussian random variable.
}

\boxedproof{
\textbf{Proof}. Let $\xi_0 = \xi(0)$ be the initial condition. Then, the solution of the SDE takes the form
\begin{equation}
\xi(t) = \xi_0e^{-\gamma t} + \int_0^t e^{-\gamma(t-t')} \Gamma(t') dt'.
\label{eq: Linear Langevin solution}
\end{equation}
At equilibrium when $t \rightarrow \infty$, we can drop $\xi_0 e^{-\gamma t}$. Moreover, by letting $\tau = t-t'$, we can write the solution as
\begin{align*}
\xi(t) = \int_0^{\infty} e^{-\gamma (t-t')} \Gamma(t') dt' = \int_0^{\infty} e^{-\gamma \tau} \Gamma(t-\tau) d\tau.
\end{align*}

The probability density function $p(\xi)$ can be determined by taking the inverse Fourier transform of the characteristic function. Recall that the characteristic function of a random variable $v(t)$ is
\begin{align*}
C(u) = \E[\exp\{iu \cdot \xi(t)\}] = 1 + \sum_{n=1}^\infty \frac{(iu)^n}{n!}\E[\xi(t)^n].
\end{align*}
So, to find $C(u)$ we need to determine the moments $\E[\xi(t)^n]$. Using a result in Risken (Chapter 3 Eqns 3.26 and 3.27), we can show that
\begin{align}
\E[\xi(t)^{2n+1}] &= 0 \notag \\
\E[\xi(t)^{2n}]   &= \frac{(2n)!}{2^n n!}\left[\int_0^{\infty} \int_0^{\infty} e^{-\gamma(\tau_1+\tau_2)} q \delta(\tau_1-\tau_2) d\tau_1 d\tau_2\right]^n \\
&= \frac{(2n)!}{2^n n!} \left[ q \int_0^{\infty} e^{-2\gamma \tau_2} d\tau_2\right]^n = \frac{(2n)!}{2^n n!} \left[ \frac{q}{2\gamma} \right]^n. \label{eq: Linear Langevin solution 1}
\end{align}
Substituting this result into the characteristic function will give us
\begin{align}
C(u) &= 1 + 0 + \frac{2!}{2} \left[ \frac{q}{2\gamma} \right] + 0 + \frac{(2\cdot 2)!}{2^2 2!} \left[ \frac{q}{2\gamma} \right]^2 + \ldots \notag \\
&= \sum_{n=0}^{\infty} \frac{(iu)^{2n} \E[\xi(t)^{2n}]}{(2n)!} \notag \\
&= \sum_{n=0}^{\infty} \frac{(iu)^{2n}}{(2n)!} \cdot \frac{(2n)!}{2^n n!} \left[ \frac{q}{2\gamma} \right]^n \notag\\
&= \sum_{n=0}^{\infty} \frac{1}{n!} \left( -\frac{u^2 q}{4\gamma}\right)^n = e^{-\frac{u^2q}{4\gamma}}.
\end{align}
Recognizing that this is the characteristic function of a Gaussian, we can use inverse Fourier transform to retrieve the probability density function
\begin{equation*}
p(x) = \sqrt{\frac{\gamma}{\pi q}}e^{-\frac{\gamma x^2}{q}}.
\end{equation*}
}

\boxedeg{
\textbf{(Forward DDPM Distribution at Equilibrium)}
Let's do a sanity check by applying our result to the forward DDPM equation, and see what probability distribution will we obtain at the equilibrium state.

For simplicity let's assume a constant learning rate for the DDPM equation
\begin{equation*}
d\vx =
-\frac{\beta}{2} \; \vx
\; dt +
\sqrt{\beta}d \vw.
\end{equation*}
The associate Langevin equation is
\begin{equation*}
\dot{\xi}(t) + \frac{\beta}{2}\xi(t) = \sqrt{\beta}\Gamma(t).
\end{equation*}
As $t \rightarrow \infty$, our theorem above suggests that
\begin{equation*}
p(x) = \sqrt{\frac{\gamma}{\pi q}}e^{-\frac{\gamma x^2}{q}} = \frac{1}{\sqrt{2\pi}}e^{-\frac{x^2}{2}},
\end{equation*}
where we substituted $\gamma = \beta/2$, and $q = \sqrt{\beta}^2 = \beta$. Therefore, the probability distribution of $\xi(t)$ when $t \rightarrow \infty$ is $\calN(0,1)$. This is consistent with what we expect.
}

\vspace{2ex}
\textbf{Wiener Process}. In the special case where $\gamma = 0$, the linear Langevin equation is simplified to
\begin{equation*}
\dot{\xi} = \Gamma(t).
\end{equation*}
By letting $\gamma = 0$ in \eref{eq: Linear Langevin solution}, we can show that
\begin{align*}
\xi(t) = \xi_0 + \int_0^t \Gamma(t') dt',
\end{align*}
which is also known as the Wiener process. The probability distribution of the solution of the Wiener process can be derived as follows.

\boxedthm{\textbf{Wiener Process}. Consider the Wiener process
\begin{equation}
\dot{\xi} = \Gamma(t),
\label{eq: Wiener process}
\end{equation}
where $\Gamma(t)$ is the Gaussian white noise with $\E[\Gamma(t)]=0$ and $\E[\Gamma(t)\Gamma(t')]=q\delta(t-t')$. The probability distribution $p(x,t)$ of the solution $\xi(t)$ where $\xi(t) = x$ is
\begin{equation}
p(x,t) = \frac{1}{\sqrt{2\pi q t}} e^{-\frac{(x-\xi_0)^2}{2qt}}.
\end{equation}
}

\boxedproof{
\textbf{Proof}. The main difference between this result and Theorem~\ref{thm: Langevin Brownian solution} is that here we are interested in the distribution at any time $t$. To do so, we notice that $\xi(t) = \xi_0 + \int_0^t \Gamma(t') dt'$. So, to eliminate the non-zero mean, we can consider $\xi(t) - \xi_0$ instead. Substituting this into \eref{eq: Linear Langevin solution 1}, we can show that
\begin{align*}
\E[(\xi(t)-\xi_0)^{2n+1}] &= 0 \notag \\
\E[(\xi(t)-\xi_0)^{2n}]   &= \frac{(2n)!}{2^n n!} \left[ q \int_0^t e^{0} d\tau_2\right]^n = \frac{(2n)!}{2^n n!} (qt)^n.
\end{align*}
This implies that the characteristic function of $\xi(t)-\xi_0$ is
\begin{align}
C(u) &= \sum_{n=0}^{\infty} \frac{1}{n!} \left( -\frac{u^2 qt}{2}\right)^n = e^{-\frac{u^2 qt}{2}}.
\end{align}
Taking the inverse Fourier transform will give us the probability distribution for $\xi(t)$:
\begin{equation}
p(x,t) = \frac{1}{\sqrt{2\pi q t}}e^{-\frac{(x-\xi_0)^2}{2qt}}.
\end{equation}
}

To gain insights about this equation, let's assume $\xi_0 = 0$ and $q = 2k$ for some constant $k$. This will give us
\begin{equation*}
p(x,t) = \frac{1}{\sqrt{4\pi k t}}e^{-\frac{x^2}{4kt}}.
\end{equation*}
An interesting observation of this result, which can be found in many thermal dynamics textbooks, is that the probability distribution $p(x,t)$ derived above is in fact the solution of the \textbf{heat equation}:
\begin{equation}
\frac{\partial}{\partial t} p(x,t) = k\frac{\partial^2 }{\partial x^2} p(x,t),
\end{equation}
assuming that the initial condition is $p(x,0) = \delta(x)$. To see this, we just need to substitute the probability distribution into the heat equation. We can then see that
\begin{align*}
\frac{\partial}{\partial t} p(x,t)
&= \frac{1}{2t} \cdot \left(\frac{x^2}{2kt}-1\right)\cdot \frac{1}{\sqrt{4\pi k t}}e^{-\frac{x^2}{4kt}}\\
\frac{\partial^2}{\partial x^2} p(x,t)
&= \frac{1}{2kt} \cdot \left(\frac{x^2}{2kt}-1\right)\cdot \frac{1}{\sqrt{4\pi k t}}e^{-\frac{x^2}{4kt}}.
\end{align*}

\begin{figure}[h]
\begin{tabular}{cc}
\includegraphics[width=0.475\linewidth]{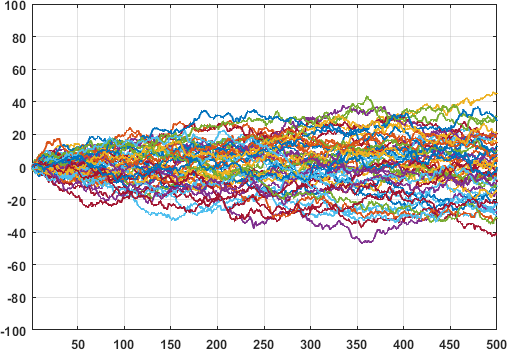}&
\hspace{-2ex}
\includegraphics[width=0.475\linewidth]{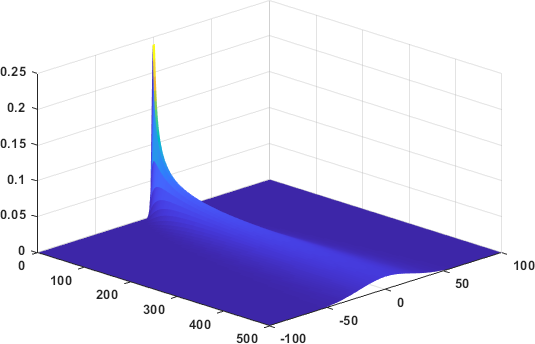}\\
(a) Realizations of $\xi(t)$ & (b) $p(x,t)$
\end{tabular}
\caption{Realization of a Wiener process. (a) The random process follows the stochastic differential equation. We show a few realizations of the random process. (b) The underlying probability distribution $p(x,t)$. As $t$ increases, the variance of the Gaussian also increases.}
\label{fig: Heat equation}
\end{figure}

The solution to the heat equation behaves like a Gaussian starting at the origin and expanding outward as time increases. The significance of this result is that while it is relatively difficult to know the exact trajectory of the random process $\xi(t)$ defined by the Langevin equation, the heat equation provides a full picture of the probability distribution. \fref{fig: Heat equation} shows the random realization of a Wiener process $\xi(t)$, and its corresponding probability distribution $p(x,t)$.

For more complicated Langevin equations (involving nonlinear terms), it seems natural to expect a similar partial equation characterizing the probability distribution. More specifically, it seems reasonable to expect that on one side of the equation, we will have $\partial/\partial t$. And on the other side of the equation, we will have $\partial^2/\partial x^2$. As we will show later, the Fokker-Planck equation will have a form similar to this. Indeed, one can derive the heat equation from the Fokker-Planck equation.

\vspace{2ex}
\textbf{Remark}: For a system of homogeneous differential equations of the form
\begin{align*}
\dot{\xi}_i + \sum_{j=1}^N \gamma_{ij} \xi_j = \Gamma_i(t), \qquad i = 1,\ldots,N,
\end{align*}
with $\E[\Gamma_i(t)] = 0$ and $\E[\Gamma_i(t)\Gamma_j(t)] = q_{ij}\delta(t-t')$, and $q_{ij} = q_{ji}$, the corresponding random process is known as the \textbf{Ornstein-Uhlenbeck} process.

\subsection{Masters Equation}
Thus far we have been studying the linear Langevin equation $\dot{\xi}(t) + \gamma \xi(t) = \Gamma(t)$. This equation allows us to handle most of the \emph{forward} diffusions where the goal is to add noise to the sample (i.e., convert the input distribution to a Gaussian.) For \emph{reverse} diffusions such as the reverse DDPM equation and the reverse SMLD equation, we would need something more general. The equation we consider now is the nonlinear Langevin equation, expressed as follows.
\boxeddef{
\label{def: nonlinear Langevin equation}
A \textbf{Nonlinear Langevin Equation} takes the form of
\begin{equation}
\dot{\xi} = h(\xi, t) + g(\xi,t)\Gamma(t),
\end{equation}
where $h(\xi,t)$ and $g(\xi,t)$ are functions denoting the drift and diffusion, respectively. Like before, we assume that $\Gamma(t)$ is a Gaussian white noise so that $\E[\Gamma(t)] = 0$ for all $t$, and $\E[\Gamma(t)\Gamma(t')] = 2\delta(t-t')$.
}
Readers can refer to Example~\ref{example: reverse DDPM} to see how the reverse DDPM would fit this equation.

The difficulty of analyzing the nonlinear Langevin equation is that there is no simple closed-form solution. Therefore, we need to develop some mathematical tools to help us understand the nonlinear Langevin equation.

\vspace{4ex}
\textbf{Markov Property}. Let's first define a Markov process. Suppose that $\xi(t)$ has a value $x_n = \xi(t_n)$ at time $t_n$, and let $t_1 \le t_2 \ldots \le t_n$. We will use the notation $p(x_n, t_n)$ to describe the probability density of having $\xi(t_n) = x_n$. We also introduce the following short-hand notation
\begin{equation*}
\vx_n = [x_n,\ldots,x_1], \qquad\mbox{and}\qquad \vt_n = [t_n,\ldots,t_1].
\end{equation*}
Therefore, $p(\vx_n,\vt_n) = p(x_n,t_n,\ldots,x_1,t_1)$ is the joint distribution of $(\xi(t_n),\ldots,\xi(x_1))$.

Let's define a Markov process. We say that a random process $\xi(t)$ is Markovian if the following memoryless condition is met.
\boxeddef{
A random process $\xi(t)$ is said to be a \textbf{Markov process} if
\begin{equation}
p(x_n, t_n \;|\;\vx_{n-1}, \vt_{n-1}) = p(x_n, t_n \;|\; x_{n-1}, t_{n-1}).
\end{equation}
That is, the probability of getting state $x_n$ at $t_n$ given \emph{all} the previous states is the same as if we are only conditioning on the immediate previous state $x_{n-1}$ at $t_{n-1}$.
}

The random process $\xi(t)$ satisfying the nonlinear Langevin equation defined in Definition~\ref{def: nonlinear Langevin equation} is Markov, as long as $\Gamma(t)$ is $\delta$-correlated. That means, the conditional probability at $t_n$ only depends on that value at $t_{n-1}$. The reason was summarized by Risken \cite{Risken_1989}: (i) A first-order differential equation is uniquely determined by its initial value; (ii) A $\delta$-correlated Langevin force $\Gamma(t)$ at a former time $t < t_{n-1}$ cannot change the conditional probability at a later time $t > t_{n-1}$. Risken further elaborates that the Markovian property is destroyed if $\Gamma(t)$ is no longer $\delta$-correlated. For example, if $\Gamma(t)$ is such that $\E[\Gamma(t)\Gamma(t')] = \frac{q}{2\gamma}e^{-\gamma|t-t'|}$, then the process described by $\dot{\xi}(t) = h(\xi) + \Gamma(t)$ will be non-Markovian. From now on, we will focus only on the Markov processes.

\vspace{4ex}
\textbf{Chapman-Kolmogorov Equation}. Consider a Markov process $\xi(t)$. We can derive a useful result known as the Chapman-Kolmogorov equation. The Chapman-Kolmogorov equation states that the joint distribution at $t_3$ and $t_1$ can be found by integrating the conditional probabilities of $t_3$ given $t_2$ and then $t_2$ given $t_1$. The two key arguments here are the Bayes Theorem plus the definition of marginalization, and the memoryless property of a Markov process.

\boxedthm{
\textbf{Chapman-Kolmogorov Equation}. Let $\xi(t)$ be a Markov process, and let $x_n = \xi(t_n)$ be the state of $\xi(t)$ at time $t_n$. Then
\begin{equation}
p(x_3,t_3 \;|\; x_1,t_1) = \int p(x_3,t_3|x_2,t_2)p(x_2,t_2|x_1,t_1)dx_2,
\end{equation}
assuming $t_1 \le t_2 \le t_3$.
}

\boxedproof{
\textbf{Proof}. For notational simplicity, let's denote $\vx_n = \{x_n,\ldots,x_1\}$ and $\vt_n = \{t_n,\ldots,t_1\}$. If $\xi(t)$ is Markov, then by the memoryless property of Markov, we have that
\begin{align*}
p(\vx_n,\vt_n)
&= p(x_n,t_n|\vx_{n-1},\vt_{n-1})p(\vx_{n-1},\vt_{n-1})\\
&= p(x_n,t_n|x_{n-1},t_{n-1})p(\vx_{n-1},\vt_{n-1}).
\end{align*}
Repeating the above argument will give us
\begin{align*}
p(\vx_n,\vt_n) = p(x_n,t_n|x_{n-1},t_{n-1})  \cdots p(x_2,t_2|x_1,t_1) \cdot p(x_1,t_1).
\end{align*}
Consequently, by marginalizing $p(\vx_3,\vt_3)$ over $x_2$, we can show that
\begin{align*}
p(x_3,t_3,x_1,t_1)
&= \int p(\vx_3,\vt_3) dx_2\\
&= \int p(x_3,t_2|x_2,t_2) p(x_2,t_2|x_1,t_1) p(x_1,t_1) dx_2.
\end{align*}
By writing $p(x_3,t_3,x_1,t_1) = p(x_3,t_3 \;|\; x_1,t_1)p(x_1,t_1)$, we can show that
\begin{equation}
p(x_3,t_3 \;|\; x_1,t_1) = \int p(x_3,t_2|x_2,t_2) p(x_2,t_2|x_1,t_1) dx_2.
\end{equation}
This completes the proof.
}

As a corollary of the Chapman-Kolmogorov equation, we can let $x = x_3$ be the current state, $x_0 = x_1$ be the initial state, and $x' = x_2$ be the intermediate state. Then the equation can be written as
\begin{equation}
p(x,t \;|\; x_0,t_0) = \int p(x,t|x',t')p(x',t'|x_0,t_0)dx',
\end{equation}
If we further drop the conditioning on $x_0$, we can write
\begin{equation*}
p(x,t ) = \int p(x,t|x',t')p(x',t')dx'.
\end{equation*}

\vspace{2ex}
\textbf{Masters Equation}. Based on the Chapman-Kolmogorov equation, we can derive a fundamental equation for Markov processes. This equation is called the Masters Equation.

\boxedthm{
Let $\xi(t)$ be a Markov process. The \textbf{Masters Equation} states that
\begin{equation}
\frac{\partial}{\partial t} p(x,t) = \int \Big[ W(x|x') p(x',t) - W(x'|x)p(x,t) \Big] \; dx',
\end{equation}
where $W(x|x')$ is the probability density function per unit time.
}

\boxedproof{
\textbf{Proof}.
Recall the Chapman-Kolmogorov Equation
\begin{equation}
p(x_3,t_3 \;|\; x_1,t_1) = \int p(x_3,t_3|x_2,t_2)p(x_2,t_2|x_1,t_1)dx_2.
\end{equation}
We consider the following mapping:
\begin{align*}
(x_1,t_1) &\longrightarrow (x_0,t_0)\\
(x_2,t_2) &\longrightarrow (x,t) \\
(x_3,t_3) &\longrightarrow (x,t+\Delta t)
\end{align*}
Then, the Chapman-Kolmogorov Equation can be written as
\begin{equation}
p(x, t+\Delta t \;|\; x_0, t_0)
= \int p(x,t+\Delta t \;|\; x', t) p(x',t \;|\; x_0, t_0) \; \; dx'.
\end{equation}

Since our goal is to obtain the partial derivative in time, we consider the time derivative of $p(x,t\;|\;x_0,t_0)$:
\begin{align*}
\frac{\partial }{\partial t} p(x,t \;|\; x_0, t_0)
&=
\lim_{\Delta t \rightarrow 0}
\frac{p(x,t+\Delta t \;|\; x_0, t_0)-p(x,t\;|\;x_0,t_0)}{\Delta t}\\
&=
\lim_{\Delta t \rightarrow 0}
\frac{ \int p(x,t+\Delta t \;|\; x', t) p(x',t \;|\; x_0, t_0) dx'-p(x,t\;|\;x_0,t_0)}{\Delta t}.
\end{align*}

We note that on the right-hand side of the equation above, there is an integration. If we switch the variables $x$ and $x'$, we can use the following observation:
\begin{align*}
	\int p(x',t+\Delta t \;|\; x, t) dx' = 1.
\end{align*}
We can insert it into the above equation and obtain
\begin{align*}
\lim_{\Delta t \rightarrow 0}
\frac{1}{\Delta t}
\Big[
\int p(x,t+\Delta t \;|\; x', t) p(x',t \;|\; x_0, t_0) dx'-
\int p(x',t+\Delta t \;|\; x,t) p(x,t \;|\; x_0, t_0) dx'
\Big]
\end{align*}

Next, we can move the limits into the integration. Let's define
\begin{align*}
W(x,t \;|\; x', t) &= \lim_{\Delta t \rightarrow 0} \frac{1}{\Delta t} p(x,t+\Delta t \;|\; x', t)\\
W(x',t \;|\; x, t) &= \lim_{\Delta t \rightarrow 0} \frac{1}{\Delta t} p(x',t+\Delta t \;|\; x, t)
\end{align*}
So, we have
\begin{equation}
\frac{\partial }{\partial t} p(x,t \;|\; x_0, t_0)
=
\int\Big[ W(x,t\;|\; x', t) p(x',t \;|\; x_0, t_0) - W(x',t\;|\; x,t) p(x,t \;|\; x_0, t_0)\Big] dx'
\end{equation}
If we fix $(x_0,t_0)$, then we can drop the conditioning. This will give us
\begin{equation}
	\frac{\partial }{\partial t} p(x,t)
	=
	\int\Big[ W(x,t\;|\; x', t) p(x',t ) - W(x',t\;|\; x,t) p(x,t)\Big] dx'.
\end{equation}
}
In the derivation above, the terms $W(x,t \;|\; x',t)$ and $W(x',t \;|\; x,t)$ are known as the \textbf{transition rates}. They are the transition probability \emph{per unit time}, with a unit $[\text{time}^{-1}]$. Thus, if we integrate them with respect to time, we will obtain
\begin{align*}
\int W(x,t \;|\; x',t) dt &= p(x,t \;|\; x',t)\\
\int W(x',t\;|\;  x,t) dt &= p(x',t\;|\; x, t).
\end{align*}

One way to visualize the Masters Equation is to consider $\int W(x,t\;|\; x', t) p(x',t ) dx'$ as the \emph{in-flow} and $\int W(x',t\;|\; x,t) p(x,t) dx'$ as the \emph{out-flow} of the transition probability from state $x'$ to $x$ (and from $x$ to $x'$). So if we view the probability as the density of particles in a room, then the Masters Equation says that the rate of the change of the density is the difference between the in-flow and the out-flow of the particles:
\begin{equation}
	\underset{\text{rate of change}}{\underbrace{\frac{\partial }{\partial t} p(x,t)}}
	=
	\underset{\text{in-flow of probability}}{\underbrace{\int\Big[ W(x,t\;|\; x', t) p(x',t )\Big] dx'}}
  - \underset{\text{out-flow of probability}}{\underbrace{\int\Big[W(x',t\;|\; x,t) p(x,t)\Big] dx'}}.
\end{equation}

Masters Equation is used widely in chemistry, biology, and many disciplines. The notion of in-flow and out-flow of participles is particularly useful to study the dynamics of a system. Another important aspect of the Masters Equation is that it relates time $\partial t$ with the state $d x'$. This will become prevalent in the Fokker-Planck equation.

One thing we can complain about the above proof is that although it is rigorous, it lacks the physics intuition during the proof. In what follows, we present an alternative proof which is more intuitive but less rigorous. The proof is based on a Lecture Note of Luca Donati \cite{Donati_Lecture}.
\boxedproof{
	\textbf{Intuitive Proof}. Consider a particle that can take only two states either 1 or 2. The probabilities of landing on a particular state are $p(x_1,t)$ and $p(x_2,t)$, such that $p(x_1,t)+p(x_2,t) = 1$ for any $t$. Now consider a small interval $(t,t+dt)$. In this interval, the particle can either stay at its current state or it can jump to the other state. This means that at the end of $t+dt$, we can write
	\begin{align*}
		p(x_1, t+dt)
		&= p(x_1,t)\Pb[\text{stay in $x_1$}] + p(x_2,t)\Pb[\text{move from $x_2$ to $x_1$}]\\
		&= p(x_1,t)\Big(1-\Pb[\text{move from $x_1$ to $x_2$}]\Big) + p(x_2,t)\Pb[\text{move from $x_2$ to $x_1$}]
	\end{align*}
	Let define the rate $W(x_2|x_1)$ and $W(x_1|x_2)$ such that
	\begin{align*}
		\Pb[\text{move from $x_1$ to $x_2$}] &= W(x_2|x_1) dt\\
		\Pb[\text{move from $x_2$ to $x_1$}] &= W(x_1|x_2) dt.
	\end{align*}
	Notice here we have implicitly assumed that the transition distribution is Markov so that the current state only depends on its previous state and not the entire history. Then the above equation can be written as
	\begin{align*}
		p(x_1, t+dt) = p(x_1,t)(1-W(x_2|x_1) dt) + p(x_2,t)W(x_1|x_2) dt + \calO(dt^2).
	\end{align*}
	The high-order term is there to account for multiple jumps during $(t,t+dt)$, e.g., jump from $x_1$ to $x_2$ and then from $x_2$ to $x_1$ within the interval. However, this term will vanish if $dt \rightarrow 0$. By rearranging the terms, we can write
	\begin{align*}
		\frac{dp(x_1,t)}{dt} = -W(x_2|x_1)p(x_1,t) + W(x_1|x_2)p(x_2,t).
	\end{align*}
	
	We can generalize this result to multiple states to and from $x_1$. For example,
	\begin{align*}
		\frac{dp(x_1,t)}{dt} = \sum_{j\not=1} \left[-W(x_j|x_1)p(x_1,t) + W(x_1|x_j)p(x_j,t)\right].
	\end{align*}
	To make it even more general, we can consider a continuum of $x_j$. By rearranging the terms, we will obtain
	\begin{align*}
		\frac{dp(x,t)}{dt} = \int \Big[W(x|x')p(x',t) - W(x'|x)p(x,t)\Big] dx',
	\end{align*}
	which is the Masters equation.
}

\subsection{Kramers-Moyal Expansion}
With the Masters Equation developed, we can now tackle the nonlinear Langevin Equation. Recall that the nonlinear Langevin Equation does not have a closed form solution, and hence we cannot write down the probability distribution of the solution analytically. Masters Equation allows us to write down the \emph{conditions} for the probability distribution through a partial differential equation known as the Fokker-Planck Equation. The derivation of the Fokker-Planck Equation requires a mathematical result known as the Kramers-Moyal Expansion.

\boxedthm{
\label{thm: Kramers Moyal}
Let $\xi(t)$ be a Markov process and let $p(x,t)$ be the probability distribution of $\xi(t)$ taking a value $x$ at time $t$. The \textbf{Kramers-Moyal Expansion} states that
\begin{align*}
\frac{\partial}{\partial t} p(x,t) = \sum_{m=1}^\infty \left[ -\frac{\partial^m}{\partial x^m} D^{(m)}(x,t)p(x,t) \right],
\end{align*}
where the Kramers-Moyal expansion coefficients are defined as
\begin{equation*}
D^{(m)}(x,t) = \frac{1}{m!} \lim_{\Delta t\rightarrow 0} \left[\frac{1}{\Delta t} \E\left[(\xi(t+\Delta t)-x)^m\right] \Big|_{\xi(t)=x}\right].
\end{equation*}
}

\boxedproof{
\textbf{Proof}. Let's start with the Masters Equation. The Masters Equation states that
\begin{align*}
\frac{\partial }{\partial t} p(x,t \;|\; x_0, t_0)
&= \lim_{\Delta t \rightarrow 0}
\frac{1}{\Delta t}
\Bigg[
\int p(x,t+\Delta t \;|\; x', t) p(x',t \;|\; x_0, t_0) dx' \\
&\qquad\qquad\qquad -
\int p(x',t+\Delta t \;|\; x,t) p(x,t \;|\; x_0, t_0) dx'
\Bigg].
\end{align*}
We can inject a test function $\varphi(x)$ such that
\begin{align*}
\frac{\partial }{\partial t} \textcolor{blue}{\int  \varphi(x)} p(x,t \;|\; x_0, t_0) \textcolor{blue}{dx}
&=
\lim_{\Delta t \rightarrow 0} \frac{1}{\Delta t} \Bigg\{  \textcolor{blue}{\int  \varphi(x)} \int p(x,t+\Delta t \;|\; x', t) p(x',t \;|\; x_0, t_0) dx' \textcolor{blue}{dx} \\
&\qquad
- \textcolor{blue}{\int  \varphi(x)} \int p(x',t+\Delta t \;|\; x, t) p(x,t \;|\; x_0, t_0) dx' \textcolor{blue}{dx} \Bigg\}.
\end{align*}
Taylor expansion of the test function will give us an infinite series
\begin{align*}
\varphi(x) = \varphi(x') + \sum_{m=1}^{\infty} \frac{(x-x')^m}{m!} \frac{\partial^m}{\partial x^m} \varphi(x) \Big|_{x = x'}.
\end{align*}
Substituting the expansion in the Masters Equation will give us
\begin{align}
&\frac{\partial }{\partial t} \int  \varphi(x) p(x,t \;|\; x_0, t_0) dx \notag \\
&=
\lim_{\Delta t \rightarrow 0} \frac{1}{\Delta t}
\Bigg[\iint \textcolor{blue}{\varphi(x')} p(x,t+\Delta t \;|\; x', t) p(x',t \;|\; x_0, t_0) dx' dx \notag \\
&\qquad\qquad\qquad+ \iint \textcolor{blue}{\sum_{m=1}^{\infty} \frac{(x-x')^m}{m!} \frac{\partial^m}{\partial x^m} \varphi(x) \Big|_{x = x'}} p(x,t+\Delta t \;|\; x', t) p(x',t \;|\; x_0, t_0) dx'  dx \notag\\
&\qquad\qquad\qquad- \iint \varphi(x) p(x',t+\Delta t \;|\; x, t) p(x,t \;|\; x_0, t_0) dx'  dx
\Bigg] \label{eq: Kramers Moyal derivation 1}
\end{align}

We notice that the last double integral in the equation above has dummy variables $x'$ and $x$. We can switch the dummy variables, and write
\begin{align*}
\iint \varphi(\textcolor{red}{x}) p(\textcolor{blue}{x'},t+\Delta t \;|\; \textcolor{red}{x}, t) p(\textcolor{red}{x},t \;|\; x_0, t_0) dx'  dx
= \iint \varphi(\textcolor{red}{x'}) p(\textcolor{blue}{x},t+\Delta t \;|\; \textcolor{red}{x'}, t) p(\textcolor{red}{x'},t \;|\; x_0, t_0) dx'  dx
\end{align*}
Then, the first and the third double integrals in \eref{eq: Kramers Moyal derivation 1} can be canceled. This will leave us
\begin{align}
&\frac{\partial }{\partial t} \int  \varphi(x) p(x,t \;|\; x_0, t_0) dx \notag \\
&=
\lim_{\Delta t \rightarrow 0} \frac{1}{\Delta t}
\iint \sum_{m=1}^{\infty} \frac{\textcolor{blue}{(x-x')^m}}{m!} \frac{\partial^m}{\partial x^m} \varphi(x) \Big|_{x = x'}  \textcolor{blue}{p(x,t+\Delta t \;|\; x', t)} p(x',t \;|\; x_0, t_0) dx'  dx \label{eq: Kramers Moyal derivation 2}
\end{align}
Now let's define
\begin{equation*}
D^{(m)}(x',t) = \frac{1}{m!} \lim_{\Delta t \rightarrow 0} \frac{1}{\Delta t} \int (x-x')^m p(x,t+\Delta t \;|\; x', t) \; dx.
\end{equation*}
Then, the Masters Equation becomes
\begin{alignat*}{2}
&\frac{\partial }{\partial t} \int  \varphi(x) p(x,t \;|\; x_0, t_0) dx \\
&= \int \sum_{m=1}^{\infty} D^{(m)}(x',t) \frac{\partial^m}{\partial x^m} \varphi(x) \Big|_{x = x'} p(x',t \;|\; x_0, t_0) dx' & \qquad & \text{Substitute $D^{(m)}(x',t)$} \\
&= \sum_{m=1}^{\infty} \int D^{(m)}(x',t) p(x',t \;|\; x_0, t_0) \frac{\partial^m}{\partial x^m} \varphi(x) \Big|_{x = x'} dx' & \qquad & \text{Switch summation and integration} \\
&= \sum_{m=1}^{\infty} (-1)^m \int \varphi(x') \frac{\partial^m}{\partial x^m} \Big[ D^{(m)}(x,t)p(x,t,x_0,t_0) \Big] dx'      & \qquad & \text{Generalized integration by part},
\end{alignat*}
where the last step is known as the generalized integration by part which states that for any continuously differentiable functions $f$ and $g$,
\begin{align*}
\int g \cdot \frac{\partial^m f}{ \partial x^m} dx = (-1)^m \int f \frac{\partial^m g}{\partial x^m} dx.
\end{align*}

Combining all these, and recognizing that the above result holds for any arbitrary $\varphi(x)$, it follows that
\begin{equation}
\frac{\partial }{\partial t} p(x,t \;|\; x_0, t_0)
= \sum_{m=1}^{\infty} (-1)^m \frac{\partial^m}{\partial x^m} \Big[ D^{(m)}(x,t)p(x,t,x_0,t_0) \Big].
\end{equation}
If we further drop the conditioning on $(x_0,t_0)$, we will obtain
\begin{equation}
\frac{\partial }{\partial t} p(x,t)
= \sum_{m=1}^{\infty} \frac{1}{m!} (-1)^m \frac{\partial^m}{\partial x^m} \Big[ D^{(m)}(x,t)p(x,t) \Big]
\end{equation}
which completes the proof.
}

Kramers-Moyal expansion expresses the time-derivative $\partial t$ of the probability distribution of any Markov process (including the solution of the nonlinear Langevin Equation) through the spatial-derivative $\partial x$. However, the expansion has infinitely many terms. An important question now is whether we allowed to truncate any of these terms. If so, how many terms can be truncated? Pawula Theorem provides an answer to this question: \cite{Pawula_1967}
\boxedthm{
\textbf{Pawula Theorem}. The Kramers-Moyal expansion may stop at one of the following three cases:
\begin{itemize}
\item $m = 1$: The resulting differential equation is known as the Liouville Equation which is a deterministic process.
\item $m = 2$: The resulting differential equation is known as the Fokker-Planck Equation.
\item $m = \infty$, i.e., the expansion cannot be truncated.
\end{itemize}
}

\boxedproof{
\textbf{Proof}. Recall that the Kramer's-Moyal coefficients are defined as
\begin{equation*}
D^{(m)}(x,t) = \frac{1}{m!} \lim_{\Delta t\rightarrow 0} \left[\frac{1}{\Delta t} \E\left[(\xi(t+\Delta t)-x)^m\right] \Big|_{\xi(t)=x}\right].
\end{equation*}
Denote $x' = \xi(t+\Delta t)$, the subject of interest here is the $m$-th moment $\E\left[(x'-x)^m\right]$.

We apply Cauchy-Schwarz inequality which states that for any functions $f$ and $g$, and random variables $X$ and $Y$, it follows that
\begin{equation*}
\E[f(X)g(Y)]^2 \le \E[f(X)^2]\E[g(Y)^2].
\end{equation*}
We consider two possibilities:
\begin{itemize}
\item Suppose that $m \ge 3$ and $m$ is odd, then
\begin{align*}
\E[(x'-x)^m]^2 = \E\left[(x'-x)^{\frac{m-1}{2}} (x'-x)^{\frac{m+1}{2}}\right]^2 \le \E\left[(x'-x)^{m-1}\right] \E\left[(x'-x)^{m+1}\right].
\end{align*}
\item Suppose that $m \ge 4$ and $m=$ is even, then
\begin{align*}
\E[(x'-x)^m]^2 = \E\left[(x'-x)^{\frac{m-2}{2}} (x'-x)^{\frac{m+2}{2}}\right]^2 \le \E\left[(x'-x)^{m-2}\right] \E\left[(x'-x)^{m+2}\right].
\end{align*}
\end{itemize}
Note that we cannot apply the above arguments for $m = 0, 1, 2$ because they will give trivial equalities.

From these two relationship, and suppose that we denote $D_m = D^{(m)}(x,t)$, then the above two cases can be written as
\begin{align*}
D_m^2 \le D_{m-1}D_{m+1}, &\qquad \text{$m$ odd and $m \ge 3$},\\
D_m^2 \le D_{m-2}D_{m+2}, &\qquad \text{$m$ even and $m \ge 4$}.
\end{align*}
Our goal now is to show that this recurring relationship will give us $D_m = 0$ for any $m \ge 3$.

Suppose first that $D_4 = 0$. Then $D_6 \le D_4D_8$ implies that $D_6 = 0$. But if $D_6 = 0$ then $D_8 \le D_6D_{10}$ implies that $D_{8} = 0$. Repeating the process will give us $D_{m} = 0$ for $m = 4, 6, 8, 10, \ldots$. Similarly, suppose that $D_6 = 0$. Then $D_4 \le D_2 D_6$ implies that $D_4 = 0$. But if $D_4 = 0$, we can go back to the first case and show that $D_m = 0$ for $m = 4, 6, 8, 10, \ldots$. In general, all even $m \ge 4$ should be zero if any one of these even $m \ge 4$ is zero. For the odd $m$'s: If $D_4 = 0$, then $D_3 \le D_2D_4$ implies that $D_3 = 0$. Similarly, if $D_6 = 0$, we will have $D_5 = 0$. So if $D_4 = D_6 = D_8 = \ldots = 0$, then $D_3 = D_5 = D_7 = \ldots = 0$. Therefore, if $D_m = 0$ for any even $m$ such that $m \ge 4$, then $D_m = 0$ all integers $m \ge 3$.

The above analysis suggests that if Kramers-Moyal expansion is truncated up to $m = 3$ so that $D_3 \not= 0$ and $D_4 = D_5 = \ldots = 0$, then $D_4 = 0$ will force $D_3 = 0$. So we will have $D_m = 0$ for all $m \ge 3$. Similarly, if the Kramers-Moyal expansion is truncated up to $m = 4$ so that $D_4 \not= 0$ and $D_5 = D_6 = \ldots = 0$, then $D_6 = 0$ will force $D_4 = 0$. So we will have $D_m = 0$ for all $m \ge 3$ again.

By repeating the above argument for other $m \ge 3$, we see that it is impossible to have Kramers-Moyal expansion be truncated for any $m \ge 3$. In other words, we can either truncate the expansion for $m=1$, $m=2$, or we never truncate it.
}

Pawula Theorem does not say that the Fokker-Planck Equation (truncating Kramers-Moyal Expansion up to $m=2$) is a good approximation to the underlying Masters Equation. It only says that we can either exactly approximate the Masters Equation using $m = 1$ or $m = 2$, or we cannot approximate at all.

\subsection{Fokker-Planck Equation}
We can now discuss the Fokker-Planck Equation. The Fokker-Planck Equation is the truncation of the Kramers-Moyal's Expansion using $m = 2$.
\boxeddef{
The \textbf{Fokker-Planck Equation} is obtained by truncating the Kramers-Moyal expansion to $m = 2$. That is, for any Markov process $\xi(t)$, the probability distribution $p(x,t)$ of $\xi(t) = x$ at time $t$ will satisfy the following partial differential equation:
\begin{equation}
\frac{\partial}{\partial t} p(x,t) = -\frac{\partial}{\partial x} D^{(1)}(x,t)p(x,t) + \frac{\partial^2}{\partial x^2} D^{(2)}(x,t)p(x,t).
\end{equation}
}
Fokker-Planck Equation is a general result for \emph{any} Markov random processes because it is a consequence of the Chapman-Kolmogorov Equation and the Masters Equation. Processes we study in this tutorial, e.g., Langevin Equation, are special cases of this big family of random processes. Therefore, if we have a Langevin Equation, it is necessary that the solution $\xi(t)$ will have a probability distribution satisfying the Fokker-Planck Equation.

\vspace{2ex}
\textbf{Nonlinear Langevin Equation}. If we focus on the nonlinear Langevin equation
\begin{equation*}
\dot{\xi} = h(\xi,t) + g(\xi,t)\Gamma(t),
\end{equation*}
we can evaluate the Kramers-Moyal coefficients $D^{(m)}(x,t)p(x,t)$. The following theorem summarizes the coefficients. We remark that during the proof of this theorem, it will become clear why $D^{(m)}(x,t)p(x,t)$ is only limited to $m=1$ and $m=2$.

\boxedthm{\textbf{Fokker-Planck for nonlinear Langevin Equation}. Consider the nonlinear Langevin equation
\begin{equation*}
\dot{\xi} = h(\xi,t) + g(\xi,t)\Gamma(t),
\end{equation*}
for functions $h(\xi,t)$ and $g(\xi,t)$. The Fokker-Planck Equation for this nonlinear Langevin equation will have Kramers-Moyal coefficients:
\begin{alignat}{2}
\text{(Drift)}        &\qquad & D^{(1)}(x,t) &= h(x,t) + g'(x,t)g(x,t)\\
\text{(Diffusion)}    &\qquad & D^{(2)}(x,t) &= g^2(x,t).
\end{alignat}
}

\boxedproof{
\textbf{Proof}. Recall the definition of the Kramers-Moyal coefficient:
\begin{equation*}
D^{(m)}(x,t) = \frac{1}{m!} \lim_{\tau\rightarrow 0} \frac{\E\left[(\xi(t+\tau)-x)^m\right]}{\tau} \Big|_{\xi(t)=x}.
\end{equation*}
The hard part is how to evaluate the moments $\E\left[(\xi(t+\tau)-x)^m\right]$.

We start by looking at the nonlinear Langevin equation:
\begin{equation*}
\dot{\xi} = h(\xi,t) + g(\xi,t)\Gamma(t).
\end{equation*}
Expressing $\xi(t+\tau)-\xi(t) = \int_{t}^{t+\tau} \dot{\xi}(t') dt'$ for a small $\tau$ and let $\xi(t) = x$, we can write
\begin{align*}
\xi(t+\tau) - x = \int_{t}^{t+\tau} \Big[ h(\xi(t'),t') + g(\xi(t'),t')\Gamma(t')\Big] dt'.
\end{align*}
We assume that $h$ and $g$ can be expanded as
\begin{align*}
h(\xi(t'),t') = h(x,t') + h'(x,t')(\xi(t')-x) + \ldots \\
g(\xi(t'),t') = g(x,t') + g'(x,t')(\xi(t')-x) + \ldots.
\end{align*}
This will give us
\begin{align*}
\xi(t+\tau) - x
&= \int_{t}^{t+\tau} \Big[h(x,t') + h'(x,t')(\xi(t')-x) + \ldots\Big] dt' \\
&\qquad + \int_{t}^{t+\tau} \Big[g(x,t') + g'(x,t')(\xi(t')-x) + \ldots\Big]\Gamma(t') d't\\
&= \int_{t}^{t+\tau} h(x,t') dt' + \int_{t}^{t+\tau} h'(x,t')(\xi(t')-x) dt' + \ldots \\
&\qquad + \int_{t}^{t+\tau} g(x,t') \Gamma(t') dt' + \int_{t}^{t+\tau} g'(x,t')(\xi(t')-x) \Gamma(t') dt' + \ldots.
\end{align*}
Now, we iterate the above equation and replace $\xi(t')-x$ by the integrations. This will give us
\begin{align}
\xi(t+\tau) - x
&= \int_{t}^{t+\tau} h(x,t') dt' + \int_{t}^{t+\tau} h'(x,t')  \left[\int_{t}^{t'} h(x,t'')dt''\right] dt' + \notag \\
&\qquad + \int_{t}^{t+\tau} h'(x,t') \left[\int_{t}^{t'} g(x,t'') \Gamma(t'') dt''\right] dt' + \ldots \notag \\
&\qquad + \int_{t}^{t+\tau} g(x,t') \Gamma(t') dt' + \int_{t}^{t+\tau} g'(x,t') \left[ \int_t^{t'} h(x,t'') dt''\right] \Gamma(t') dt' \notag \\
&\qquad + \int_{t}^{t+\tau} g'(x,t') \left[ \int_{t}^{t'} g(x,t'') \Gamma(t'') dt''\right] \Gamma(t') dt' + \ldots  \label{eq: Kramers Moyal nonlinear steps 2}
\end{align}
where we only write the terms involving $h$, $g$ and $\Gamma$. Terms involving $\xi(t'')-x$ are not dropped.

Take expectation, and noticing that $\E[\Gamma(t)] = 0$, we can show that only the first two terms and the last term in \eref{eq: Kramers Moyal nonlinear steps 2} will survive. Thus, we have
\begin{align}
\E[\xi(t+\tau) - x ]
&= \int_{t}^{t+\tau} h(x,t') dt' + \int_{t}^{t+\tau} \int_{t}^{t'} h'(x,t') h(x,t'')dt'' dt' + \ldots \notag \\
&\qquad + \int_{t}^{t+\tau} g'(x,t') \underset{=g(x,t')}{\underbrace{\int_t^{t'} g(x,t'') 2\delta(t''-t')  dt''}} dt' + \ldots.
\label{eq: Kramers Moyal nonlinear steps 3}
\end{align}
where we follow Risken's definition that $\int_t^{t'} 2\delta(t''-t') dt'' = 1$ \cite{Risken_1989}. As $\tau \rightarrow 0$, it follows that the first and third terms of \eref{eq: Kramers Moyal nonlinear steps 3} are
\begin{align*}
&\lim_{\tau \rightarrow 0} \int_{t}^{t+\tau} h(x,t') dt' = h(x,t),\\
&\lim_{\tau \rightarrow 0} \int_{t}^{t+\tau} g'(x,t') g(x,t') dt' = g'(x,t)g(x,t).
\end{align*}
For the second term in \eref{eq: Kramers Moyal nonlinear steps 3}, we can show that
\begin{align*}
\lim_{\tau \rightarrow 0} \int_{t}^{t+\tau} \int_{t}^{t'} h'(x,t') h(x,t'')dt'' dt'
&= \lim_{\tau \rightarrow 0} \int_{t}^{t+\tau} h'(x,t') (H(x,t')-H(x,t)) dt' \\
&= h'(x,t) (H(x,t)-H(x,t)) = 0,
\end{align*}
where we assume $h(x,t)$ is integrable over $t$ and so we can define $H(x,t) = \int_0^t h(x,t')dt'$. Therefore, we arrive at
\begin{align*}
D^{(1)}(x,t) = h(x,t) + g'(x,t)g(x,t).
\end{align*}

The derivation of $D^{(2)}(x,t)$ follows essentially the same set of arguments. The key to note here is that when we take the square in $\E[(\xi(t+\tau) - x )^2]$, the integrals in \eref{eq: Kramers Moyal nonlinear steps 3} will give us contributions proportional to $\tau^2$. When $\tau \rightarrow 0$, all these terms will vanish because there is only one $1/\tau$ in the definition of $D^{(2)}(x,t)$. As a result, the only term that can survive is
\begin{align*}
D^{(2)}(x,t)
&=
\frac{1}{2} \lim_{\tau\rightarrow 0} \frac{1}{\tau}
\int_{t}^{t+\tau} \int_t^{t+\tau} g(x,t') g(x,t'') 2\delta(t'-t'') dt' dt'' \\
&= g^2(x,t),
\end{align*}
which completes the proof.
}

\vspace{2ex}
\boxedeg{
Consider the following Langevin Equation
\begin{equation*}
\dot{\xi} = A(\xi) \xi + \sigma \Gamma(t).
\end{equation*}
Then, the probability distribution $p(x,t)$ for the solution $\xi(t)$ will satisfy the following Fokker-Planck equation
\begin{align*}
\frac{\partial}{\partial t} p(x,t)
&= -\frac{\partial}{\partial x} \Big[h(x,t) + g'(x,t)g(x,t)\Big]p(x,t) + \frac{\partial^2}{\partial x^2} \Big[g^2(x,t) p(x,t)\Big]\\
&= -\frac{\partial}{\partial x} \left[ \left(A(x) + 0 \cdot \sigma \right) p(x,t) \right] + \frac{\partial^2}{\partial x^2} \Big[\sigma^2 p(x,t)\Big]\\
&= -\frac{\partial}{\partial x} \left[ A(x) p(x,t) \right] + \sigma^2  \frac{\partial^2}{\partial x^2} [p(x,t)].
\end{align*}
}

\boxedeg{
For the special case where $A(x) = 0$, the Langevin equation is simplified to a Wiener process:
\begin{equation*}
\dot{\xi} = \sigma \Gamma(t).
\end{equation*}
The corresponding Fokker-Planck equation is
\begin{equation*}
\frac{\partial}{\partial t} p(x,t)  = \sigma^2  \frac{\partial^2}{\partial x^2} p(x,t).
\end{equation*}
This equation is known as the \textbf{heat equation} or the \textbf{diffusion equation}. If the initial condition is that $p(x,0)=\delta(x)$, the solution is (See derivation below.)
\begin{equation*}
p(x,t) = \frac{1}{\sqrt{4\pi \sigma^2 t}} e^{-\frac{x^2}{4\sigma^2 t}}.
\end{equation*}
}

\boxedproof{
\textbf{Solution to Heat Equation}. The heat equation can be solved using Fourier transforms. For notational simplicity we denote $u_t = \frac{\partial u}{\partial t}$ and $u_{xx} = \frac{\partial^2 u}{\partial x^2}$. Consider a generic heat equation:
\begin{equation*}
u_t(x,t) = k u_{xx}(x,t), \qquad x \in \R, t > 0,
\end{equation*}
with initial condition $u(x,0) = \phi(x)$. We can take Fourier transform (to map between $x \leftrightarrow \omega$) on both sides by defining
\begin{align*}
\widehat{u_t}(\omega,t)    &= \calF\Big\{u_t(x,t)\Big\}    = \int_{-\infty}^{\infty} u_t(x,t)    e^{j \omega x} dx\\
\widehat{u}_{xx}(\omega,t) &= \calF\Big\{u_{xx}(x,t)\Big\} = \int_{-\infty}^{\infty} u_{xx}(x,t) e^{j \omega x} dx.
\end{align*}
This will give us
\begin{align*}
\widehat{u_t}(\omega,t) = k \widehat{u}_{xx}(\omega,t).
\end{align*}
Using the differentiation property of Fourier transform, we can write the right hand side of the equation as
\begin{align*}
k \widehat{u}_{xx}(\omega,t) = k (j \omega)^2 \widehat{u}(\omega,t) = - k \omega^2 \widehat{u}(\omega,t),
\end{align*}
which will give us an ordinary differential equation
\begin{equation}
\widehat{u_t}(\omega,t) = - k \omega^2 \widehat{u}(\omega,t).
\label{eq: Heat equation Fourier}
\end{equation}
The solution to this differential equation (in $t$) is given by
\begin{equation}
\widehat{u}(\omega,t) = \widehat{\phi}(\omega)e^{-k\omega^2 t}.
\end{equation}
(Remark: \eref{eq: Heat equation Fourier} is just a simple differential equation $f'(t) = a f(t)$ whose solution can be found by integration.) Therefore, if we take the inverse Fourier transform on $\widehat{u}(\omega,t) $(with respect to $\omega \leftrightarrow x$), we will have
\begin{align*}
u(x,t)  &=  \calF^{-1}\{\widehat{u}(\omega,t)\} = \calF^{-1}\left\{\widehat{\phi}(\omega)e^{-k\omega^2 t}\right\}
\end{align*}
which is the inverse Fourier transform on the product of $\widehat{\phi}(\omega)$ and $\widehat{f}(\omega) \bydef e^{-k\omega^2 t}$. Since multiplication in the Fourier domain is convolution in the spatial domain, it follows that $u(x,t)$ is the convolution of $\phi(x)$ and $f(x) = \calF^{-1}(e^{-k\omega^2 t})$. But $\calF^{-1}(e^{-k\omega^2 t}) = \frac{1}{\sqrt{2k t}} e^{-x^2/(4kt)}$. Therefore, we can show that the solution is
\begin{align*}
u(x,t)
&= \frac{1}{\sqrt{2\pi}}\int_{-\infty}^{\infty} \phi(x-x') f(x') dx' \\
&= \frac{1}{\sqrt{2\pi}}\int_{-\infty}^{\infty} \phi(x-x') \frac{1}{\sqrt{2k t}} e^{-(x')^2/(4kt)} dx'.
\end{align*}
So, if $\phi(x) = \delta(x)$ so that $\phi(x-x') = \delta(x-x')$, it follows that
\begin{equation}
u(x,t) = \frac{1}{\sqrt{4k \pi t}}e^{-\frac{x^2}{4kt}}.
\end{equation}
}

\vspace{2ex}
\textbf{Probability Current}. The Fokker-Planck Equation has some interesting physics interpretations. Recall that the Fokker-Planck Equation is
\begin{equation}
\frac{\partial}{\partial t} p(x,t) = -\frac{\partial}{\partial x} D^{(1)}(x,t)p(x,t) + \frac{\partial^2}{\partial x^2} D^{(2)}(x,t)p(x,t).
\end{equation}
Let's define a quantity
\begin{equation}
S(x,t) = \left[D^{(1)}(x,t) - \frac{\partial}{\partial x} D^{(2)}(x,t)\right]p(x,t).
\end{equation}
Then, the Fokker-Planck Equation can be written as
\begin{equation}
\frac{\partial p}{\partial t} + \frac{\partial S}{\partial x} = 0.
\label{eq: Probability Current}
\end{equation}
One way to interpret \eref{eq: Probability Current} is that it is the \textbf{probability current}.

\boxedproof{
\textbf{Intuitive Derivation of \eref{eq: Probability Current}}. Conversation of energy tells us that if some increases/decreases in a spatial region (e.g., particles or charges), the change in the amount should be equal to the change in its surface. So, if $p(x,t)$ represents some sort of density, then $p(x,t)dx$ will be the amount of particles sitting between $(x,x+dx)$ at time $t$. Here, $S(x,t)$ can be viewed as the current of the particle flowing per unit time across $x$. For a time interval $(t,t+dt)$ and a spatial interval $(x,x+dx)$, the change in amount of particles is
\begin{equation*}
\text{number of particles increased/decreased} = [p(x,t+dt) - p(x,t)]dx.
\end{equation*}
Because of the conversation of energy, for this change to happen, there must be some flow of the current. The current over the interval is
\begin{equation*}
\text{number of particles flowing in/out} = [S(x,t) - S(x+dx,t)]dt.
\end{equation*}
By equating the two, we will obtain
\begin{equation*}
[p(x,t+dt) - p(x,t)]dx = [S(x,t) - S(x+dx,t)]dt.
\end{equation*}
Taylor approximation to the first order will give us $p(x,t+dt) - p(x,t) = \frac{\partial p}{\partial t}$ and $S(x,t) - S(x+dx,t) = \frac{\partial S}{\partial x}$. This will then give us
\begin{equation}
\frac{\partial p}{\partial t} + \frac{\partial S}{\partial x} = 0.
\end{equation}
}
Therefore, the Fokker-Planck Equation can be regarded as one form of conversation of energy where the change in $p$ (over time) should be equal to change in $S$ (over space).

\textbf{Equilibrium Solution}. At equilibrium, the current vanishes and so we have $S = 0$. Consequently, we can show the following.
\boxedthm{
At equilibrium, since the probability current vanishes, the probability distribution $p(x)$ satisfies
\begin{equation*}
D^{(1)}(x,t)p(x) = \frac{\partial}{\partial x} \left[D^{(2)}(x,t)p(x)\right].
\end{equation*}
}

\boxedeg{
For example, if $D^{(1)} = -\gamma x$ and $D^{(2)} = \frac{\gamma k T}{m}$, then $S = 0$ will give us
\begin{equation*}
\left(-\gamma x - \frac{\gamma k T}{m}\frac{\partial}{\partial x}\right)p(x) = 0.
\end{equation*}
Then $-\gamma x p(x) = \frac{\gamma k T}{m} \frac{\partial}{\partial x}p(x)$. Solving this differential equation will give us
\begin{equation*}
p(x) = \sqrt{\frac{m}{2 \pi k T}}e^{-\frac{m x^2}{2 k T}}.
\end{equation*}
}

\boxedeg{\textbf{Connection with SMLD}
Let's try to map our results with \eref{eq: Langevin dynamics main equation} defined in Definition~\ref{def: Langevin discrete}. We shall consider the 1D case. Consider the following Langevin equation
\begin{equation*}
\underset{\dot{\xi}}{\underbrace{\frac{\partial x}{\partial t} }}
=
\underset{h(\xi,t)}{\underbrace{\tau \frac{\partial }{\partial x} \log p(x)}}
+
\underset{g(\xi,t)}{\underbrace{\sigma}} \Gamma(t).
\end{equation*}
To avoid notational confusions, we let $W(x,t)$ be the probability distribution of the solution $x(t)$ for this Langevin equation. The Kramers-Moyal coefficients for this Langevin Equation are
\begin{align*}
D^{(1)}(x,t) &= h(x,t) + g'(x,t)g(x,t) = \tau \frac{\partial}{\partial x}\log p(x) \bydef A(x)\\
D^{(2)}(x,t) &= g(x,t)^2 = \sigma^2.
\end{align*}
So, the corresponding Fokker-Planck Equation is
\begin{align*}
\frac{\partial}{\partial t} W(x,t)
&= -\frac{\partial}{\partial x} \left[ D^{(1)}(x,t) W(x,t) \right] + \frac{\partial^2}{\partial x^2} \left[D^{(2)}(x,t)W(x,t)\right]\\
&= -\frac{\partial}{\partial x} \left[ A(x) W(x,t) \right] + \sigma^2 \frac{\partial^2}{\partial x^2} W(x,t).
\end{align*}

At equilibrium when $ t \rightarrow \infty$, the probability distribution $W(x,t)$ can be written as $W(x)$. Since the probability current vanishes, it follows that
\begin{equation*}
A(x) W(x) = \sigma^2 \frac{\partial}{\partial x} W(x).
\end{equation*}
Recall that $A(x) = \tau \frac{\partial}{\partial x}\log p(x)$, it follows that
\begin{equation*}
\tau \frac{\partial}{\partial x}\log p(x) = \sigma^2 \frac{\partial}{\partial x} W(x).
\end{equation*}
Since we have the freedom to choose $\sigma$, we will just make it $\sigma = \sqrt{\tau}$. Then the above equation is simplified to $\frac{\partial}{\partial x}\log p(x) = \frac{\partial}{\partial x}W(x)$. Integrating both sides with respect to $x$ will give us
\begin{equation}
\log p(x) = W(x) + C,
\label{eq: Fokker-Planck SMLD final solution}
\end{equation}
for some constant $C$. Let $U(x) = e^{W(x)}$ be a probability distribution so that $W(x) = \log U(x)$ is the log-likelihood, \eref{eq: Fokker-Planck SMLD final solution} will give us $p(x) = U(x)e^C$. Since $p(x)$ and $U(x)$ are probability distributions, we must have $\int p(x) dx = 1$ and $\int U(x) dx = 1$. Thus, we can show that $C = 0$.

Therefore, we conclude that if we run the Langevin equation until convergence, the probability distribution $W(x)$ of the solution is exactly the ground truth distribution $p(x)$. Moreover, the noise level $\sigma$ and the step size $\tau$ is related by $\sigma = \sqrt{\tau}$.
}

\subsection{Concluding Remark}
In this section we discussed the physics behind Brownian motion, Langevin equation, and Fokker-Planck equation, and demonstrated several classical theorems in the statistical physics literature. Many of our results are general, as they can be applied to any Markov random process. Going beyond the Markov processes can be done by limiting the spatial/temporal correlation to a small interval.

There are a plethora of references on this topic, many of which originated from physics. Risken's textbook \cite{Risken_1989} is a classic reference on this subject which contains essentially all the ingredients. For readers looking for something slightly more general, Reichl's statistical physics book \cite{Reichl_1998} would serve the purpose.

\newpage
\section{Conclusion}
This tutorial covers a few basic concepts underpinning the diffusion-based generative models in the recent literature. We find it particularly important to go deeper into these fundamental principles rather than staying at the surface of Python programming.

As we write this tutorial, a few lessons we learned are worth sharing. The development of DDPM is in many senses an extension of the VAE, both in terms of the structure, the usage of the evidence lower bound, and the re-parameterization trick. While DDPM is still one of the state-of-the-art methods, it would be more ideal if future models can avoid any iteration. Some recent papers are beginning to explore the feasibility of using knowledge distillation to reduce the number of iterations. Some are investigating acceleration methods by borrowing ideas from the differential equation literature.

For readers who are interested in imaging, the score matching Langevin dynamics would likely continue to play a fundamental role in solving inverse problems. The training of the score-matching function is nearly identical to training an image denoiser, and the application of a score-matching step is identical to a denoising step. Therefore, as soon as we know how to split the inverse problem into the forward model and prior distribution, we will be able to leverage score matching to perform the posterior sampling. Recent approaches have also begun to explore different forms of Bayesian diffusion models to connect the physical model and the data-driven model.

The SDE and Fokker-Planck equations offer great theoretical intuitions as to why the SMLD equation was derived in a certain way. Historically speaking, the development of SMLD does not appear to start with the SDE but the realization today helps us understand the behavior of the solution statistics.

Looking into the future, the biggest challenges of diffusion models are the consistency with the physical world, let alone their high computational complexity. Class-specific learning will continue to be influential. E.g., one can train a customized diffusion model using the gallery images on a cell phone. Temporal consistency would demand for larger models with more memory to store the frames. New architectures are needed to leverage the increasing number of temporal inputs to the model. Another open question is how to bring language and semantics into image generation. Should images continue to be represented as an array of pixels (or pixels of features), or are there new ways to describe the scene using a few words without losing information? Finally, information forensics is going to be the biggest challenge for the next decades until we can develop effective countermeasures (or policies).

\vspace{4ex}
\subsection*{Acknowledgement}
This work is supported, in part, by the National Science Foundation under the awards 2030570, 2134209, and 2133032, as well as by SRC JUMP 2.0 Center, and research awards from Samsung Research America. Since this draft was posted on the internet in March 2024, we received numerous constructive feedback from readers from all over the world. Thank you all for your input. Thanks also to many of the graduate students at Purdue who shared good thoughts about the content of the tutorial. We want to give a special thanks to William Chi-Kin Yau who worked tirelessly with us on the section about Langevin and Fokker-Planck equations.

\newpage
\bibliography{references}
\bibliographystyle{plain}

\end{document}